\documentclass[lettersize,journal]{IEEEtran}
\usepackage{amsmath,amsfonts, amsthm}
\usepackage{algorithmic}
\usepackage{algorithm}

\usepackage{array}
\usepackage{textcomp}
\usepackage{stfloats}
\usepackage{url}
\usepackage{verbatim}
\usepackage{xspace}
\usepackage{graphicx}
\usepackage{cite}
\usepackage{tikz}
\usepackage{mathtools}
\usepackage{booktabs}
\usepackage{tabularx}
\usepackage[inline]{enumitem}
\usepackage[edges]{forest}
\usepackage{longtable}
\usepackage{booktabs,siunitx}
\usetikzlibrary{
    calc,
    shapes,
    arrows,
    arrows.meta,
    decorations.markings,
    decorations.pathreplacing,
    shadows,
    positioning,
    fit,
    shapes.geometric,
    patterns,
    patterns.meta,
}

\usepackage{listings} 
\usepackage[table]{xcolor}
\usepackage{subcaption}
\usepackage[hidelinks]{hyperref}
\usepackage{cleveref}
\usepackage{qrcode}

\usepackage[acronyms,nopostdot,nomain]{glossaries}
\newif\ifprintgloss
\printglosstrue

\ifprintgloss
  \makeglossaries
\else
  \glsdisablehyper
\fi
\setacronymstyle{long-short}

\newacronym{ipm}{IPM}{interior point methods}
\newacronym{sqp}{SQP}{sequential quadratic programming}
\newacronym{nlp}{NLP}{nonlinear programming}
\newacronym{scp}{SCP}{sequential convex programming}
\newacronym{ocp}{OCP}{optimal control problem}
\newacronym{ddp}{DDP}{differential dynamic programming}
\newacronym{ilqr}{iLQR}{iterative linear quadratic regulator}
\newacronym{ilqg}{iLQG}{iterative linear quadratic Gaussian}
\newacronym{prm}{PRM}{probabilistic roadmap}
\newacronym{rrt}{RRT}{rapidly-exploring random trees}
\newacronym{scvx}{SCvx}{successive convexification}
\newacronym{ctscvx}{CT-SCvx}{continuous-time successive convexification}
\newacronym{ctcs}{CTCS}{continuous-time constraint satisfaction}
\newacronym{dcp}{DCP}{disciplined convex programming}
\newacronym{soc}{SOC}{second-order cone}
\newacronym{dpp}{DPP}{disciplined parametric programming}
\newacronym{dsl}{DSL}{domain-specific language}
\newacronym{dag}{DAG}{directed acyclic graph}
\newacronym{qp}{QP}{quadratic program}
\newacronym{stl}{STL}{signal temporal logic}
\newacronym{foh}{FOH}{first-order hold}
\newacronym{zoh}{ZOH}{zero-order hold}
\newacronym{alv}{ALV}{ascent launch vehicle problem}
\newacronym{rrv}{RRV}{reusable reentry vehicle problem}
\newacronym{hs}{HS}{hypersensitive problem}
\newacronym{mpc}{MPC}{model predictive control}
\newacronym{dof}{DoF}{degree of freedom}
\newacronym{stc}{STC}{state-triggered constraint}
\newacronym{pdg}{PDG}{powered descent guidance}
\newacronym{relu}{ReLU}{rectified linear unit}
\newacronym{dnlp}{DNLP}{disciplined nonlinear programming}
\newacronym{licq}{LICQ}{linear independence constraint qualification}
\newacronym{gmsr}{GMSR}{generalized mean-based smooth robustness measures}

\crefname{listing}{List.}{Lists.}
\Crefname{listing}{Listing}{Listings}

\crefname{section}{Sec.}{Secs.}
\crefname{subsection}{Sec.}{Secs.}
\crefname{subsubsection}{Sec.}{Secs.}
\crefname{paragraph}{Sec.}{Secs.}
\Crefname{section}{Section}{Sections}
\Crefname{subsection}{Section}{Sections}
\Crefname{subsubsection}{Section}{Sections}
\Crefname{paragraph}{Section}{Sections}

\crefname{equation}{Eq.}{Eqs.}
\crefname{subequation}{Eq.}{Eqs.}
\Crefname{equation}{Equation}{Equations}
\Crefname{subequation}{Equation}{Equations}

\crefname{figure}{Fig.}{Figs.}
\crefname{subfigure}{Fig.}{Figs.}
\Crefname{figure}{Figure}{Figures}
\Crefname{subfigure}{Figure}{Figures}

\crefname{table}{Tab.}{Tabs.}
\crefname{subtable}{Tab.}{Tabs.}
\Crefname{table}{Table}{Tables}
\Crefname{subtable}{Table}{Tables}

\crefname{appendix}{Appendix}{Appendices}
\crefname{subappendix}{Appendix}{Appendices}
\Crefname{appendix}{Appendix}{Appendices}
\Crefname{subappendix}{Appendix}{Appendices}

\crefname{algorithm}{Alg.}{Algs.}
\Crefname{algorithm}{Algorithm}{Algorithms}

\definecolor{sky-blue}{rgb}{0.36, 0.68, 0.89} % Node circle color RGB(93, 173, 226)
\definecolor{tan-color}{rgb}{0.84, 0.74, 0.69} % Tan text color RGB(213, 189, 175)
\definecolor{light-blue-color}{rgb}{0.52, 0.76, 0.91} % Light blue text color RGB(133, 193, 233)
\definecolor{ann-text-color}{rgb}{0,0,0} % Black for annotation text

\tikzset{
    mono-code-text/.style={font=\ttfamily\Large, text=black, align=center},
    operator-circle/.style={circle, draw=none, fill=sky-blue, inner sep=0pt, minimum size=1.5cm, text=white, font=\Huge},
    tree-text-tan/.style={font=\ttfamily\Large, color=tan-color, draw=none, inner sep=0.1cm},
    tree-text-blue/.style={font=\ttfamily\Large, color=light-blue-color, draw=none, inner sep=0.1cm},
    arrow-black/.style={->, >=stealth, color=black, thick},
    annotation-line/.style={-, color=black, thin},
    annotation-text/.style={font=\normalsize, color=ann-text-color},
    squiggle-red/.style={decorate, decoration={snake,amplitude=0.3mm,segment length=1.5mm}, color=red, thick}
}

\usepackage[most]{tcolorbox}
\tcbset {
  base/.style={
    arc=3mm, 
    bottomtitle=0.5mm,
    boxrule=0mm,
    colbacktitle=black!10!white,
    coltitle=black, 
    fonttitle=\bfseries, 
    left=2.5mm,
    leftrule=1mm,
    right=3.5mm,
    title={#1},
    toptitle=0.75mm, 
  }
}

\definecolor{brandblue}{rgb}{0.34, 0.7, 1}
\definecolor{brandgreen}{rgb}{0.14, 0.7, 0.4}
\definecolor{brandred}{rgb}{1, 0.2, 0.2}

\newtcolorbox{mainbox}[1]{
  enhanced,
  breakable,
  colframe=ttc-blue,
  base={\textcolor{ttc-blue}{\fontfamily{put}\selectfont\bfseries #1}}
}

\newtcolorbox{mathnote}[1]{
  enhanced,
  breakable,
  colframe=ttc-blue,
  base={\textcolor{ttc-blue}{\fontfamily{put}\selectfont\bfseries #1}}
}

\newcommand{\openscvx}{\textsc{OpenSCvx}\xspace}
\newcommand{\ctscvx}{\textsc{CT-SCvx}\xspace}

\newcommand{\jax}{\textsf{JAX}}
\newcommand{\cvxpy}{\textsf{CVXPY}}
\newcommand{\mujoco}{\textsf{MuJoCo}}
\newcommand{\fraxpkg}{\textsf{frax}}

\newcommand{\casadi}{\textsf{CasADi}}
\newcommand{\cusadi}{\textsf{CusADi}}
\newcommand{\drake}{\textsf{Drake}}
\newcommand{\jump}{\textsf{JuMP}}

\newcommand{\crocoddyl}{\textsf{Crocoddyl}}
\newcommand{\altro}{\textsf{ALTRO}}
\newcommand{\aligator}{\textsf{Aligator}}
\newcommand{\lqrax}{\textsf{LQRax}}
\newcommand{\ocs}{\textsf{OCS2}}

\newcommand{\gpops}{\textsf{GPOPS-II}}
\newcommand{\iclocs}{\textsf{ICLOCS2}}
\newcommand{\psopt}{\textsf{PSOPT}}
\newcommand{\maptor}{\textsf{MAPTOR}}
\newcommand{\acados}{\textsf{acados}}

\newcommand{\chomp}{\textsf{CHOMP}}
\newcommand{\stomp}{\textsf{STOMP}}
\newcommand{\gpmp}{\textsf{GPMP2}}
\newcommand{\trajopt}{\textsf{TrajOpt}}
\newcommand{\scptoolbox}{\textsf{SCPToolbox}}

\newcommand{\ipopt}{\textsf{IPOPT}}
\newcommand{\snopt}{\textsf{SNOPT}}

\definecolor{tablebg}{HTML}{E2EBEB} % Light gray-cyan background
\definecolor{tableteal}{HTML}{2B6A7F} % Bold teal for the top line and title

\newcommand{\behcetFull}{Beh\c{c}et~A\c{c}\i kme\c{s}e}

\newcommand{\wProx}[1]{\lambda_{\mathrm{Prox},#1}}
\newcommand{\wVc}{\lambda_{\mathrm{vc}}}

\newcommand{\wVbh}{\lambda_{\mathrm{vb},H}}
\newcommand{\wVbg}{\lambda_{\mathrm{vb},G}}

\definecolor{ttc-red}{RGB}{218, 37, 29}
\definecolor{ttc-yellow}{RGB}{248, 195, 0}
\definecolor{ttc-green}{RGB}{0, 146, 63}
\definecolor{ttc-blue}{RGB}{0, 130, 201}
\definecolor{ttc-purple}{RGB}{162, 26, 104}
\definecolor{ttc-orange}{RGB}{231, 120, 23}
\definecolor{ttc-gray}{RGB}{150, 149, 148}

\colorlet{ttc-red-dark}{ttc-red!85!black}
\colorlet{ttc-red-light}{ttc-red!20}
\colorlet{ttc-yellow-dark}{ttc-yellow!85!black}
\colorlet{ttc-yellow-light}{ttc-yellow!20}
\colorlet{ttc-green-dark}{ttc-green!85!black}
\colorlet{ttc-green-light}{ttc-green!20}
\colorlet{ttc-blue-dark}{ttc-blue!85!black}
\colorlet{ttc-blue-light}{ttc-blue!20}
\colorlet{ttc-gray-dark}{ttc-gray!85!black}
\colorlet{ttc-gray-light}{ttc-gray!20}
\colorlet{ttc-orange-dark}{ttc-orange!85!black}
\colorlet{ttc-orange-light}{ttc-orange!20}

\usepackage{float}
\floatstyle{ruled}
\newfloat{listing}{tbp}{lol}
\floatname{listing}{Listing}

\definecolor{codekw}{HTML}{0B5394}
\definecolor{codestr}{HTML}{B23A2E}
\definecolor{codecomment}{HTML}{5E7A4E}

\lstdefinestyle{oxapi}{
  language=Python,
  basicstyle=\ttfamily\footnotesize,
  keywordstyle=\color{black},
  stringstyle=\color{black},
  commentstyle=\color{black}\itshape,
  showstringspaces=false,
  numbers=none,
  xleftmargin=1.8em,
  aboveskip=0pt, belowskip=0pt, 
  breaklines=true,
  columns=fullflexible,
  keepspaces=true,
  escapeinside={(*@}{@*)},
}

\newcounter{oxsec}

\begin{document}
% Define the Assumption environment
\theoremstyle{definition} % Optional: gives upright text rather than italics
\newtheorem{assumption}{Assumption}
    
\title{\textsc{OpenSCvx}: An Open-Source Modular and Extensible Nonlinear Trajectory Planning Package}

\author{Christopher R. Hayner$^{*,1}$,~\IEEEmembership{Student Member,~IEEE,}
Griffin J. Norris$^{*,2}$, Fabio Spada$^{4}$, Samet Uzun$^{5}$, Avi Mittal$^{1}$, \behcetFull$^{3,4}$,~\IEEEmembership{Fellow,~IEEE}, and Karen Leung$^{1}$~\IEEEmembership{Member,~IEEE}% <-this % stops a space
\thanks{$^*$These authors contributed equally to this work.}%
\thanks{$^1$ Dept. of Aeronautics and Astronautics, University of Washington, Seattle, WA, USA. Corresponding Email: {\tt\small haynec@uw.edu}}
\thanks{$^2$ General Robotics, Seattle, WA, USA.}
\thanks{$^3$ Dept. of Electrical Engineering and Computer Sciences, University of California, Berkeley, Berkley CA, USA.}
\thanks{$^4$ Dept. of Mechanical Engineering, University of California, Berkeley, Berkley CA, USA.}
\thanks{$^5$ Zoox, San Francisco, CA, USA.}
\thanks{The authors would like to acknowledge Davis Adams, Teming Tse, and Haru Tidmore for their patient and gracious help in running experiments on the NASA SENSS Laboratory, as well as the Mitsubishi Electric Research Laboratory for their support and early testing of this work.}
\thanks{This work was supported by a NASA Space Technology Graduate Research Opportunity under grant 80NSSC23K1178.}
}
\markboth{Under Review}%
{Hayner \MakeLowercase{\textit{et al.}}: \textsc{OpenSCvx}: An Open-Source Nonlinear Trajectory Optimization Package}

\maketitle

\begin{abstract}
Trajectory optimization computes dynamically feasible motions that enable autonomous systems to accomplish complex tasks while satisfying operational and environmental constraints. This tutorial presents \textsc{OpenSCvx}, an open-source Python framework that bridges the gap between high-level problem specification and efficient numerical optimization. Rather than requiring users to derive solver-specific mathematical formulations, \textsc{OpenSCvx} provides a symbolic modeling interface that automatically constructs and solves trajectory optimization problems from modular descriptions of objectives, dynamics, and constraints. Beyond simplifying problem formulation, \textsc{OpenSCvx} supports (i) continuous-time constraint modeling, (ii) temporal and logical specifications, (iii) automatic vectorization for scalable and batched optimization, and (iv) a modular architecture that enables new algorithms, models, and solver backends to be incorporated with minimal effort. These capabilities allow researchers and practitioners to rapidly prototype, solve, and extend state-of-the-art trajectory optimization methods.
\end{abstract}

\begin{IEEEkeywords}
Robotics, Trajectory Optimization, Path Planning, Successive Convexification, Open-Source Software, JAX, Autonomous Navigation.
\end{IEEEkeywords}

\section*{Supplementry Material}
Source code: \href{https://github.com/OpenSCvx/OpenSCvx}{https://github.com/OpenSCvx/OpenSCvx}

Documentation: \href{https://openscvx.github.io/OpenSCvx/}{https://openscvx.github.io/OpenSCvx/}

\section{Introduction}

Autonomous systems increasingly operate in environments where motion decisions directly determine mission success, safety, and performance. 
Whether navigating cluttered environments, manipulating objects, or executing precision aerospace maneuvers, these systems must generate trajectories that satisfy nonlinear dynamics and constraints while optimizing task-specific objectives.
Trajectory planning provides the computational foundation for this capability by determining how a system should move through its environment while balancing competing objectives such as efficiency, safety, and robustness.

\begin{figure}
\centering
\includegraphics[width=1\linewidth]{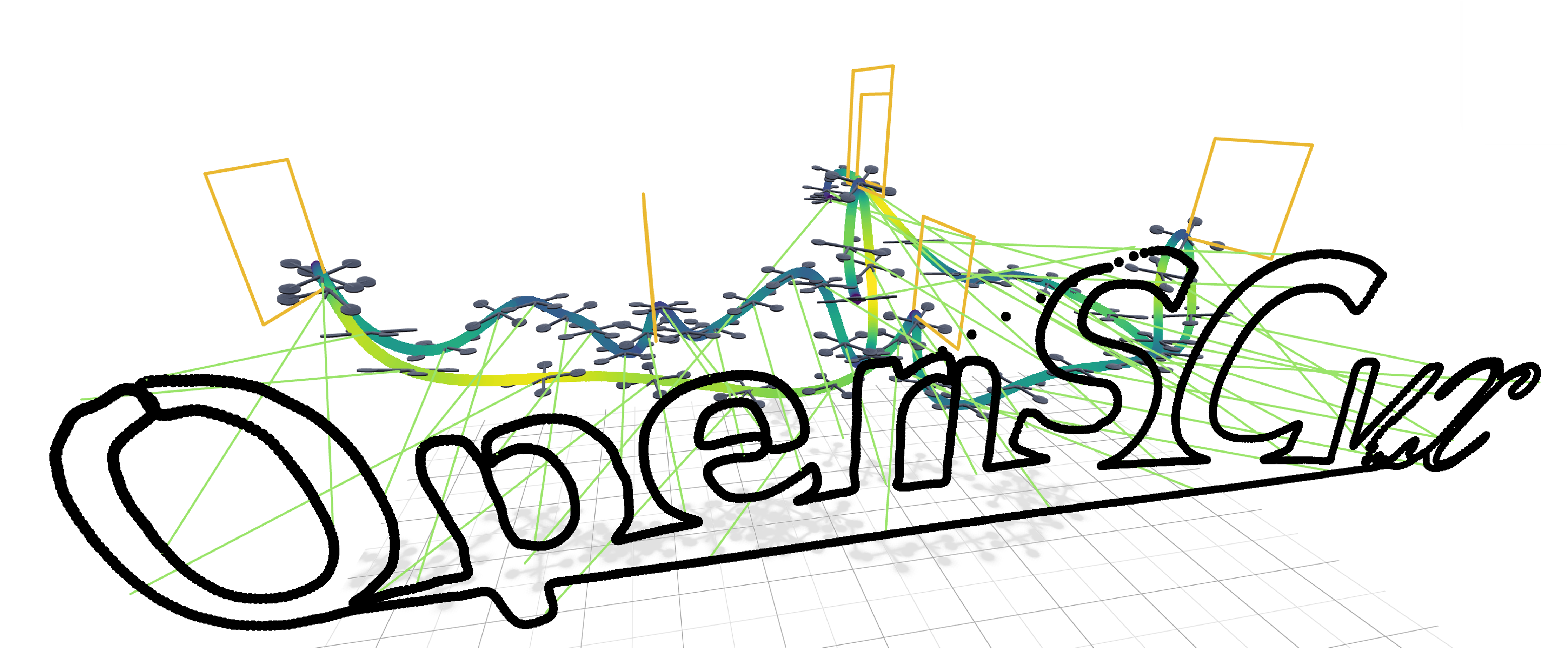}
\caption{Example trajectory generated with \openscvx: a quadrotor flies through a sequence of gates while controlling a laser pointer to trace the \openscvx logo on a target surface.}
\label{fig:main_fig}
\end{figure}

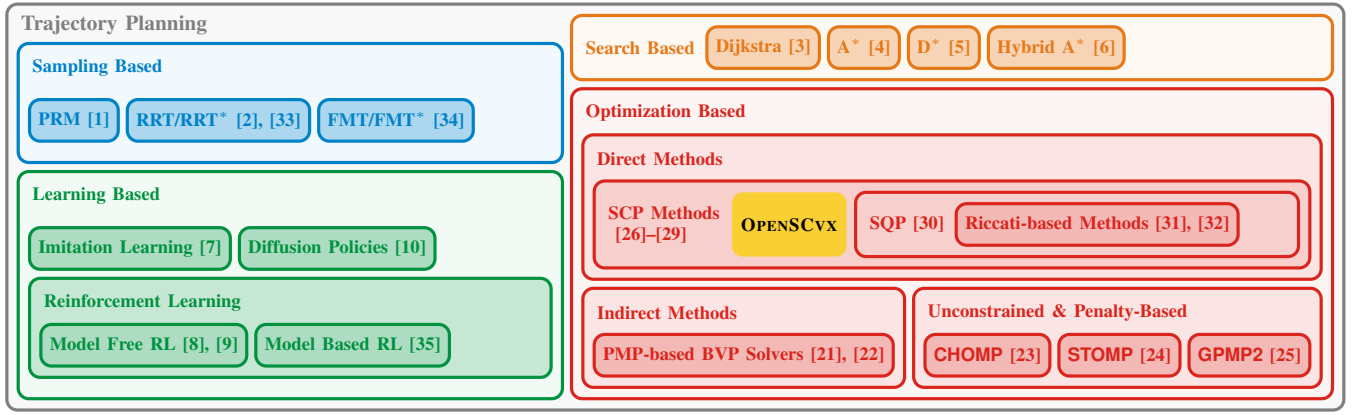
\begin{figure*}[t]
\centering

% Trajectory-planning taxonomy as nested containers: hierarchy is shown by
% containment and tint depth rather than edges.  Containers are fit around
% their children, so each nesting depth paints on its own background layer.
% Spacing: \pad between a border and its content, \gap between siblings.
\pgfdeclarelayer{d1}\pgfdeclarelayer{d2}\pgfdeclarelayer{d3}\pgfdeclarelayer{d4}\pgfdeclarelayer{d5}\pgfdeclarelayer{d6}
\pgfsetlayers{d1,d2,d3,d4,d5,d6,main}

% Layout constants.
\newcommand{\pad}{1.5mm}    % border-to-content clearance, every container, all sides
\newcommand{\gap}{1.2mm}    % gap between siblings
\newcommand{\chiph}{5.5mm}  % uniform chip height
\newcommand{\titleh}{3.4mm} % minimum title height, uniform at every depth
\newcommand{\colw}{72mm}    % width of the left column
% Derived (evaluated by pgfmath):
\newcommand{\thead}{(\pad+\titleh+\gap)}  % top border -> title -> first child
\newcommand{\tdrop}{(\thead+\gap)}        % box bottom -> next sibling's first child
\newcommand{\hdrop}{(2*\pad+\gap)}        % child east  -> next sibling's first child
\newcommand{\boxh}{(\thead+\chiph+\pad)}  % natural height of a one-chip-row container

\begin{tikzpicture}[
    % chip = leaf method, cont = container, ctitle = its title; #1 = family hue.
    chip/.style={
        rounded corners, draw=ttc-#1, fill=ttc-#1!32,
        very thick, minimum height=\chiph, font=\scriptsize\bfseries,
        text=ttc-#1, align=center},
    cont/.style 2 args={
        rounded corners, draw=ttc-#1, fill=ttc-#1!#2,
        very thick, inner sep=\pad},
    ctitle/.style={font=\scriptsize\bfseries, text=ttc-#1,
        inner xsep=0.4mm, inner ysep=0pt, minimum height=\titleh, align=left},
    grp/.style={inner sep=0pt, draw=none},
    outer sep=0pt,  % anchors sit on the stroke: \pad is measured edge to edge
]

% \gbox{name}{hue}{tint}{depth}{title}{fit list}: wrap already-placed nodes in
% a corner-titled container on layer d<depth>.  Phantom coordinates in the fit
% list stretch the box to an exact edge.
\newcommand{\gbox}[6]{%
    \node[grp, fit=#6] (#1-b) {};
    \node[ctitle=#2, anchor=south west] (#1-t) at ([yshift=\gap]#1-b.north west) {#5};
    \begin{pgfonlayer}{d#4}
        \node[cont={#2}{#3}, fit=(#1-b)(#1-t)] (#1) {};
    \end{pgfonlayer}%
}

% \stitle{name}{hue}{title}{at} + \gboxside{name}{hue}{tint}{depth}{fit list}:
% side-title variant for single-row containers.  The title is placed first,
% beside the children, and joins the fit list like any other child.
\newcommand{\stitle}[4]{%
    \node[ctitle=#2, anchor=west] (#1-t) at #4 {#3};%
}
\newcommand{\gboxside}[5]{%
    \begin{pgfonlayer}{d#4}
        \node[cont={#2}{#3}, fit=#5] (#1) {};
    \end{pgfonlayer}%
}

% ===== Optimization Based (red).  Laid out first, at natural height; the rest
% of the figure hangs off it.  The Indirect / Penalty-Based row is the widest
% thing in the figure and sets the width of everything above it.
\node[chip=red, anchor=north west] (pmp) at (0,0)
    {PMP-based BVP Solvers \cite{Pontryagin2018-fn, Bryson2018-sl}};
\gbox{indirect}{red}{12}{3}{Indirect Methods}{(pmp)}

\node[chip=red, anchor=north west] (chompn) at ([xshift={\hdrop}]pmp.north east)
    {\chomp{} \cite{Ratliff2009-cx}};
\node[chip=red, right=\gap of chompn] (stompn) {\stomp{} \cite{Kalakrishnan2011-cc}};
\node[chip=red, right=\gap of stompn] (gpmpn)  {\gpmp{} \cite{Mukadam2018-jl}};
\gbox{unc}{red}{12}{3}{Unconstrained \& Penalty-Based}{(chompn)(stompn)(gpmpn)}

% Direct Methods: SCP > SQP > Structure-Exploiting nested on one side-titled
% band, anchored to the centre line dm-mid.
\coordinate (dm-mid) at
    ([xshift=\pad, yshift={\thead+\gap+3*\pad+\chiph/2}]pmp.north west);
\stitle{scp}{red}{SCP Methods\\%
    \cite{Mao2016-yr, Bonalli2019gusto, Oguri2023alscvx, Uzun2025proxconvex}}{(dm-mid)}
\node[chip=red, anchor=west, fill=ttc-yellow!80, draw=none, text=black,
      minimum height={\chiph+2*\pad}]  % as tall as the SQP box beside it
      (oscvx) at ([xshift=\gap]scp-t.east) {\openscvx};
\stitle{sqp}{red}{SQP \cite{Gill2002snopt}}{([xshift={\gap+\pad}]oscvx.east)}
\node[chip=red, anchor=west] (se) at ([xshift=\gap]sqp-t.east)
    {Riccati-based Methods \cite{David-H-Jacobson1970-ddp, Todorov2005ilqr}};
\path let \p1 = ([xshift={-3*\pad}]unc.east), \p2 = (se.east) in
    coordinate (sqp-e) at ({max(\x1,\x2)}, \y2);  % stretch SQP to the row below
\gboxside{sqp}{red}{22}{5}{(sqp-t)(se)(sqp-e)}
\gboxside{scp}{red}{22}{4}{(scp-t)(oscvx)(sqp)}

\gbox{direct}{red}{12}{3}{Direct Methods}{(scp)}
\gbox{opt}{red}{5}{2}{Optimization Based}{(direct)(indirect)(unc)}

% ===== Search Based (orange): a side-titled row riding \gap above
% Optimization Based, between the same border lines.
\coordinate (rinL) at ([xshift=\pad]opt.west);
\coordinate (rinR) at ([xshift=-\pad]opt.east);
\coordinate (srch-mid) at ([yshift={\gap+\pad+\chiph/2}]opt.north);
\stitle{search}{orange}{Search Based}{(rinL |- srch-mid)}
\node[chip=orange, anchor=west] (dijkstra) at ([xshift=\gap]search-t.east)
    {Dijkstra \cite{Dijkstra1959-ib}};
\node[chip=orange, right=\gap of dijkstra] (astar) {A$^*$ \cite{Hart1968a*}};
\node[chip=orange, right=\gap of astar] (dstar) {D$^*$ \cite{Stentz1994d*}};
\node[chip=orange, right=\gap of dstar] (hybrida)
    {Hybrid A$^*$ \cite{Dmitri2008hybrida*}};
\coordinate (srch-e) at (rinR |- srch-mid);
\gboxside{search}{orange}{5}{2}{(search-t)(dijkstra)(astar)(dstar)(hybrida)(srch-e)}

% ===== Left column: Sampling over Learning, stretched to the right column's
% height.  The top is asymmetric: Search clears the root border by \pad while
% the left column starts below the root title.  colS is the shared border,
% splitting the column's slack evenly between the two boxes.
\coordinate (colR) at ([xshift={-\pad-\gap}]indirect.west);
\coordinate (colL) at ([xshift=-\colw]colR);
\coordinate (inL)  at ([xshift=\pad]colL);
\coordinate (inR)  at ([xshift=-\pad]colR);

\node[font=\footnotesize\bfseries, text=ttc-gray-dark, inner xsep=0.4mm, inner ysep=0pt,
      anchor=north west] (root-t) at (colL |- search.north) {Trajectory Planning};

\coordinate (colT) at ([yshift=-\gap]root-t.south);
\coordinate (colS) at
    ($([yshift={-\boxh}]colT)!0.5!([yshift={2*\boxh+2*\gap}]opt.south)$);

% Sampling Based (blue).
\coordinate (s-top) at ([yshift={-\thead}]colT);
\coordinate (s-bot) at ([yshift=\pad]colS);
\coordinate (s-mid) at ($(s-top)!0.5!(s-bot)$);
\node[chip=blue, anchor=west] (prm) at (inL |- s-mid) {PRM \cite{Kavraki1996prm}};
\node[chip=blue, right=\gap of prm] (rrt) {RRT/RRT$^*$ \cite{lavalle1998rrt, Karaman2011rrt*}};
\node[chip=blue, right=\gap of rrt] (fmt) {FMT/FMT$^*$ \cite{Janson2015-br}};
\coordinate (s-nw) at (inL |- s-top);
\coordinate (s-se) at (inR |- s-bot);
\gbox{samp}{blue}{5}{2}{Sampling Based}{(prm)(rrt)(fmt)(s-nw)(s-se)}

% Learning Based (green), its chip row and RL sub-container centred as a group.
\coordinate (l-top) at ([yshift={-\gap-\thead}]samp.south);
\coordinate (l-bot) at ([yshift=\pad]opt.south);
\coordinate (l-mid) at ($(l-top)!0.5!(l-bot)$);
\node[chip=green, anchor=north west] (imit)
    at ([yshift={(\chiph+\gap+\boxh)/2}]inL |- l-mid)
    {Imitation Learning \cite{Ross2010dagger}};
\node[chip=green, right=\gap of imit] (diff) {Diffusion Policies \cite{Janner2022diffusion}};
\node[chip=green, anchor=north west] (mfrl) at ([xshift=\pad, yshift={-\tdrop}]imit.south west)
    {Model Free RL \cite{Schulman2017ppo, Konda1999actorcritic}};
\node[chip=green, right=\gap of mfrl] (mbrl)
    {Model Based RL \cite{Amos2018-ex}};
\coordinate (rl-e) at ([xshift=-\pad]inR |- mfrl.north);
\gbox{rl}{green}{22}{4}{Reinforcement Learning}{(mfrl)(mbrl)(rl-e)}
\coordinate (l-nw) at (inL |- l-top);
\coordinate (l-se) at (inR |- l-bot);
\gbox{learn}{green}{5}{2}{Learning Based}{(imit)(diff)(rl)(l-nw)(l-se)}

% Root frame; the title is already placed in its corner.
\node[grp, fit=(samp)(search)(learn)(opt)(root-t)] (all) {};
\begin{pgfonlayer}{d1}
    \node[rounded corners, draw=ttc-gray-dark, fill=ttc-gray!7,
          very thick, inner sep=\pad, fit=(all)] {};
\end{pgfonlayer}
\end{tikzpicture}

\caption{A taxonomy of trajectory planning methods. The trajectory optimization categories reflect differences in problem formulation and optimization structure; numerical solvers used to solve intermediate subproblems such as \gls{ipm}'s or active set methods are not shown in this figure.}
\label{fig:trajopt_taxonomy}
\end{figure*}

Representative motion-planning paradigms include search-based, sampling-based, learning-based, and optimization-based methods, as summarized in \cref{fig:trajopt_taxonomy}. These categories are not mutually exclusive, and many modern planning systems combine components from multiple paradigms.
Sampling-based methods, such as \gls{prm} \cite{Kavraki1996prm} and \gls{rrt} \cite{lavalle1998rrt}, construct feasible paths by exploring the state or configuration space through randomized search.
Search-based methods discretize the state or configuration space into a graph and use graph-search algorithms to find minimum-cost paths. 
Classical algorithms such as Dijkstra's algorithm and A* \cite{Dijkstra1959-ib,Hart1968a*} provide foundational approaches, while incremental and replanning methods such as D* \cite{Stentz1994d*} extend search to environments whose costs may change. 
More specialized approaches, such as Hybrid A* \cite{Dmitri2008hybrida*}, incorporate continuous vehicle states or motion primitives into the search process to better account for vehicle dynamics and nonholonomic constraints.
State-lattice and motion-primitive approaches can encode differential constraints into the graph construction, yielding dynamically feasible motions by design while retaining the computational advantages of heuristic search.

Learning-based methods instead leverage demonstrations, simulation data, or interaction to learn policies or trajectory generators. 
Approaches including imitation learning \cite{Ross2010dagger}, reinforcement learning \cite{Schulman2017ppo,Konda1999actorcritic}, and generative trajectory models \cite{Janner2022diffusion, kazukidiffusion} can provide extremely fast online inference and handle high-dimensional sensory inputs.
However, these methods generally require representative training distributions and may provide limited guarantees on constraint satisfaction, safety, or behavior outside the training environment.

Trajectory optimization provides a complementary approach by formulating motion planning as a constrained \gls{ocp}, where trajectories are computed by optimizing continuous-time state and control trajectories subject to system dynamics, path constraints, boundary conditions, and performance objectives \cite{Betts2010-vx}. 
Rather than discretizing the planning space into a graph, exploring it through random sampling, or learning a trajectory from data, trajectory optimization obtains solutions by numerically solving the resulting constrained optimization problem.
This formulation provides a systematic framework for incorporating physical models, actuator limits, environmental constraints, and mission-specific objectives, making trajectory optimization a fundamental tool in aerospace guidance, robotics, and autonomous systems. 

Most modern direct trajectory-optimization methods transform an infinite-dimensional continuous-time optimal control problem into a finite-dimensional numerical optimization problem.
This process involves several distinct computational layers that are often conflated in both literature and software implementations. 
First, the continuous-time optimal control problem must be formulated through system dynamics, objectives, and constraints. 
The resulting problem must then be parameterized through a transcription method, such as single shooting \cite[Section 3.3]{Betts2010-vx} \cite{Hicks1971single}, multiple shooting \cite[Section 3.4]{Betts2010-vx} \cite{Bock1984multiple, Diehl2007-qf}, direct collocation \cite{Hargraves1987-bk, Kelly2017-zv}, or pseudospectral discretization \cite{Elnagar1995-fs, Ross2004pseudospectral, Garg2010-yu}.
A finite-dimensional nonlinear program is then formed and iteratively solved until convergence.
Because the resulting problems are generally nonconvex, numerical trajectory-optimization methods typically seek locally optimal or stationary solutions, and their performance can depend on initialization and globalization mechanisms.

The methods in \cref{fig:trajopt_taxonomy} are complemented by additional software layers, including symbolic modeling systems, automatic differentiation frameworks, numerical optimization backends, and execution strategies such as parameter caching, batching, and hardware acceleration.
Although each component has been studied extensively, integrating these layers into a flexible and reusable trajectory optimization framework remains challenging.
Over the past several decades, numerous trajectory optimization frameworks have emerged that address different portions of this computational pipeline.
Direct optimal control frameworks such as \gpops{} \cite{Patterson2014GPOPII}, \iclocs{} \cite{Nie2018ICLOC2}, \psopt{} \cite{Becerra2010PSOPT}, and \maptor{} \cite{maptor2025} provide high-level formulations of continuous-time optimal control problems and automatically perform transcription into sparse nonlinear programs solved using mature \gls{nlp} algorithms.
Other frameworks, such as \acados{} \cite{Verschueren2022acados}, exploit problem structure through efficient \gls{sqp} methods and code generation to enable high-performance embedded optimal control.
\gls{ddp} \cite{David-H-Jacobson1970-ddp} and \gls{ilqr}-based methods \cite{Todorov2005ilqr, Kavuncu2021-dpilqr, Amatucci2026-rr} exploit the stagewise structure of discrete-time optimal-control problems through backward dynamic-programming or Riccati-style recursions, often yielding low per-iteration computational cost. Frameworks such as \altro{} \cite{Howell2019-we}, \crocoddyl{} \cite{Mastalli2020-on}, and \ocs{} extend these ideas to constrained and robotics-oriented problems using augmented-Lagrangian, barrier, and feasibility-restoration mechanisms.

Despite these advances, software support for \gls{scvx} remains comparatively limited.
\Gls{scvx} \cite{Mao2016-yr,Mao2017-zj} has emerged as a powerful trajectory optimization paradigm that transforms difficult nonlinear optimal control problems into sequences of convex optimization problems through iterative convexification.
Convex optimization problems are a very well-studied class of optimization problems that leverage convexity within the costs and constraints to efficiently solve them with guarantees that every local minimizer of a feasible convex problem is globally optimal for that problem \cite{Boyd2004cvx}.
Unlike general \gls{nlp} methods, \gls{scvx} can preserve convex structure within the original problem while providing reliable convergence behavior for challenging aerospace and robotic applications.
Recent extensions, including Guaranteed Sequential Trajectory Optimization (GuSTO) \cite{Bonalli2019gusto} and \gls{ctscvx} \cite{Elango2024ctscvx,Elango2025auto}, have further expanded the applicability of these methods.

However, existing \gls{scvx} implementations are generally research-oriented and application-specific.
They often require users to manually construct problem transcriptions, provide derivatives, implement algorithmic components, or modify solver internals when extending the method to new applications.
This is a particularly high barrier of entry for practitioners who wish to prototype trajectory planning algorithms across a wide range of problems.
Consequently, researchers frequently develop independent implementations rather than building upon a common, extensible software framework.

In this work, we introduce \openscvx, an open-source general-purpose trajectory optimization framework built around continuous-time successive convexification.
The objective of \openscvx is to provide a general software environment for formulating, solving, and extending nonlinear optimal control problems across robotics, aerospace, and autonomous systems.

\openscvx combines a high-level symbolic optimal control modeling interface with a modular \gls{ctscvx} optimization architecture and a high-performance computational backend.
The framework separates problem specification from algorithmic implementation, enabling users to define nonlinear dynamics, objectives, and constraints while independently modifying transcription strategies, convexification methods, update rules, and numerical backends.
By leveraging \jax, \openscvx additionally supports automatic differentiation, vectorized computation, parameterized problem updates, and GPU acceleration for scalable trajectory optimization.

\subsection*{Contributions}

The primary contributions of \openscvx are:

\begin{itemize}

\item[\textbf{C.1}]
A symbolic modeling framework for nonlinear optimal control that enables users to specify dynamics, objectives, and constraints through composable mathematical expressions. 
The framework automatically constructs the computational representations required for trajectory optimization, reducing implementation effort when developing new problems and algorithms.

\item[\textbf{C.2}]
A modular \gls{ctscvx} framework supporting interchangeable transcription methods, convexification strategies, and algorithmic update schemes.
This architecture enables systematic development and evaluation of \gls{scvx} variants across a broad range of trajectory optimization problems.

\item[\textbf{C.3}]
A parameterized optimization architecture that separates problem structure from numerical parameters, enabling efficient repeated optimization with updated problem data without requiring complete reconstruction of the optimization model.

\item[\textbf{C.4}]
A \jax-based computational backend that leverages automatic differentiation, vectorized execution, and GPU acceleration to enable scalable trajectory optimization, including batched optimization and integration with learned or simulation-based dynamics models.

\end{itemize}

\subsection*{How to follow this tutorial}

This tutorial presents the complete workflow of formulating and solving trajectory optimization problems using \openscvx.
\Cref{sec:related_work} first reviews trajectory optimization algorithms and software frameworks, emphasizing the distinction between optimal control formulations, transcription methods, optimization algorithms, and computational backends.
\Cref{sec:nocp} introduces the nonlinear optimal control problem class supported by \openscvx.
\Cref{sec:modeling-interface} presents the symbolic modeling framework and demonstrates how high-level mathematical descriptions are translated into optimization problems.
\Cref{sec:openscvx-framework} describes the \gls{ctscvx} algorithm and discretization strategies.
Finally, \cref{sec:case_study} evaluates \openscvx through verification problems, benchmark comparisons, and representative robotics and aerospace applications.

\medskip
\noindent\begin{minipage}{0.70\linewidth}
To follow along, \openscvx can be installed with \texttt{uv pip install openscvx}; the source code and documentation are available at \href{https://github.com/OpenSCvx/OpenSCvx}{github.com/OpenSCvx/OpenSCvx} and \href{https://openscvx.github.io/OpenSCvx/}{openscvx.github.io/OpenSCvx}, respectively.
\end{minipage}\hfill
\begin{minipage}{0.26\linewidth}
\centering
\qrcode[height=0.8in]{https://github.com/OpenSCvx/OpenSCvx}
\end{minipage}

\section{Trajectory Optimization Software and Algorithmic Paradigms}
\label{sec:related_work}

Trajectory optimization approaches can broadly be categorized according to how they exploit the structure of the optimal control problem. 
Classical indirect methods solve the necessary conditions of optimality derived from Pontryagin's maximum principle \cite{Pontryagin2018-fn, Bryson2018-sl}, while modern software frameworks have largely focused on direct transcription, structure-exploiting Bellman-based methods, and \gls{scp} approaches.
These paradigms provide different tradeoffs between computational efficiency, modeling flexibility, and the ability to represent complex optimal control problems.

Existing software ecosystems broadly fall into five categories.
General-purpose optimization modeling frameworks provide symbolic abstractions for constructing optimization problems but require users to manually formulate the trajectory optimization transcription.
Structure-exploiting Bellman-based methods exploit the structure of optimal control problems to obtain highly efficient algorithms, but typically require stronger assumptions on the problem formulation.
Direct transcription frameworks automatically convert continuous-time optimal control problems into \glspl{nlp} that are solved using mature nonlinear optimization algorithms.
Finally, \gls{scp} methods solve a sequence of convex approximations, preserving convex problem structure while handling nonlinear dynamics and constraints.

\Cref{tab:trajopt_ecosystem} summarizes representative frameworks spanning these categories and compares their transcription methods, solver paradigms, modeling requirements, computational backends, and primary application domains.
This taxonomy highlights both the diversity of existing approaches and the design space that motivates \openscvx.

\begin{table*}[t]
\centering
\caption{
Representative trajectory optimization software and algorithmic paradigms. 
\openscvx\ combines the high-level problem specification of modern trajectory optimization frameworks with the structure-exploiting advantages of continuous-time successive convexification. 
Objective capability features replace subjective ratings: 
\emph{Auto Trans.} indicates automatic continuous-to-discrete transcription; 
\emph{Auto Diff} indicates automatic differentiation without manual Jacobians; 
\emph{CTCS} indicates continuous-time path constraint satisfaction; 
\emph{STL / Logic} indicates native support for Signal Temporal Logic or state-triggered constraints; 
\emph{Mesh Adapt.} indicates adaptive mesh refinement; 
\emph{JIT / Compile} indicates JIT/AOT compilation or C-code generation; 
and \emph{Batch / GPU} indicates native vectorized batching or GPU acceleration.
}
\label{tab:trajopt_ecosystem}

\footnotesize
\setlength{\tabcolsep}{2.5pt}
\renewcommand{\arraystretch}{1.2}

\rowcolors{2}{gray!12}{white}

\begin{tabularx}{\textwidth}{
l                                                   % Framework
>{\raggedright\arraybackslash}X                     % Transcription
>{\centering\arraybackslash}p{0.8cm}                % Auto Trans.
>{\centering\arraybackslash}p{0.8cm}                % Auto Diff
>{\centering\arraybackslash}p{0.7cm}                % CTCS
>{\centering\arraybackslash}p{0.8cm}                % STL / Logic
>{\centering\arraybackslash}p{0.8cm}                % Mesh Adapt.
>{\centering\arraybackslash}p{0.9cm}                % JIT / Compile
>{\centering\arraybackslash}p{0.8cm}                % Batch / GPU
l                                                   % Language
}

\toprule
\rowcolor{gray!25}
\textbf{Framework} &
\textbf{Transcription} &
\textbf{\shortstack{Auto\\Trans.}} &
\textbf{\shortstack{Auto\\Diff}} &
\textbf{\shortstack{CTCS}} &
\textbf{\shortstack{STL /\\Logic}} &
\textbf{\shortstack{Mesh\\Adapt.}} &
\textbf{\shortstack{JIT /\\Compile}} &
\textbf{\shortstack{Batch /\\GPU}} &
\textbf{Language} \\
\midrule

\rowcolor{gray!18}
\multicolumn{10}{l}{\textbf{General Modeling Frameworks}}\\
\midrule

\casadi{} \cite{andersson2018casadi} &
User-defined &
-- &
\checkmark &
-- &
-- &
-- &
\checkmark &
-- &
C++ / Python / MATLAB \\

\cusadi{} \cite{Jeon2025-iw} &
User-defined &
-- &
\checkmark &
-- &
-- &
-- &
\checkmark &
\checkmark &
C++ / Python \\

\drake{} \cite{drake} &
Collocation &
\checkmark &
\checkmark &
-- &
-- &
-- &
\checkmark &
-- &
C++ / Python \\

\cvxpy{} \cite{diamond2016cvxpy}/ DNLP \cite{Cederberg2026dnlp} &
User-defined &
-- &
\checkmark &
-- &
-- &
-- &
-- &
-- &
Python \\

\jump{} \cite{Dunning2015-ut} & 
User-defined &
-- &
\checkmark &
-- &
-- &
-- &
\checkmark &
-- &
Julia \\

\midrule
\rowcolor{gray!18}
\multicolumn{10}{l}{\textbf{Direct Methods}}\\
\midrule

\gpops{} \cite{Patterson2014GPOPII, Agamawi2020CGPOP} &
LGR Collocation &
\checkmark &
\checkmark &
-- &
-- &
\checkmark &
\checkmark &
-- &
MATLAB / C \\

\maptor{} \cite{maptor2025} &
LGR Collocation &
\checkmark &
\checkmark &
-- &
-- &
\checkmark &
-- &
-- &
Python \\

\iclocs{} \cite{Nie2018ICLOC2} &
Multiple Shooting / Collocation &
\checkmark &
\checkmark &
-- &
-- &
\checkmark &
-- &
-- &
MATLAB \\

\psopt{} \cite{Becerra2010PSOPT} &
LGR Collocation &
\checkmark &
\checkmark &
-- &
-- &
\checkmark &
\checkmark &
-- &
C++ \\

\acados{} \cite{Verschueren2022acados} &
Multiple Shooting &
\checkmark &
\checkmark &
-- &
-- &
-- &
\checkmark &
-- &
C / Python / MATLAB \\

\midrule
\rowcolor{gray!18}
\multicolumn{10}{l}{\textbf{Riccati-based QP Methods}}\\
\midrule

\crocoddyl{} \cite{Mastalli2020-on} &
Multiple Shooting &
\checkmark &
\checkmark &
-- &
-- &
-- &
\checkmark &
-- &
C++ / Python \\

\altro{} \cite{Howell2019-we} &
Multiple Shooting &
\checkmark &
\checkmark &
-- &
-- &
-- &
\checkmark &
-- &
Julia \\

\aligator{} \cite{Jallet2025-lg} &
Multiple Shooting &
\checkmark &
\checkmark &
-- &
-- &
-- &
\checkmark &
-- &
C++ / Python \\

\textsf{DP-iLQR} \cite{Kavuncu2021-dpilqr} &
Single Shooting &
\checkmark &
\checkmark &
-- &
-- &
-- &
-- &
-- &
Python \\

\lqrax{} \cite{sun_lqrax_2025} &
Single Shooting &
\checkmark &
\checkmark &
-- &
-- &
-- &
\checkmark &
\checkmark &
Python \\

\midrule
\rowcolor{gray!18}
\multicolumn{10}{l}{\textbf{Unconstrained \& Penalty-based Methods}}\\
\midrule

\chomp{} \cite{Ratliff2009-cx} &
Discrete &
-- &
-- &
-- &
-- &
-- &
\checkmark &
-- &
C++ \\

\stomp{} \cite{Kalakrishnan2011-cc} &
Discrete &
-- &
-- &
-- &
-- &
-- &
\checkmark &
-- &
C++ / ROS \\

\gpmp{} \cite{Mukadam2018-jl} &
GP Interpolation &
\checkmark &
-- &
-- &
-- &
-- &
\checkmark &
-- &
C++ \\

\midrule
\rowcolor{gray!18}
\multicolumn{10}{l}{\textbf{SCP Methods}}\\
\midrule

\trajopt{} \cite{Schulman2014-cm} &
Discrete &
-- &
-- &
-- &
-- &
-- &
\checkmark &
-- &
C++ \\

\scptoolbox{} \cite{SCPToolboxCSM2022} &
Multiple Shooting &
\checkmark &
-- &
-- &
-- &
-- &
\checkmark &
-- &
Julia \\

\textsc{SCvxGEN}\textsuperscript{a} \cite[Chapter 7]{Kamath2025thesis} &
Multiple Shooting &
\textbf{\checkmark} &
\textbf{\checkmark} &
\textbf{\checkmark} &
\textbf{\checkmark} &
\textbf{--} &
\textbf{\checkmark} &
-- &
Python/C++ \\

\textbf{\openscvx} &
\textbf{Multiple Shooting} &
\textbf{\checkmark} &
\textbf{\checkmark} &
\textbf{\checkmark} &
\textbf{\checkmark} &
\textbf{--} &
\textbf{\checkmark} &
\textbf{\checkmark} &
\textbf{Python} \\

\bottomrule
\end{tabularx}
\vspace{2pt}
{\raggedright\footnotesize \textsuperscript{a}Not yet publicly released.\par}
\end{table*}

\subsection{General modeling frameworks}

The development of high-level optimization modeling frameworks has significantly simplified the formulation of mathematical programs by separating problem specification from numerical implementation. 
Rather than requiring users to construct solver-specific matrices, these frameworks allow optimization problems to be expressed using symbolic variables, objectives, and constraints, while automatically generating the corresponding numerical representations.

A foundational example is \cvxpy{} \cite{diamond2016cvxpy}, which introduced \gls{dcp}, enabling users to describe convex optimization problems using mathematical expressions while automatically verifying convexity and canonicalizing the resulting problem. 
Similar modeling abstractions have been adopted in frameworks such as \jump{} \cite{Dunning2015-ut} and extended to nonlinear optimization through symbolic systems such as \casadi{} \cite{andersson2018casadi}. 
Other frameworks, including \drake{} \cite{drake}, focus on high-fidelity modeling of complex dynamical systems, while recent work on \gls{dnlp} \cite{Cederberg2026dnlp} extends structured modeling principles beyond convex  within the \cvxpy{} framework.

These frameworks provide powerful abstractions for constructing optimization problems; however, they remain general-purpose tools rather than trajectory optimization frameworks. 
For nonlinear optimal control, users must still manually select a transcription method, discretize continuous-time dynamics, formulate the resulting finite-dimensional program, and manage numerical considerations associated with solving the resulting nonlinear optimization problem. 
Consequently, while modern modeling frameworks have simplified optimization problem specification, much of the trajectory optimization workflow remains the responsibility of the user.

The success of these abstractions motivates a similar separation between optimal control problem specification and numerical solution. 
However, achieving this separation for trajectory optimization requires additional capabilities to automatically handle dynamic constraints, trajectory transcription, and the iterative solution of nonconvex optimal control problems.

\subsection{Direct trajectory optimization frameworks}

Direct methods instead discretize the continuous-time optimal control problem and transcribe it into a finite-dimensional \gls{nlp}.
The resulting optimization problem is then solved using mature nonlinear optimization algorithms such as \ipopt{} \cite{Wachter2006ipopt}, \snopt{} \cite{Gill2002snopt}, or \textsf{KNITRO} \cite{Byrd2006knitro}.
This paradigm forms the foundation of many of the most widely used trajectory optimization frameworks.

Frameworks including \gpops{} \cite{Patterson2014GPOPII,Agamawi2020CGPOP}, \iclocs{} \cite{Nie2018ICLOC2}, \psopt{} \cite{Becerra2010PSOPT}, \maptor{} \cite{maptor2025}, and \acados{} \cite{Verschueren2022acados} provide high-level interfaces that automatically perform transcription while allowing users to specify continuous-time dynamics, objectives, and constraints.
These frameworks have demonstrated remarkable success across aerospace guidance, robotics, and embedded \gls{mpc}.

Despite their maturity, direct transcription fundamentally replaces the original continuous-time optimal control problem with a discrete approximation.
As a consequence, state and path constraints are typically enforced only at discretization nodes, requiring increasingly dense grids or mesh refinement techniques \cite{kumagai2024adaptive_mesh_scp_space} to approximate continuous-time feasibility.
Likewise, mission specifications that naturally evolve over continuous trajectories, such as visibility requirements, collision avoidance certificates, or \gls{stl} specifications, often require problem-specific reformulations or additional discretization heuristics.
Furthermore, transcription into a general nonlinear program discards much of the convex structure inherent to many optimal control problems, preventing modern convex optimization techniques from being directly exploited.

\subsection{Functional gradient, stochastic, and probabilistic trajectory optimization}

A complementary class of trajectory optimization methods avoids formulating the problem as a general nonlinear program or sequence of convex subproblems. 
Instead, these approaches iteratively refine trajectories using functional gradients, stochastic sampling, or probabilistic inference, often exploiting problem-specific structure for efficient continuous-time motion planning.

Covariant Hamiltonian Optimization for Motion Planning (\chomp) \cite{Ratliff2009-cx} performs trajectory optimization using functional gradient descent in a covariant metric that promotes smoothness while avoiding obstacles. 
Obstacle costs are evaluated from workspace distance fields, enabling continuous trajectory refinement from simple initial guesses.

Stochastic Trajectory Optimization for Motion Planning (\stomp) \cite{Kalakrishnan2011-cc} replaces gradient computations with stochastic trajectory perturbations. 
Candidate rollouts are sampled, evaluated, and combined to update the nominal trajectory, allowing optimization of non-differentiable objectives and constraints.

Gaussian Process Motion Planning (\textsf{GPMP} and \gpmp) \cite{Mukadam2018-jl} formulates trajectory optimization as probabilistic inference, representing trajectories as Gaussian processes with sparse Gauss--Markov priors. 
The resulting maximum a posteriori estimation problem is solved using structure-exploiting nonlinear least-squares methods, while Gaussian process interpolation enables efficient continuous-time collision checking and online replanning.

These methods have proven highly effective for robotic motion planning, particularly in high-dimensional configuration spaces. 
However, they are primarily tailored to motion planning problems rather than general nonlinear optimal control, making them complementary to direct transcription and \gls{scvx} approaches.

\subsection{Sequential convex programming}

\Gls{scp} provides an alternative direct optimization methodology that solves a nonconvex optimization problem through a sequence of convex subproblems. At each iteration, nonlinear dynamics and nonconvex constraints are locally approximated to obtain a convex optimization problem that can be solved using mature convex optimization techniques. Unlike general \gls{nlp} methods, \gls{scp} can preserve convex structure already present in the original formulation, allowing convex objectives and nonlinear convex constraints to remain in the subproblem rather than being replaced by local quadratic approximations.

\paragraph{Sequential quadratic programming} A useful specialization within this framework is \gls{sqp}, in which each local model is restricted to a \gls{qp}, typically through a quadratic approximation of the objective and linearization of the constraints. When the quadratic model is convex, for example under a positive-semidefinite Hessian or Hessian approximation, \gls{sqp} can be viewed as a \gls{qp}-specialized subset of \gls{scp}: both methods sequentially construct tractable local optimization problems, but general \gls{scp} permits convex subproblems beyond \gls{qp}s. This distinction allows \gls{scp} methods to preserve convex nonlinear structure, including second-order cone constraints, that would otherwise need to be approximated within a standard \gls{sqp} formulation. The resulting \gls{qp}s in \gls{sqp} can be solved using closed-form methods for specialized problem structures or numerical solvers such as \gls{ipm}s.

\paragraph{Riccati-based methods} The additional stagewise structure present in optimal control problems can sometimes be exploited through Riccati-based solution procedures that decompose quadratic subproblems into smaller sequential operations. More broadly, however, \gls{ddp} \cite{David-H-Jacobson1970-ddp}, \gls{ilqr} \cite{Todorov2005ilqr}, and \gls{ilqg} \cite{Todorov2005ilqr} are better viewed as a separate structure-exploiting trajectory-optimization family: they construct local approximations of the dynamics, cost, and value function and compute updates through efficient backward-forward recursions closely related to discrete-time Riccati methods. These algorithms are therefore connected to, but not merely subroutines for, \gls{sqp}. While generic \gls{sqp} approaches may pass a convex \gls{qp} to a general-purpose numerical solver, Riccati-based methods instead exploit the temporal structure of the optimal-control problem directly to reduce computational cost and latency.

\paragraph{General \Gls{scp} Solvers} Within the general \gls{scp} methodology, several algorithms have been developed specifically for trajectory optimization. Early work such as \trajopt{} \cite{Schulman2014-cm} demonstrated the effectiveness of \gls{scp} for robotic motion planning, but was restricted to kinematics. 
Subsequent developments include SCvx* \cite{Oguri2023alscvx}, which embeds SCvx within an augmented-Lagrangian framework to strengthen feasibility guarantees; AutoSCvx \cite{Mceowen2026autoscvx}, which introduces automated parameter adaptation; and CT-SCvx \cite{Elango2025auto}, which addresses continuous-time constraint satisfaction.
These methods instantiate the general \gls{scp} methodology in specialized trajectory-optimization algorithms, with different choices for convexification, constraint handling, globalization, and convergence mechanisms. They are more general than \gls{sqp} in that they can preserve the convex structure present in the original \gls{ocp}, such as \gls{soc} or semidefinite program constraints, provided an appropriate convex solver is used.

The convergence-guaranteed prox-linear framework \cite{Drusvyatskiy2019-sg} provides an important theoretical foundation for this class of sequential convex methods and is particularly relevant to trajectory optimization because it provides conditions under which successive convexification of a nonconvex problem converges to stationary solutions. More generally, prox-linear and its generalized prox-convex formulations \cite{Uzun2025proxconvex} provide a principled framework for constructing and analyzing convex local models, providing important reliability and convergence insights for \gls{scp}-based trajectory optimization.

These advances have enabled \gls{scp} to address increasingly challenging applications including planetary landing, spacecraft guidance, perception-aware planning, and information-aware trajectory optimization \cite{uzun2025information_acquisition_ctscvx,spada2025impulsive_relative_ctcs,uzun2025sequential,kumagai2024adaptive_mesh_scp_space}.

Despite the computational efficiency of structure-exploiting methods, their performance relies on assumptions about problem structure and local approximations. Constraints are typically enforced along discretized trajectories, making continuous-time feasibility between nodes challenging. Additionally, local linearization or quadratic approximation of all problem components can prevent direct exploitation of existing convex structure within the formulation. Free-final-time, nonsmooth, and complex path-constrained formulations often require additional reformulation or specialized algorithmic mechanisms.

Similarly, despite the theoretical and algorithmic advances in \gls{scp}, software support for these methods remains comparatively immature. Existing implementations, such as \scptoolbox{} \cite{SCPToolboxCSM2022}, largely expose algorithm-level interfaces requiring users to manually construct transcriptions, provide derivative information, or formulate convex subproblems directly. Consequently, many of the software engineering advances that have made direct \gls{nlp} frameworks broadly accessible—including symbolic problem specification, automatic differentiation, backend abstraction, and modular solver interfaces—have yet to be fully realized within the \gls{scp} ecosystem.

\subsection{\openscvx}

\openscvx is designed to bridge this gap by bringing modern optimization software abstractions to continuous-time successive convexification.
Rather than exposing algorithmic implementation details, \openscvx allows users to formulate continuous-time optimal control problems using a high-level symbolic modeling interface that closely resembles modern optimization frameworks while automatically generating the convex subproblems required by \gls{ctscvx}.

\openscvx distinguishes itself through the integration of continuous-time constraint handling, \gls{stl}-based mission specifications, a composable symbolic interface, modular CT-SCvx components, and \jax-accelerated computation within a common Python framework.
It also provides automatic differentiation, modular algorithm components, and interchangeable computational backends.
\jax provides automatic differentiation, vectorization, compilation, and accelerator execution. 
Within the \jax ecosystem, we make use of \textsf{Diffrax} for integration.
Parameterized problem updates and DPP caching come from \cvxpy, while customized C-code generation comes from CVXPYgen
Combined with an extensive benchmark suite spanning more than one hundred optimal control problems, \openscvx aims to provide a general-purpose software platform for continuous-time trajectory optimization across robotics, aerospace, and autonomous systems.
\section{Supported Problem Class}
\label{sec:nocp}

Before describing the modeling language, we first define the class of \glspl{ocp} supported by \openscvx. The symbolic interface introduced in \cref{sec:modeling-interface} is ultimately a mechanism for constructing instances of the following continuous-time \gls{ocp}.

The primary objective of \openscvx is to solve generic nonlinear \glspl{ocp} posed in the continuous-time Mayer form. From a user's perspective, utilizing the package requires mapping their physical system and mission objectives directly into this standardized mathematical structure:

\begin{mathnote}{Mayer Form}
\begin{subequations}
\label{prob:mayer}
\begin{align}
\min_{x, u,\mu, t} \quad & J(t_f, x(t_f)) \label{eqn:mayer:cost}\\
\text{s.t.} \quad & \dot{x}(t) = f(t, x(t), u(t)), \quad t \in [t_0, t_f] \\
& x(t^+) = f_\textrm{d}(t^-, x(t^-), \mu(t^-)), \quad t \in \mathcal{T} \\
& g(t, x(t), u(t)) \leq 0_{n_g}, \quad t \in [t_0, t_f] \\
& h(t, x(t), u(t)) = 0_{n_h}, \quad t \in [t_0, t_f] \\
& P(t_0, x(t_0), t_f, x(t_f)) \leq 0_{n_P} \\
& Q(t_0, x(t_0), t_f, x(t_f)) = 0_{n_Q}
\end{align}
\end{subequations}
where $x(t) \in \mathbb{R}^{n_x}$ is the state vector, $u(t) \in \mathbb{R}^{n_u}$ is the control vector, $\mu(t) \in \mathbb{R}^{n_\mu}$ is the impulsive control vector and $t$ is the time. The functions $f$, $f_\textrm{d}$ define respectively the continuous-time and the discrete-time dynamics of the system, $\mathcal{T}$ gathers the discrete dynamics timings, $g$ and $h$ define general path inequality and equality constraints, and $P$ and $Q$ enforce the initial and terminal boundary conditions. 
\end{mathnote}

To ensure mathematical well-posedness and exact sensitivity propagation, we assume the continuous control inputs $u(t)$ are piecewise continuous, while discrete impulsive controls $\mu_k$ act instantaneously at specified event times $\mathcal{T} = \{t_1^\mathcal{T}, \dots, t_M^\mathcal{T}\} \subset [t_0, t_f]$ (e.g., impulsive orbital $\Delta v$ maneuvers). Consequently, the state trajectory $x(t)$ is piecewise continuous, permitting finite jump discontinuities across $\mathcal{T}$.

\begin{assumption}[Function Smoothness]
The continuous dynamics vector field $f(t, x, u)$, discrete transition map $f_\textrm{d}(t, x, \mu)$, path constraints $g(t, x, u), h(t, x, u)$, boundary conditions $P, Q$, and terminal cost $J$ are continuously differentiable ($C^1$) with respect to their time, state and control arguments $(t, x, u, \mu)$, possessing locally Lipschitz continuous Jacobians.
\end{assumption}

Under Assumption~1, forward sensitivity Jacobians exist everywhere along the trajectory, allowing \jax{} automatic differentiation to generate exact subproblem linearizations across both continuous intervals and discrete state jumps.

Crucially, no assumptions of convexity are made for any functions within the Mayer form. 
Because \openscvx\ handles generic nonlinearities directly, this formulation naturally accommodates a wide class of challenging trajectory optimization scenarios without requiring manual analytical simplifications. 
Furthermore, this continuous-time formulation allows for enforcement
of non-convex spatial, logical, and physical constraints, and
temporal specifications described in \cref{par:stl}.
To define $f$, $f_\textrm{d}$, $g$, $h$, $P$, and $Q$, the user utilizes the provided atomic operators within the \openscvx\ interface outlined in \cref{apx:atoms}. 

As \openscvx\ targets Mayer form, which minimizes only a terminal cost, running costs must be recast through state augmentation. For a running cost $\int_{t_0}^{t_f} J_r(t, x(t), u(t)) \, dt$, introduce a state $\xi$ with dynamics $\dot{\xi} = J_r(t, x(t), u(t))$ and initial condition $\xi(t_0) = 0$. The running cost is then minimized as the terminal Mayer cost $\xi(t_f)$. 
\section{Building an Optimal Control Problem with \openscvx}
\label{sec:modeling-interface}

Inspired by modeling frameworks such as \cvxpy{} \cite{diamond2016cvxpy, agrawal2018rewriting} and \casadi{} \cite{andersson2018casadi}
as well as symbolic and numerical computing libraries such as \textsf{SymPy} \cite{meuer2017sympy} and \jax{} \cite{jax2018github},
\openscvx provides a symbolic expression layer with which users can define their problems.
A guiding principle during development was to keep the programmatic interface as close to the mathematical notation of \cref{prob:mayer} as practical, while simultaneously remaining familiar for users of the scientific Python ecosystem,
adopting the problem-construction idiom of \cvxpy, symbolic operators in the style of \textsf{SymPy}, and \textsf{NumPy}-style \cite{Harris2020numpy} array semantics.

The remainder of this section mirrors the user's workflow.
\Cref{sec:symbolic-expression-layer} introduces the symbolic expression system, an embedded \gls{dsl} on which the modeling interface is built.
\Cref{sec:tutorial} then follows a guided tutorial, implementing the Brachistochrone problem from its mathematical statement through problem assembly and solution.
\Cref{sec:beyond-tutorial} presents modeling constructs that reach beyond this example, and \cref{sec:working-with-solutions} shows how solutions are accessed and updated between solves using the same symbolic vocabulary that defined the problem.
Throughout, we restrict attention to what the user writes; how \openscvx solves the assembled problem is the subject of \cref{sec:openscvx-framework}.

\subsection{Symbolic Expression Layer}
\label{sec:symbolic-expression-layer}

\begin{figure}[!t]
  \centering
  \tikzset{
    opnode/.style={circle, draw, ttc-blue, minimum size=6.5mm, inner sep=0.5pt, very thick, font=\scriptsize},
    leafnode/.style={circle, draw, ttc-red, minimum size=6.5mm, inner sep=0.5pt, very thick, font=\scriptsize},
    dagedge/.style={->, ttc-blue, thick, >={Latex[round]}}}
\begin{tikzpicture}[x=1mm, y=1mm]
  % ===== LEFT: authored cone constraint  a @ (p-w) >= ||p-w|| =====
  \node[opnode]   (ge)  at (0,0)             {$\geq$};
  \node[opnode]   (mm)  at ($(ge)+(-120:12)$) {$@$};
  \node[opnode]   (nm)  at ($(ge)+(-60:12)$)  {$\|\cdot\|$};
  \node[leafnode] (a)   at ($(mm)+(-120:12)$) {$a$};
  \node[opnode]   (d)   at ($(nm)+(-120:12)$) {$-$};
  \node[leafnode] (p)   at ($(d)+(-120:11)$)  {$p$};
  \node[leafnode] (w)   at ($(d)+(-60:11)$)   {$w$};
  \draw[dagedge] (ge) -- (mm);  \draw[dagedge] (ge) -- (nm);
  \draw[dagedge] (mm) -- (a);   \draw[dagedge] (mm) -- (d);
  \draw[dagedge] (nm) -- (d);
  \draw[dagedge] (d)  -- (p);   \draw[dagedge] (d)  -- (w);

  % ===== RIGHT: canonicalized  (a@(p-w) - ||p-w||) >= 0 =====
  \node[opnode]   (ge2) at ($(ge)+(50,0)$)     {$\geq$};
  \node[opnode]   (sb0) at ($(ge2)+(-120:12)$) {$-$};
  \node[leafnode] (z)   at ($(ge2)+(-60:12)$)  {$0$};
  \node[opnode]   (mm2) at ($(sb0)+(-120:12)$) {$@$};
  \node[opnode]   (nm2) at ($(sb0)+(-60:12)$)  {$\|\cdot\|$};
  \node[leafnode] (a2)  at ($(mm2)+(-120:12)$) {$a$};
  \node[opnode]   (d2)  at ($(nm2)+(-120:12)$) {$-$};
  \node[leafnode] (p2)  at ($(d2)+(-120:11)$)  {$p$};
  \node[leafnode] (w2)  at ($(d2)+(-60:11)$)   {$w$};
  \draw[dagedge] (ge2) -- (sb0);  \draw[dagedge] (ge2) -- (z);
  \draw[dagedge] (sb0) -- (mm2);  \draw[dagedge] (sb0) -- (nm2);
  \draw[dagedge] (mm2) -- (a2);   \draw[dagedge] (mm2) -- (d2);
  \draw[dagedge] (nm2) -- (d2);
  \draw[dagedge] (d2)  -- (p2);   \draw[dagedge] (d2)  -- (w2);

  % ===== transform arrow =====
  \draw[-{Latex[round]}, very thick, ttc-gray]
    let
        \p1 = (nm.east),
        \p2 = ($(ge)!0.5!(w)$),
        \p3 = (mm2.west),
    in
        (\x1+1mm, \y2) -- node [above, pos=0.5, font=\scriptsize\itshape] {canonicalize} (\x3-1mm, \y2);
\end{tikzpicture}
\caption{
    The constraint \texttt{a @ (p - w) >= Norm(p - w)} as authored (left) and after canonicalization (right).
    Leaf nodes are red; operator nodes and the edges to their children are blue.
    Because the subexpression \texttt{p - w} feeds both the inner product and the norm, it has two parents -- the graph is a directed acyclic graph rather than a tree.
    Canonicalization rewrites the inequality into residual form, moving all terms to one side against a constant~$0$ while reusing the shared subgraph.
}
\label{fig:lower}
\end{figure}
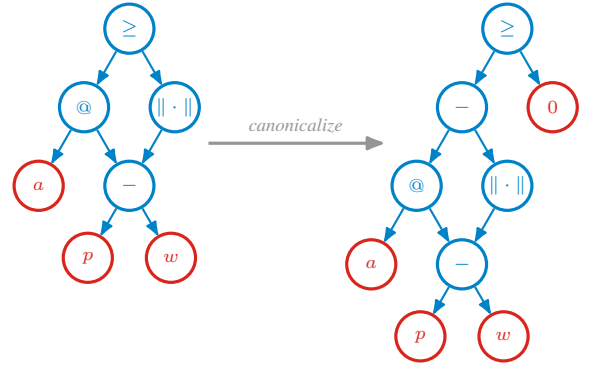

The symbolic expression system is implemented as an embedded \gls{dsl} in Python; when a user writes \texttt{Norm(p - w) <= 1.0} the expression is not eagerly evaluated, but stored as a graph for subsequent processing.
Indeed, in trajectory optimization, \texttt{p} and \texttt{w} will often represent variables, which would make eager evaluation nonsensical.
Instead, \openscvx stores the expressions symbolically; variables and constants make up leaf nodes while operators combine their child nodes into composite expressions, resulting in a \emph{graph} structure as illustrated in \cref{fig:lower}.
Because subexpressions may be reused by several parent expressions, and each node may only reference already constructed nodes, our graph is more precisely a \gls{dag}.

This \gls{dag} is the central data structure of \openscvx.
Every step between problem definition and numerical solution operates on the graph, which is successively inspected, transformed, augmented, canonicalized (\cref{fig:lower}, right), and finally lowered to executable code for the algorithm at hand.
The same graph is lowered to each of the two computation backends, \jax{} and \cvxpy, as \cref{sec:lowering} describes.

\subsubsection{Symbolic Lowering and Backend Compilation}
\label{sec:lowering}

To formulate and solve trajectory optimization problems, \openscvx translates symbolic expression graphs into executable code for its computational backends.
This process, known as \emph{lowering}, is a recursive traversal of the \gls{dag} in which each operator node is mapped to its counterpart in the target backend, a \jax{} primitive or a \cvxpy{} atom, and composed with the already-lowered translations of its children.
The translation is structure-preserving: the lowered computation mirrors the shape of the symbolic graph, and shared subexpressions can be reused.
Each backend is thus fully specified by a set of per-operator translation rules, and the same graph can be lowered to different backends at different points in the \gls{scvx} pipeline.

Each of the two main backends, \jax{} and \cvxpy, serves a different purpose.
The system dynamics, costs, and nonconvex constraints are lowered to pure \jax{} functions, after which no problem-specific numerical code is required: Jacobians come from automatic differentiation rather than hand derivation, and the discrete-time linearized dynamics consumed by the convex subproblem are produced by numerical integration of the lowered dynamics (\cref{sec:lindisc}).
Vectorization over trajectory segments, hardware acceleration, and ahead-of-time compilation then make these local approximations cheap to rebuild at every \gls{scvx} iteration.

\cvxpy{} is used to assemble the convex subproblem posed at each \gls{scvx} iteration, which combines three ingredients.
First, the local approximations produced by the \jax{} backend enter as \cvxpy{} \emph{parameters}:
the Jacobians and residuals of the dynamics and nonconvex constraints are declared symbolically as parameter matrices, whose numerical values are overwritten with the output of automatic differentiation at every iteration.
Second, expressions that are already convex, such as convex nodal constraints, bypass the linearization pipeline entirely and are lowered directly from the symbolic graph into \cvxpy{} atoms.
Third, the terms prescribed by the particular \gls{scvx} formulation, such as trust-region penalties and virtual-control slack variables, are appended by the algorithm itself.
The assembled subproblem satisfies the \gls{dcp} and \gls{dpp} rulesets by construction:
all nonconvexity has already been handled by linearization, so parameters enter only affinely, which is exactly the form \gls{dpp} requires, and \cvxpy{} verifies compliance when the problem is built. 
Building the subproblem layer on \cvxpy{} also gives \openscvx access to a mature ecosystem of conic and quadratic programming solvers, with optional solver code generation through \textsf{CVXPYGen}~\cite{schaller2022cvxpygen}.

This architecture is what keeps \openscvx modular.
The expression graph is the sole interface between the modeling layer and everything downstream, so the user-facing syntax, the \gls{scvx} algorithms, and the computational backends can each be developed and understood in isolation.
Extending the library therefore reduces to small, local additions.
A new operator is one node type plus one translation rule per backend, after which it composes freely with the existing vocabulary.
A new backend is only a set of translation rules over the existing nodes; the built-in \LaTeX~renderer is nothing more than a third such rules set alongside \jax{} and \cvxpy.
None of this is visible from the modeling layer, whose vocabulary the remainder of this section introduces by way of a worked example (\cref{sec:tutorial}); the full operator catalog appears in \cref{apx:atoms}.

\subsection{Tutorial: The Brachistochrone}
\label{sec:tutorial}

\begin{mathnote}{Brachistochrone Problem}
\begin{subequations}
\label{prob:brach}
\begin{align}
\min_{\theta, t_f} \quad & t_f \label{eq:brach:cost}\\
\text{s.t.} \quad
  & \dot{p}(t) = \begin{bmatrix}v(t)\sin\theta(t) \\ -v(t)\cos\theta(t) \end{bmatrix}, && t \in [0, t_f], \label{eq:brach:dyn-p}\\
  & \dot{v}(t)   = g\cos\theta(t),              && t \in [0, t_f], \label{eq:brach:dyn-v}\\
  & p(0) = [0, 10]^\top, && \label{eq:brach:p0}\\
  & p(t_f) = [10, 5]^\top,     && \label{eq:brach:pf}\\
  & v(0) = 0,                                   && \label{eq:brach:v0}\\
  & p(t) \in \begin{bmatrix}[0, 10] \\ [0,10]\end{bmatrix}, && t \in [0, t_f], \label{eq:brach:box-p}\\
  & v(t) \in [0, 10], && t \in [0, t_f], \label{eq:brach:box-v}\\
  & \theta(t) \in [0, \bar{\theta}],            && t \in [0, t_f]. \label{eq:brach:box-theta}
\end{align}
\end{subequations}
\end{mathnote}

We now put this vocabulary to work on a classical example: the Brachistochrone problem of finding the frictionless curve down which a bead, starting at rest, slides between two fixed points in minimum time.
Parameterizing the curve by its local angle $\theta(t)$ from the vertical yields the optimal control formulation below, which the remainder of this subsection translates into \openscvx piece by piece.

\subsubsection{Decision Variables \& Parameters}
\label{sec:decision-variables-parameters}

The leaf nodes of the expression graph are the decision variables, constants, and parameters of the \gls{ocp}.
\Cref{lst:openscvx-api-leaf} declares every leaf the Brachistochrone problem needs; we walk through its lettered sections in turn.

\begin{listing}[t]
\caption{Leaf node definition of the Brachistochrone problem in \openscvx. Continued with the expression definition and problem assembly in \cref{lst:openscvx-api-expr}.}
\label{lst:openscvx-api-leaf}
\setcounter{oxsec}{0}% restart the section letters at A for this listing
\begin{lstlisting}[style=oxapi]
import numpy as np
import openscvx as ox

n = 2  # decision nodes

(*@\oxsec@*)# States
position = ox.State("position", shape=(2,))
velocity = ox.State("velocity", shape=(1,))

# Define Bounds
position.max = [10.0, 10.0]
position.min = [0.0, 0.0]
velocity.max = [10.0]
velocity.min = [0.0]

# Define Boundary Conditions
position.initial = [0.0, 10.0]
position.final = [10.0, 5.0]
velocity.initial = [0.0]
velocity.final = [ox.Free(10.0)]

(*@\oxsec@*)# Controls
theta = ox.Control("theta", shape=(1,))
theta.max = [100.5 * np.pi / 180]
theta.min = [0.0]
theta.guess = np.zeros((n, 1))

(*@\oxsec@*)# Time horizon
time = ox.Time(
    initial=0.0,
    final=ox.Minimize(2),
    min=0.0,
    max=2,
)

(*@\oxsec@*)# Parameter
g = ox.Parameter("g", shape=(1,), value=9.81)

# ... continued in (*@\cref{lst:openscvx-api-expr}@*)
\end{lstlisting}
\end{listing}

\paragraph{States}
The bead carries two states, its position $p$ and speed $v$, each declared in section~A of \cref{lst:openscvx-api-leaf} with a name and a shape.
The problem data of \cref{prob:brach} then attaches directly as attributes: the box bounds (\cref{eq:brach:box-p,eq:brach:box-v}) become \texttt{min} and \texttt{max}, and the endpoint conditions (\crefrange{eq:brach:p0}{eq:brach:v0}) become \texttt{initial} and \texttt{final}.
A boundary value given as a plain number is held fixed; wrapping it in a marker relaxes it instead, as with \texttt{ox.Free(10.0)}, which leaves the final speed unconstrained and supplies only its initial guess.
Between the endpoints, the initial state trajectory defaults to linear interpolation and can be overridden through the \texttt{guess} attribute (\cref{lst:guesses} shows the accepted forms).

\paragraph{Controls}
The single control is the wire angle \texttt{theta}, declared in section~B with the same name, shape, and bound attributes (\cref{eq:brach:box-theta}).
Two things distinguish controls from states.
First, the continuous control signal is reconstructed from its values at the decision nodes by a choice of parameterization: \gls{foh} by default, with \gls{zoh} and impulsive schemes available (\cref{par:control_param}).
Second, there is no default initial guess: a good control guess is inherently problem specific, so \texttt{guess} must be set explicitly.

\paragraph{Time and objective}
\label{sec:objective}
Time is declared through a dedicated \texttt{Time} object (section~C), which behaves as an additional state.
Its final boundary condition is where the objective (\cref{eq:brach:cost}) enters the problem: the marker \texttt{ox.Minimize(2)} leaves $t_f$ free, supplies $2$ as its initial guess, and adds it to the objective, making this a minimum-time problem.
The same markers apply to the boundary condition of \emph{any} state, with \texttt{Maximize} available analogously.
Running costs are expressed through the same mechanism via an auxiliary state (\cref{sec:beyond-tutorial}).

\paragraph{Parameters}
The gravitational acceleration $g$ is declared as a \texttt{Parameter} (section~D): a symbolic placeholder that is constant during optimization but whose value can be changed between solves without recompilation, the mechanism behind the compile-once, solve-many workflow of \cref{sec:working-with-solutions}.

\subsubsection{Composing Expressions}
\label{sec:composing-expressions}

With the leaves in place, the operators of the symbolic layer combine them into the \emph{expressions} that complete the problem: the dynamics and the constraints of \cref{prob:brach}.

\begin{listing}[t]
\caption{Expression definition for an \openscvx optimal control problem. Continues from leaf node definition from \cref{lst:openscvx-api-leaf}.}
\label{lst:openscvx-api-expr}
\setcounter{oxsec}{0}% restart the section letters at A for this listing
\begin{lstlisting}[style=oxapi]
# ... continued from (*@\cref{lst:openscvx-api-leaf}@*)

(*@\oxsec@*)# Dynamics
dynamics = {
    "position": ox.Concat(
        velocity * ox.Sin(theta),   # x_dot
        -velocity * ox.Cos(theta),  # y_dot
    ),
    "velocity": g * ox.Cos(theta),
}


(*@\oxsec@*)# Constraints
constraints = [
  ox.ctcs(position <= position.max),
  ox.ctcs(position.min <= position),
  ox.ctcs(velocity <= velocity.max),
  ox.ctcs(velocity.min <= velocity),
]  # more options: (*@\crefrange{lst:constraints-dt}{lst:constraints-ctcs}@*)

# ... continued in (*@\cref{lst:openscvx-api-prob}@*)
\end{lstlisting}
\end{listing}

\paragraph{Dynamics}
The dynamics (\cref{eq:brach:dyn-p,eq:brach:dyn-v}) are written exactly as they appear on paper: a dictionary maps each state's name to the expression for its time derivative, as shown in section~A of \cref{lst:openscvx-api-expr}.
The expressions compose the leaves with ordinary operators (\texttt{ox.Sin}, \texttt{ox.Cos}, multiplication, concatenation), and the parameter \texttt{g} enters symbolically, so its value can change between solves without touching the dynamics.

\paragraph{Constraints}
The box bounds of \crefrange{eq:brach:box-p}{eq:brach:box-v} are imposed in section~B, each wrapped in \texttt{ox.ctcs}, which enforces the inequality in continuous time rather than only at the decision nodes (\cref{sec:ctcs-reformulation}).
Continuous-time satisfaction is what makes the two-node grid in \cref{lst:openscvx-api-leaf} sufficient: the constraints hold between nodes by the constraint augmentation technique in \cref{sec:ctcs-reformulation}, and along the optimal cycloid the wire angle varies linearly in time, so a single first-order-hold segment represents the optimal control exactly.
Richer constraint forms, including nodal constraints, sub-interval enforcement, and penalty selection, are covered in \cref{sec:beyond-tutorial}.

\subsubsection{Assembly and Solution}
\label{sec:problem-assembly}

All ingredients of \cref{prob:brach} are now in hand, and the \texttt{Problem} constructor collects them into a single object, as shown in \cref{lst:openscvx-api-prob}; the only new information is the number of decision nodes \texttt{N}.
Two calls then produce a solution.
\texttt{initialize()} compiles the problem: the expression graph is canonicalized, augmented, and lowered as described in \cref{sec:lowering}, the \jax{} functions are compiled ahead of time, and a first throwaway solve of the convex subproblem caches its \gls{dpp} canonicalization for reuse in every subsequent solve.
\texttt{solve()} then runs the \gls{scvx} loop of \cref{sec:openscvx-framework} and returns the optimized trajectory; working with the returned solution is the subject of \cref{sec:working-with-solutions}.

\begin{listing}[t]
\caption{The Brachistochrone problem is assembled from the leaves (\cref{lst:openscvx-api-leaf}) and expressions (\cref{lst:openscvx-api-expr}) and solved with two calls.}
\label{lst:openscvx-api-prob}
\begin{lstlisting}[style=oxapi]
# ... continued from (*@\cref{lst:openscvx-api-expr}@*)

prob = ox.Problem(
    dynamics=dynamics,
    constraints=constraints,
    states=[position, velocity],
    controls=[theta],
    time=time,
    N=n,
)

prob.initialize()
results = prob.solve()
\end{lstlisting}
\end{listing}

Every algorithmic component of the resulting pipeline is an argument of this assembly with a sensible default.
\Cref{lst:openscvx-api-options} shows the overrides users reach for most: the discretization scheme, whose families and trade-offs are discussed in \cref{sec:lindisc}, the convex solver backend, and the weight-update policy of the \gls{scvx} loop (\cref{sec:hyperparameter-update}).
Discretizers, autotuners, subproblem solvers, and \gls{scvx} algorithms all follow this pattern: each is a swappable module behind a small interface, so a custom implementation can be supplied without touching the problem definition (\cref{sec:modularity}).

\begin{listing}[t]
\caption{Algorithmic components are selected at assembly through keyword arguments with sensible defaults, shown here for the discretization scheme, the convex solver backend, and the \gls{scvx} weight-update policy.}
\label{lst:openscvx-api-options}
\begin{lstlisting}[style=oxapi]
prob = ox.Problem(
    ...,  # as in (*@\cref{lst:openscvx-api-prob}@*)
    discretizer=ox.DiscretizeLinearizeVectorize(),
    solver=ox.MoreauPTRSolver(),
    algorithm={"autotuner": ox.AdaptiveProximalWeight()},
)
\end{lstlisting}
\end{listing}

\subsection{Beyond the Tutorial}
\label{sec:beyond-tutorial}

The tutorial needed only box constraints enforced in continuous time.
This subsection surveys the rest of the modeling vocabulary through self-contained snippets, from richer constraint forms to alternative sources of dynamics, derivatives, and initial guesses.

\paragraph{Nodal constraints}
A constraint expression left unwrapped is enforced at the decision nodes, every node by default; \texttt{.at(...)} restricts it to a chosen subset, the natural form for waypoints, racing gates, and other pointwise requirements (\cref{lst:constraints-dt}).
Chaining \texttt{.convex()} additionally asserts that the expression is convex: it then bypasses linearization entirely and is lowered directly into the convex subproblem, where it is enforced exactly rather than penalized (\cref{eq:scp_subproblem}).
The assertion is checked, not trusted: expressions that are not \gls{dcp}-compliant are rejected when the problem is built.

\begin{listing}[t]
\caption{Nodal constraints: \texttt{.at()} selects the nodes at which a constraint is enforced, and \texttt{.convex()} routes a \gls{dcp}-compliant expression directly into the convex subproblem.}
\label{lst:constraints-dt}
\begin{lstlisting}[style=oxapi]
constraints = [
    (pos == waypoint_pos).at(2),
    (ox.linalg.Norm(pos) <= max_dist).at(n-1).convex(),
]
\end{lstlisting}
\end{listing}

\paragraph{Continuous-time constraints}
The \texttt{ox.ctcs} wrapper from the tutorial (\cref{lst:openscvx-api-expr}) generalizes along two axes, shown in \cref{lst:constraints-ctcs} for an obstacle-avoidance constraint.
\texttt{.over((3, 5))} narrows enforcement from the full horizon, the default, to the sub-interval between two nodes.
Each such window receives its own violation integrator, so it can be tuned and monitored independently of the others (\cref{sec:ctcs-reformulation}).
The \texttt{penalty} argument selects the differentiable exterior penalty through which violations are measured: one of the built-in names, or any penalty authored in the symbolic layer itself, supplied as a function of the constraint residual and subject to the admissibility conditions of \cref{sec:ctcs-reformulation}.
The choice mainly affects the convergence behavior of constraint satisfaction; the default is the squared \gls{relu}.

\begin{listing}[t]
\caption{Declaring a continuous-time (\texttt{ctcs}) constraint: over the full
horizon (default), restricted to a sub-interval with \texttt{.over()}, with a
built-in \texttt{penalty} by name, or with a penalty composed in the symbolic
layer.}
\label{lst:constraints-ctcs}
\begin{lstlisting}[style=oxapi]
# 1. full horizon (default)
constr = ox.ctcs(ox.linalg.Norm(pos - obs_pos) >= 1)

# 2. restricted to nodes 3..5
constr = ox.ctcs(ox.linalg.Norm(pos - obs_pos) >= 1).over((3, 5))

# 3. built-in penalty by name (default: squared ReLU)
constr = ox.ctcs(ox.linalg.Norm(pos - obs_pos) >= 1, penalty="smooth_relu")

# 4. any penalty composed from symbolic operators
constr = ox.ctcs(
    ox.linalg.Norm(pos - obs_pos) >= 1,
    penalty=lambda r: ox.Huber(ox.PositivePart(r), delta=0.5),
)
\end{lstlisting}
\end{listing}

\paragraph{Signal temporal logic specifications}
\label{par:stl}
\openscvx also supports temporal and logical specifications through Signal Temporal Logic (STL), allowing constraints to depend on how the trajectory evolves over time. 
For example, STL can express requirements such as a safety condition that must always hold over an interval or a goal that must eventually be reached within a specified time window.
This provides a concise way to encode mission-level requirements that would otherwise require manually constructing multiple time-dependent constraints.
The \texttt{ox.stl} operators compose ordinary inequality predicates into smooth \gls{stl} specifications via \gls{gmsr} \cite{uzun2024optimizationtemporallogicalspecifications, uzun2026successive} and additional smooth function provide by the \textsf{stljax} package \cite{Kapoor2025-mp}.
Logical connectives \texttt{And}, \texttt{Or}, \texttt{Not}, and \texttt{IfThen}, together with the temporal operator \texttt{Always}, are imposed on the trajectory in the same style as \texttt{ox.ctcs}, as shown in \cref{lst:constraints-stl}.
\texttt{.over((i, j))} enforces the formula continuously over a node window, while \texttt{.at([k$_1$,\ldots,k$_m$])} enforces it nodally at selected indices, exactly as for ordinary constraints (\cref{lst:constraints-dt,lst:constraints-ctcs}).

\begin{listing}[t]
\caption{Signal temporal logic specifications can be expressed \emph{over} continuous intervals (\texttt{.over()}) or \emph{at} discrete points along the trajectory (\texttt{.at()}). \texttt{reach\_a, reach\_b, near, slow} are predicate \openscvx expressions made up of the operators in \cref{apx:atoms}.}
\label{lst:constraints-stl}
\begin{lstlisting}[style=oxapi]
reach = ox.stl.Or(reach_a, reach_b)
constraints.append(reach.at([n//2, n-1]))          # nodal
constraints.append(reach.over((n-3, n-1)))         # CTCS window
constraints.append(
    ox.stl.IfThen(near, slow).over((0, n-1))
)
\end{lstlisting}
\end{listing}

\paragraph{Running costs}
The Brachistochrone objective (\cref{eq:brach:cost}) was purely terminal; a running cost such as the control effort $\int u^2\,\mathrm{d}t$ is expressed with the same boundary-condition markers after casting it into Mayer form (\cref{prob:mayer}) through state augmentation.
The augmentation uses only vocabulary already in hand: a \texttt{cost} state whose dynamics are the integrand and whose final value carries the \texttt{Minimize} marker, as in \cref{lst:objective}.
No dedicated objective API exists, and none is needed: minimum-time, terminal-cost, and running-cost problems are all the same construction.

\begin{listing}[t]
\caption{A running cost $\int u^2\,\mathrm{d}t$ in Mayer form: an augmented state integrates the cost, and minimizing its final value minimizes the integral.}
\label{lst:objective}
\begin{lstlisting}[style=oxapi]
# A state whose dynamics are the integrand
cost = ox.State("cost", shape=(1,))
cost.initial = [0.0]

dynamics = {"cost": u**2}

# Minimizing its final value minimizes the integral
cost.final = [ox.Minimize(0.0)]
\end{lstlisting}
\end{listing}

\paragraph{Initial guesses}
Guesses, too, are authored in the symbolic layer.
Beyond a plain array, \texttt{guess} accepts an expression in other variables, evaluated node-by-node with each referenced variable at its own guess, or a callable of \texttt{tau}, the normalized time running from $0$ to $1$ (\cref{lst:guesses}).

\begin{listing}[t]
\caption{Initial guesses may be arrays, expressions in other variables' guesses, or callables of the normalized time \texttt{tau}.}
\label{lst:guesses}
\begin{lstlisting}[style=oxapi]
# From other variables' guesses
speed.guess = ox.linalg.Norm(velocity)

# As a function of normalized time
position.guess = lambda tau: p0 + (pf - p0) * tau
\end{lstlisting}
\end{listing}

\paragraph{Discrete dynamics and impulsive controls}
Alongside the continuous flow, \openscvx supports a second dynamics mechanism: discrete transitions, in which a state jumps instantaneously, whether at a thruster burn, a contact event, a mass deployment, \emph{etc.}
Transitions are declared through a second dictionary, \texttt{dynamics\_discrete}, in the same state-to-expression form as the continuous dynamics.
Every state appears in the map, and one without a jump maps to itself, as \texttt{pos} does in \cref{lst:dynamics-discrete}.
Actuation enters these transitions through controls declared with the \texttt{impulsive} parameterization, which may only act at the decision nodes listed in \texttt{nodes}: here \texttt{delta\_v} kicks the velocity only at the first and last node.
The transition map itself is applied at every node, so a transition is effectively confined to its control's nodes only when it reduces to the identity at zero control; any control-independent term takes effect at all nodes.
\texttt{dynamics\_discrete} is passed to \texttt{Problem} alongside \texttt{dynamics}, required when, and only when, an impulsive control is declared.

\begin{listing}[t]
\caption{Discrete dynamics with an impulsive control: every state maps to its transition, and \texttt{delta\_v} may act only at its declared \texttt{nodes}.}
\label{lst:dynamics-discrete}
\begin{lstlisting}[style=oxapi]
# An impulsive control, nonzero only at its declared nodes
dv = ox.Control(
    "delta_v", shape=(3,),
    parameterization="impulsive",
    nodes=[0, n - 1],
)

# Instantaneous state transition, applied at the nodes
dynamics_discrete = {
    "pos": pos,       # no jump
    "vel": vel + dv,  # velocity jump
}

prob = ox.Problem(..., dynamics_discrete=dynamics_discrete)
\end{lstlisting}
\end{listing}

\paragraph{Custom Jacobians}
Linearization defaults to exact automatic differentiation, but any subexpression can carry a user-supplied Jacobian through \texttt{.with\_jacobian}, written in the same symbolic vocabulary (\cref{lst:with-jacobian}).
Directions without an override are still differentiated automatically, and the override rides discretization and linearization unchanged.
A deliberately simplified Jacobian is a classical device for conditioning \gls{scvx} iterations without altering the dynamics themselves.

\begin{listing}[t]
\caption{A user-supplied Jacobian replaces automatic differentiation for one subexpression and one variable; all other directions still differentiate automatically.}
\label{lst:with-jacobian}
\begin{lstlisting}[style=oxapi]
drag = -0.5 * rho * ox.linalg.Norm(vel) * vel

dynamics = {
    "vel": thrust / m + drag.with_jacobian({vel: J_simplified}),
}
\end{lstlisting}
\end{listing}

\paragraph{External dynamics}
Dynamics need not be authored symbolically at all.
Adapters wrap \jax-based physics engines and robot models: \texttt{FraxDynamics} builds states, controls, and dynamics from a URDF description, auto-populating joint and torque bounds (\cref{lst:external-dynamics}), and \texttt{MjxDynamics} does the same for \mujoco{} XML models through \textsf{MuJoCo XLA}, with the full setup in \cref{apx:integrations}.
The returned leaves are ordinary \texttt{State} and \texttt{Control} objects, so bounds, boundary conditions, costs, and constraints compose with them exactly as in the tutorial.

\begin{listing}[t]
\caption{A robot model becomes states, controls, and dynamics through an adapter; the \mujoco{} counterpart appears in \cref{apx:integrations}.}
\label{lst:external-dynamics}
\begin{lstlisting}[style=oxapi]
robot = frax.load_panda()   # URDF, bundled with frax
dyn = ox.FraxDynamics(robot)

q, qd = dyn.states          # ordinary State objects
(tau,) = dyn.controls       # an ordinary Control
\end{lstlisting}
\end{listing}

\subsection{Working with Solutions}
\label{sec:working-with-solutions}
The symbolic vocabulary that defined the problem carries over to its solution: every state and control in an \texttt{OptimizationResults} object is retrieved by the name it was declared with, as \cref{lst:results} shows.

\paragraph{Three solution forms}
The solution is available as optimizer nodes and through two forms of nonlinear propagation.
\texttt{results.nodes} holds the optimizer's solution at the $N$ decision nodes: \texttt{results.nodes["position"]} is an $(N, 2)$ array for the Brachistochrone.
\texttt{results.multishot\_propagation()} exposes the nonlinear integrations performed segment-by-segment between nodes during \gls{scvx} (\cref{sec:lindisc}), either per segment or stitched chronologically, and accepts \texttt{iteration=$k$} to inspect earlier iterates.
Finally, \texttt{prob.post\_process()} performs a nonlinear single-shot propagation of the converged control trajectory over the full horizon on a dense time grid, sampled with time step \texttt{prob.settings.prp.dt} (0.01 time units by default).
The single-shot trajectory populates \texttt{results.trajectory} with the propagated states, including propagation-only quantities such as forward kinematics, alongside \texttt{results.trajectory["time"]}.

\begin{listing}[t]
\caption{The solution is accessible as optimizer nodes, segment-by-segment nonlinear propagation, or full-horizon nonlinear propagation.}
\label{lst:results}
\begin{lstlisting}[style=oxapi]
results = prob.post_process()

# Optimizer solution at decision nodes
pos_nodes = results.nodes["position"]     # (n, 2)

# Segment-by-segment nonlinear propagation
prop = results.multishot_propagation()
pos_ms, t_ms = prop.state("position")

# Full-horizon nonlinear propagation
t_grid   = results.trajectory["time"]
pos_traj = results.trajectory["position"]
\end{lstlisting}
\end{listing}

\paragraph{Parameter updates}
Parameters close the loop the tutorial opened: because \texttt{g} entered the problem symbolically, its value is runtime data rather than compiled structure.
After \texttt{initialize()}, assigning through \texttt{prob.parameters} and re-solving reuses the compiled problem unchanged (\cref{lst:parameters}), the same mechanism \gls{dpp} affords the convex subproblem (\cref{sec:lowering}).
This compile-once, solve-many workflow is what suits \openscvx to \gls{mpc}, moving targets and obstacles, and model retuning between solves.

\begin{listing}[t]
\caption{A \texttt{Parameter} declared at modeling time is updated between solves without recompilation.}
\label{lst:parameters}
\begin{lstlisting}[style=oxapi]
# New problem data, same compiled problem
prob.parameters["g"] = 3.71   # the Martian Brachistochrone
results = prob.solve()
\end{lstlisting}
\end{listing}

\begin{figure}[h]
    \centering
    \includegraphics[width=\linewidth]{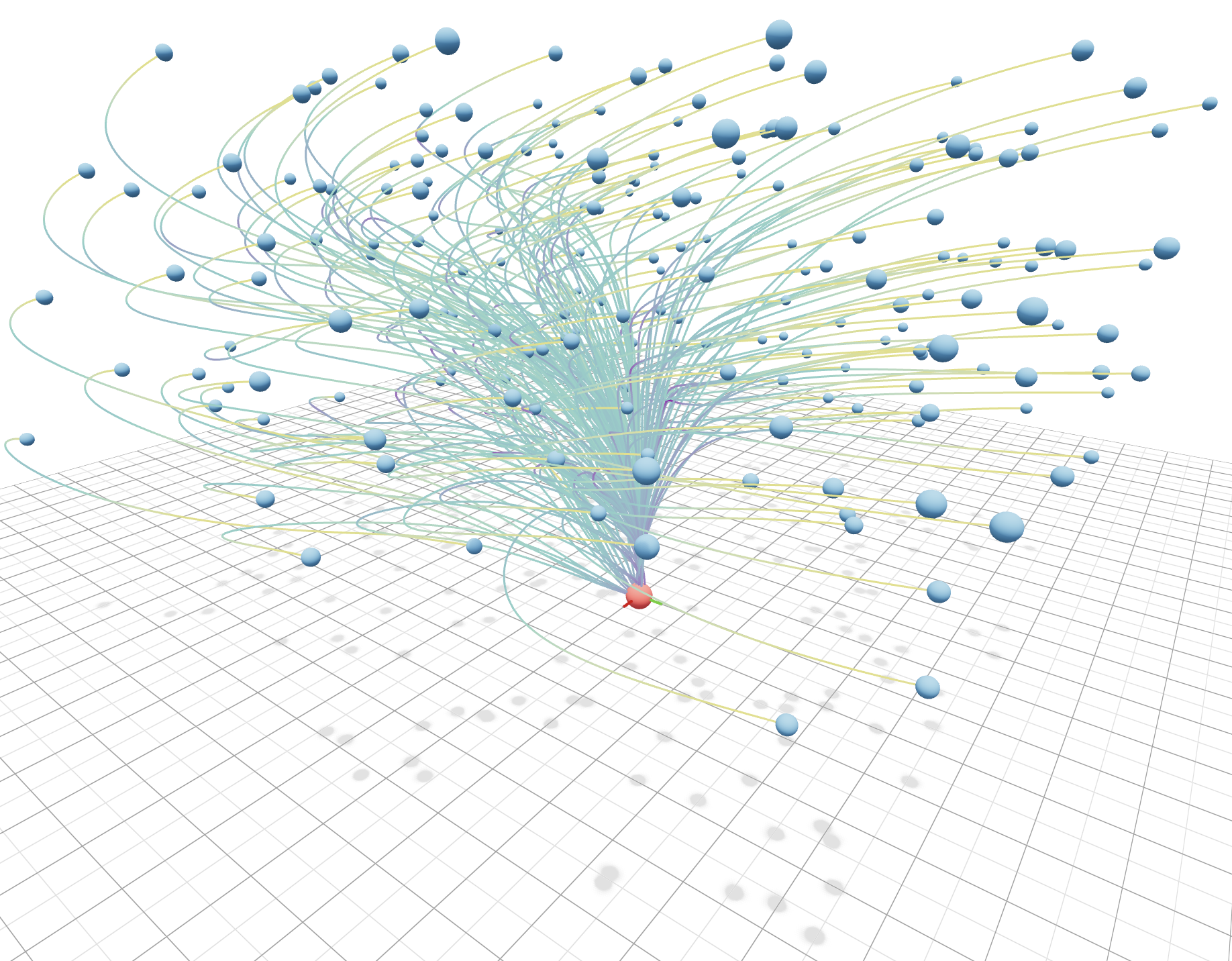}
    \caption{Monte Carlo analysis of 6DoF \glsentrylong{pdg} (\glsentryshort{pdg}): 200 random initial positions denoted by the blue points. The coloring of the trajectories denotes the 2-norm of velocity, $\|v\|_2$.}
    \label{fig:6DoF_batch}
\end{figure}

\paragraph{Batched solves}
For Monte Carlo campaigns and parameter sweeps, \texttt{solve\_batched()} solves a family of problems in one call: any input gains a leading batch dimension while the rest is shared across the batch (\cref{lst:batching}).
Underneath, \texttt{solve\_jax()} expresses the entire \gls{scvx} loop as a \jax{} function returning an \texttt{OptimizationResults} pytree, so it composes with \texttt{jax.jit} and \texttt{jax.vmap} like any other \jax{} function.
The trade-off is the loss of \texttt{solve()}'s interactive diagnostics, such as multishot propagation.
The convex backend selected at assembly (\cref{lst:openscvx-api-options}) determines how the batch is solved: \jax-native backends such as \textsf{qpax} and \textsf{moreau} solve the batched subproblems in parallel, while the \cvxpy{} backend solves them sequentially on the host at each iteration, the \gls{scvx} loops still advancing in lockstep.
With a \jax-native backend the compiled computation can additionally be exported and cached on disk (\texttt{save\_compiled} in the problem configuration), amortizing compilation across sweeps.
\Cref{fig:6DoF_batch} shows a 200-sample initial-condition sweep of a six-degree-of-freedom powered-descent problem solved this way.

\begin{listing}[t]
\caption{Common batching patterns: any input may be batched by adding a leading dimension $B$; unbatched inputs are shared across all solves.}
\label{lst:batching}
\begin{lstlisting}[style=oxapi]
# Initial-condition sweep
results = prob.solve_batched(x_initial=x0_batch)          # (B, n_x)

# Parameter sweep
results = prob.solve_batched(
    parameters={"target_pos": target_batch},              # (B, 3)
)

# Hyperparameter sweep
results = prob.solve_batched(
    algorithm={"lam_prox": lam_prox_batch},               # (B,)
)
\end{lstlisting}
\end{listing}

\section{How \openscvx solves the optimal control problem}
\label{sec:openscvx-framework}

The symbolic modeling interface introduced in \cref{sec:modeling-interface} allows users to describe \glspl{ocp} independently of the underlying solution method.
Once assembled, the resulting problem is solved using a successive convexification procedure, where the original nonlinear continuous-time \gls{ocp} is iteratively linearized and discretized to produce a sequence of convex subproblems. This is described in detail in \cref{sec:ctscvx}.

A central design principle of \openscvx is the separation between problem specification and algorithmic solution. 
This allows the same user-defined problem to be solved using different successive convexification strategies without modification of the modeling layer.
This modularity is discussed in \cref{sec:modularity}.

\subsection{\ctscvx}
\label{sec:ctscvx}

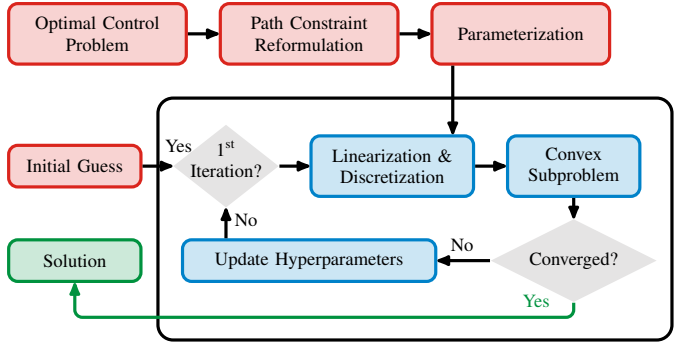
\begin{figure}[!t]
    \centering
    %% ===========================================================================
%%  OpenSCvx algorithm flowchart  --  bare TikZ fragment (\input this file)
%%
%%  Libraries (already loaded by the main document):
%%      calc, arrows.meta, shapes.geometric  (+ shadows, if `shadow' is used)
%%  Colors (defined globally in the preamble):
%%      ttc-red, ttc-blue, ttc-gray, ttc-green (+ their ttc-*-dark shades)
%%
%%  Drawn with x=1mm, y=1mm, so every number below is a millimetre on the
%%  page.  Total width is set by \xConR: ~88mm ~ 1 IEEE column, so it can be
%%  \input directly, without a \resizebox.
%% ===========================================================================
\begin{tikzpicture}[x=1mm, y=1mm, >={Latex[round, scale=0.8]}]

%% ---------------------------------------------------------------------------
%%  LAYOUT KNOBS -- everything positional lives in this block
%% ---------------------------------------------------------------------------
%% -- column centres (x, mm) -------------------------------------------------
\def\xOcp{12}        % "Optimal Control Problem"        (top row, col 1)
\def\xPcr{40}        % "Path Constraint Reformulation"  (top row, col 2)
\def\xPar{68}        % "Parameterization"               (top row, col 3)
                     % "Initial Guess" has no column knob: it is anchored so that
                     %   its west edge lines up with "Optimal Control Problem"
\def\xDia{29}        % "1st Iteration?" diamond + junction dot
\def\xLin{51}        % "Linearization & Discretization"
\def\xCvx{75}        % "Convex Subproblem" and "Converged?" share this column
\def\xUpd{40}        % "Update Hyperparameters" (must span \xDia)
                     % "Solution" has no column knob either: same west edge,
                     %   and the same size, as "Initial Guess" above it
\def\xDrop{59}       % where the Parameterization arrow drops into the loop
                     %   (must lie under the red box AND over the blue one)

%% -- row centres (y, mm; negative = downwards) ------------------------------
\def\yTop{-4}        % top row of red blocks
\def\yLoop{-21.5}    % main loop row: diamond, Linearization, Convex Subproblem
\def\yRet{-34}       % return row: "Update Hyperparameters", "Converged?", and
                     %   "Solution" share it, so the "No" arrow is horizontal
\def\yExit{-41.5}    % the horizontal run of the "Yes" line, below the return
                     %   row; it turns back up to enter Solution from the south

%% -- SCP loop container (rounded black rectangle) ---------------------------
\def\xConL{20}       % left wall  (crossed by the Initial Guess / Solution arrows)
\def\xConR{88}       % right wall (also sets the total figure width)
\def\yConT{-12.5}    % top edge
\def\yConB{-44.5}    % bottom edge

%% -- box sizes (minimum width/height; a node grows if its text is wider) ----
\def\wTop{23.5mm}    \def\hTop{8mm}   % red blocks (uniform across the top row)
\def\wGuess{17.5mm}  \def\hOne{5.5mm} % single-line blue blocks
\def\wLin{21.5mm}    \def\hTwo{8mm}   % two-line blue blocks
\def\wCvx{17.5mm}    %                  "Convex Subproblem"
\def\wUpd{33.5mm}    %                  "Update Hyperparameters"
\def\wDia{14mm}      \def\hDia{11mm}  % "1st Iteration?" decision diamond
\def\wConv{22mm}     \def\hConv{11mm} % "Converged?" decision diamond
                     %   Both diamonds are empty nodes with the label set
                     %   separately, so a long word widens neither.

%% -- corner radii (all strokes use TikZ's named `very thick') ----------------
\def\rBox{1.2mm}     % corner radius of the blocks
\def\rLoop{2mm}      % corner radius of the container
\def\rBend{1mm}      % corner radius at arrow bends
\def\gapArrow{0mm}   % gap left between an arrowhead and the block it points at

%% ---------------------------------------------------------------------------
%%  STYLES
%% ---------------------------------------------------------------------------
\tikzset{
  %% add `shadow' to `block' below to give every block a drop shadow
  shadow/.style   = {drop shadow={shadow xshift=0.35mm, shadow yshift=-0.35mm,
                                  opacity=0.45, fill=black!65}},
  block/.style    = {draw, very thick, rounded corners=\rBox,
                     align=center, font=\scriptsize, inner sep=1mm,
                     minimum height=\hTwo},
  red/.style      = {block, draw=ttc-red,   fill=ttc-red-light},
  blue/.style     = {block, draw=ttc-blue,  fill=ttc-blue-light},
  green/.style    = {block, draw=ttc-green, fill=ttc-green-light},
  %% both decision diamonds: a bare gray diamond, sized by its own knobs
  switch/.style   = {diamond, fill=ttc-gray!25, inner sep=0,
                     align=center, font=\scriptsize},
  flow/.style     = {->, very thick, rounded corners=\rBend,
                     shorten >=\gapArrow},
  lbl/.style      = {font=\scriptsize, inner sep=0.6mm},
}

%% ---------------------------------------------------------------------------
%%  SCP LOOP CONTAINER  (drawn first so the arrows cross over it)
%% ---------------------------------------------------------------------------
\draw[very thick, rounded corners=\rLoop]
      (\xConL,\yConT) rectangle (\xConR,\yConB);

%% ---------------------------------------------------------------------------
%%  NODES
%% ---------------------------------------------------------------------------
%% -- problem set-up (red, above the loop) -----------------------------------
\node[red, minimum width=\wTop] (ocp) at (\xOcp,\yTop) {Optimal Control\\Problem};
\node[red, minimum width=\wTop] (pcr) at (\xPcr,\yTop) {Path Constraint\\Reformulation};
\node[red, minimum width=\wTop] (par) at (\xPar,\yTop) {Parameterization};

%% -- loop entry / exit (outside the container, flush left with "Optimal
%%    Control Problem" and sized alike, so they read as one column) ----------
\node[red,  minimum width=\wGuess, minimum height=\hOne, anchor=west]
                                (guess) at (ocp.west |- 0,\yLoop) {Initial Guess};
\node[green, minimum width=\wGuess, minimum height=\hOne, anchor=west]
                                (sol)   at (ocp.west |- 0,\yRet) {Solution};

%% -- the SCP iteration ------------------------------------------------------
\node[switch, minimum width=\wDia, minimum height=\hDia]
      (dia) at (\xDia,\yLoop) {};
\node[lbl, align=center] at ($(dia.center)+(0, 1)$) {1\textsuperscript{st}\\Iteration?};

\node[blue, minimum width=\wLin]  (lin) at (\xLin,\yLoop) {Linearization \&\\Discretization};
\node[blue, minimum width=\wCvx]  (cvx) at (\xCvx,\yLoop) {Convex\\Subproblem};
\node[switch, minimum width=\wConv, minimum height=\hConv]
                                  (con) at (\xCvx,\yRet) {};
\node[lbl] at (con.center) {Converged?};
\node[blue, minimum width=\wUpd,  minimum height=\hOne]
                                  (upd) at (\xUpd,\yRet) {Update Hyperparameters};

%% helper: the column the Parameterization arrow falls down
\coordinate (drop) at (\xDrop,0);

%% ---------------------------------------------------------------------------
%%  CONNECTORS  (strictly horizontal / vertical, rounded at the bends)
%% ---------------------------------------------------------------------------
%% -- problem set-up chain ---------------------------------------------------
\draw[flow] (ocp.east) -- (pcr.west);
\draw[flow] (pcr.east) -- (par.west);
\draw[flow] (drop |- par.south) -- (drop |- lin.north);

%% -- into the loop: the two labeled entries to the 1st-Iteration test -------
%%    The diamond is a merge: "Yes" admits the initial guess, "No" the
%%    retuned iterate; the labels sit on the incoming edges.
\draw[flow] (guess.east) -- (dia.west);
\node[lbl, anchor=south] at ($(dia.west)+(0.8,1.2)$) {Yes};

%% -- the iteration itself ---------------------------------------------------
\draw[flow] (dia.east) -- (lin.west);
\draw[flow] (lin.east) -- (cvx.west);
\draw[flow] (cvx.south) -- (con.north);
\draw[flow] (con.west) -- (upd.east);          % not converged -> retune ...
\node[lbl, anchor=south east] at ($(con.west)+(-1.5,0.6)$) {No};

\draw[flow] (dia.south |- upd.north) -- (dia.south);   % ... and re-enter
\node[lbl, anchor=north west] at ($(dia.south)+(0.5,0)$) {No};

%% -- converged: leave the loop ----------------------------------------------
\draw[flow, ttc-green] (con.south) |- (sol.south |- 0,\yExit) -- (sol.south);
\node[lbl, ttc-green, anchor=south east]
      at ($(con.south |- 0,\yExit)+(-2.5,0.6)$) {Yes};

\end{tikzpicture}
    \caption{\Acrfull{ctscvx}. The \textit{red blocks} are problem parsing steps, the \textit{blue blocks} are the iterative steps.}
    \label{fig:scvx}
\end{figure}

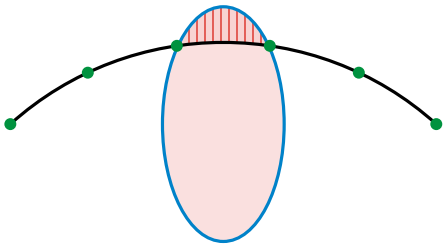
\begin{figure}
    \centering
    \begin{tikzpicture}[x=1.15cm, y=1.15cm]
\tikzset{
    constraint set/.style={fill=ttc-red-light!70},
    constraint boundary/.style={draw=ttc-blue, very thick},
    violation shade/.style={fill=ttc-red-light},
    violation hatch/.style={pattern={Lines[angle=90, distance=3pt, line width=0.5pt]},
                            pattern color=ttc-red},
    trajectory/.style={very thick, black, line cap=round, line join=round},
    node/.style={circle, fill=ttc-green, inner sep=0pt, minimum size=4.5pt}
}

% Equal arc-length samples of the circular arc.
\coordinate (n1) at (-2.45000, 0.00000);
\coordinate (n2) at (-1.55889, 0.59171);
\coordinate (n3) at (-0.53483, 0.90074);
\coordinate (n4) at ( 0.53483, 0.90074);
\coordinate (n5) at ( 1.55889, 0.59171);
\coordinate (n6) at ( 2.45000, 0.00000);

\useasboundingbox (-2.52,-1.42) rectangle (2.62,1.42);
\fill[white] (current bounding box.south west) rectangle (current bounding box.north east);

% Obstacle (keep-out ellipse)
\path[constraint set] (0,0) ellipse[x radius=0.70, y radius=1.35];

% Lens cut by the arc: ellipse minus the disk that carries the trajectory.
\begin{scope}
    \clip (0,0) ellipse[x radius=0.70, y radius=1.35];
    \fill[even odd rule, violation shade]
        (-1.2,-1.5) rectangle (1.2,1.5)
        (0, -2.72282) circle[radius=3.66282];
    \fill[even odd rule, violation hatch]
        (-1.2,-1.5) rectangle (1.2,1.5)
        (0, -2.72282) circle[radius=3.66282];
\end{scope}

\draw[constraint boundary] (0,0) ellipse[x radius=0.70, y radius=1.35];

% One circular arc, left to right over the top (clockwise sweep).
\draw[trajectory]
    (n1) arc[start angle=131.981, delta angle=-83.962, radius=3.66282];

\foreach \n in {n1, n2, n3, n4, n5, n6} {
    \node[node] at (\n) {};
}

\end{tikzpicture}
    \caption{Nodal feasibility does not imply continuous-time feasibility. All nodes (green) lie outside the obstacle (blue), yet the trajectory connecting them cuts through the obstacle (hatched red), motivating continuous enforcement.}
    \label{fig:ctcs}
\end{figure}

% \begin{figure}[!t]
%     \centering
%     \subfloat[Nodally enforced constraint.]{\includegraphics[width=0.48\linewidth]{figures/examples/dr_vp_nodal_single_camera_white.pdf}\label{fig:vel}}
%     \hfil
%     \subfloat[Continuous enforced constraint]{\includegraphics[width=0.48\linewidth]{figures/examples/dr_vp_single_camera_white.pdf}\label{fig:ang_vel}}
%     \caption{Demonstration of continuous versus discrete constraint enforcement. The traces should remain with constraint set shown in red.}
%     \label{fig:ctcs}
% \end{figure}

The \gls{ctscvx} procedure consists of four primary stages:
\begin{enumerate}
    \item \textbf{Continuous-time reformulation and parameterization:} the continuous dynamics and constraints are transcribed into a finite-dimensional representation over an adaptive time grid.
    
    \item \textbf{Discretization and linearization:} nonlinear dynamics and nonconvex constraints are approximated using local linear models, yielding a convex surrogate problem.
    
    \item \textbf{Convex subproblem:} the resulting convex optimization problem is solved to obtain candidate primal and dual variables.
    
    \item \textbf{Hyperparameter Update:} the candidate solution is evaluated, and algorithmic parameters such as trust-region radii, penalty weights, and virtual control terms are updated.
\end{enumerate}

Together, these steps define a standard successive convexification loop for continuous-time optimal control.
The following section details how each of these stages is instantiated within the \openscvx implementation and how alternative design choices recover different variants of \gls{scvx} methods.

\subsubsection{Path Constraint Reformulation and CTCS}
\label{sec:ctcs-reformulation}
Standard direct transcription enforces path constraints $g(t, x, u) \le 0_{n_g}$ and $h(t, x, u) = 0_{n_h}$ only at discrete knot points $t_k$, which can lead to intersample constraint violations during forward rollout \cite{Elango2025auto}. To evaluate feasibility continuously across each trajectory segment, \openscvx reformulates continuous path constraints into an integrated penalty state vector $y(\tau) \in \mathbb{R}^{n_y}$ using a vector violation function $\Lambda(t, x, u) \ge 0_{n_y}$:
\begin{align*}
    \Lambda_m(t, x, u) &= \sum_{i \in \mathcal{I}_m} q_i(g_i) + \sum_{j \in \mathcal{J}_m} p_j(h_j), \\
    \Lambda(t, x, u) &= [\Lambda_1, \dots, \Lambda_{n_y}]^\top,
\end{align*}
where $\mathcal{I}_m$ and $\mathcal{J}_m$ partition the inequality and equality path constraints assigned to the $m$-th integrator state. 
While constraints active across the entire horizon can be aggregated into a single scalar integrator state $y(\tau) \in \mathbb{R}$ ($n_y = 1$), \openscvx allows users to assign dedicated integrator components $y_m(\tau)$ to specific constraints for independent weight tuning and error tracking. 
Crucially, if a continuous constraint is restricted to a sub-interval $[\tau_{\text{start}}, \tau_{\text{end}}] \subset [0, 1]$ (e.g., using the \texttt{.over()} method, \cref{lst:constraints-ctcs}), it is automatically assigned its own dedicated integrator state $y_m(\tau)$ whose differential dynamics are evaluated exclusively over that sub-interval.

Following \cite[Sec. 2.3]{Elango2025auto}, the scalar exterior penalty functions $q_i : \mathbb{R} \to \mathbb{R}_+$ and $p_j : \mathbb{R} \to \mathbb{R}_+$ must satisfy two core conditions: (i) \emph{feasibility equivalence}, where $q_i(z) = 0 \iff z \le 0$ and $q_i(z) > 0 \iff z > 0$ (and $p_j(z) = 0 \iff z = 0$), and (ii) \emph{$C^1$ continuous differentiability} on $\mathbb{R}$ to ensure exact first-order ODE sensitivity integration across shooting intervals. A canonical choice satisfying both conditions is the squared \gls{relu} function $q_i(z) = (\max\{0, z\})^2$ for inequalities (and $p_j(z) = z^2$ for equalities), whose derivative $q_i'(z) = 2 \max\{0, z\}$ is continuous on $\mathbb{R}$ and vanishes smoothly at the boundary $z = 0$ \cite{Elango2025auto}.
In \openscvx the penalties are themselves symbolic expressions: the built-in choices are provided as such, and users may compose their own from the operator vocabulary, subject to conditions (i)--(ii) (\cref{lst:constraints-ctcs}).

As shown in \cite[Lemma 2]{Elango2025auto}, pointwise path constraint satisfaction almost everywhere on $[t_0, t_f]$ is equivalent to zero total integrated violation $\int_{t_0}^{t_f} \Lambda_m(t, x(t), u(t)) \, dt = 0$ for all $m = 1, \dots, n_y$. In practice, strictly enforcing zero integrated violation causes trivial \gls{licq} violations (see \cite[Lemma 10]{Elango2025auto}). \openscvx therefore enforces a small relaxation tolerance $\epsilon > 0$ elementwise across each shooting segment $k = 1, \dots, N-1$:
\begin{equation}
    y_{k+1} - y_k = \int_{\tau_k}^{\tau_{k+1}} s(\tau) \Lambda(t(\tau), x(\tau), u(\tau)) \, d\tau \le \varepsilon \, 1_{n_y}.
    \label{eq:ctcs_relaxation}
\end{equation}
Under this $\varepsilon$-relaxation, the maximum integrated constraint violation is bounded by $\varepsilon \cdot N$; we refer interested readers to \cite[Theorem 14]{Elango2025auto} for the detailed pointwise bound derivation and \gls{licq} conditions.
\subsubsection{Generalized Time Dilation}
Free-final-time \glspl{ocp} are mapped onto a fixed normalized time domain $\tau \in [0, 1]$ via generalized time dilation \cite{Elango2025auto, Kamath2023-pq}. We define a monotonic mapping $t : [0, 1] \to [t_0, t_f]$ with local rate $s(\tau) = \frac{dt(\tau)}{d\tau} > 0$. The physical time horizon identity is given by:
\begin{equation*}
    t_f - t_0 = \int_{0}^{1} s(\tau) \, d\tau.
\end{equation*}
The scalar dilation factor $s(\tau)$ is treated as an additional continuous control decision variable bounded by $0 < s_{\min} \le s(\tau) \le s_{\max}$, forming the augmented control vector $\tilde{u}(\tau) = [u(\tau)^\top, s(\tau)]^\top$.

\subsubsection{Complete Augmented System}
Combining the physical state $x(\tau)$, \gls{ctcs} violation states $y(\tau)$, and physical time state $t(\tau)$ yields the consolidated augmented state vector $\tilde{x}(\tau) \in \mathbb{R}^{n_x + n_y + 1}$:
\begin{equation*}
    \tilde{x}(\tau) := \begin{bmatrix} x(\tau) \\[2pt] y(\tau) \\[2pt] t(\tau) \end{bmatrix}.
\end{equation*}
Similarly, combining the physical continuous control vector $u(\tau)$ and the continuous time dilation factor $s(\tau)$ forms the augmented continuous control vector $\tilde{u}(\tau) \in \mathbb{R}^{n_u + 1}$:
\begin{equation*}
    \tilde{u}(\tau) := \begin{bmatrix} u(\tau) \\[2pt] s(\tau) \end{bmatrix}.
\end{equation*}
Applying the chain rule $\frac{d(\cdot)}{d\tau} = \frac{d(\cdot)}{dt} \frac{dt}{d\tau} = s(\tau) \frac{d(\cdot)}{dt}$, the complete augmented continuous differential system over normalized time $\tau \in [0, 1]$ becomes autonomous, its dynamics being expressed as follows.
\begin{equation*}
    \frac{d\tilde{x}(\tau)}{d\tau} = \begin{bmatrix} 
        s(\tau) \, f(t(\tau), x(\tau), u(\tau)) \\[2pt] 
        s(\tau) \, \Lambda(t(\tau), x(\tau), u(\tau)) \\[2pt] 
        s(\tau) 
    \end{bmatrix} =: \tilde{f}(\tilde{x}(\tau), \tilde{u}(\tau))
\end{equation*}
Discrete transitions act now at knot points $\tau_k \in \mathcal{I} := \{0, 1, \dots, N-1\}$ using the augmented discrete-time dynamics, defined as follows. 
$$
\tilde{x}(\tau_k^+) = \begin{bmatrix} 
        f_\textrm{d}(t(\tau_k^-), x(\tau_k^-), \mu((\tau_k^-))) \\[2pt] 
        y(\tau_k^-) \\[2pt] 
        t(\tau_k^-) 
    \end{bmatrix} =: \tilde{f}_\textrm{d}( \tilde{x}(\tau_k^-), \mu_k)
$$

\subsubsection{Control Parameterization}
\label{par:control_param}

In practice, the continuous control input is discretized using either a \gls{zoh} or \gls{foh} parameterization to obtain a finite-dimensional optimization problem.
Under \gls{zoh}, control is assumed to be constant on each interval, $u(\tau) = u_k$ for $\tau\in [\tau_k, \tau_{k+1})$. This might be suitable for actuators that cannot continously vary their signal.
In \gls{foh}, the control varies linearly between knot points, enforcing $u(\tau_k) = u_k$ and $u(\tau_{k+1}) = u_{k+1}$. This produces a piecewise-linear control profile that better captures smoothly varying inputs. This might be more suitable for actuators like motors that can vary their signal continuously. 

From this point on we assume the control profile has been parameterized over the whole horizon by a finite set of values $\tilde{u}_0, \dots, \tilde{u}_{N}$; however, we keeping the notation $\tilde{u}(\tau)$ for graphical ease. $\tilde{u}(\tau)$ must be then considered as a function of a finite set of numerical values, as depicted in \cref{fig:control_parameterization} for the \gls{zoh} and \gls{foh} parameterization. A more detailed mathematical description can be found at \cite{Elango2024ctscvx}. 

\textbf{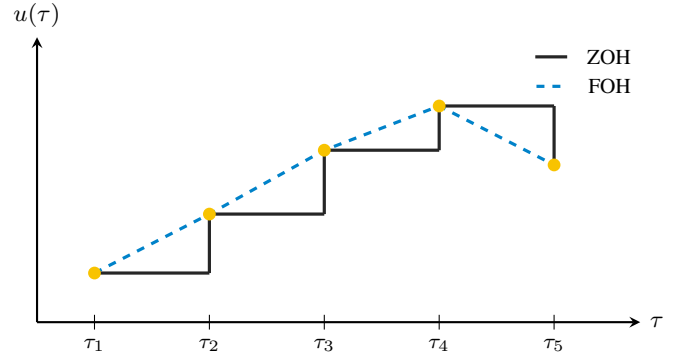
\begin{figure}[h!]
    \centering
    % Units are chosen so the picture's natural width is one IEEEtran column
% (252pt); it is included at true size, without \resizebox.
\begin{tikzpicture}[>=stealth, x=0.760cm, y=0.65cm]

% ----------------------------------------------------
% Styles
% ----------------------------------------------------
\tikzset{
    node/.style={circle, fill=ttc-yellow, draw=none, inner sep=1.7pt},
    zoh/.style={very thick, black!85},
    foh/.style={very thick, ttc-blue, dashed},
    grid/.style={thick, ttc-gray!40!white, dashed},
    axis/.style={thick, ->},
    axis label/.style={font=\small},
    tick label/.style={font=\footnotesize},
    legend label/.style={font=\footnotesize, anchor=east}
}

% ----------------------------------------------------
% Axes
% ----------------------------------------------------
\draw[axis] (0,0) -- (10.5,0) node[axis label, right] {$\tau$};
\draw[axis] (0,0) -- (0,5.8) node[axis label, above] {$u(\tau)$};

% ----------------------------------------------------
% Time nodes
% ----------------------------------------------------
\foreach \x/\label in {1/\tau_1, 3/\tau_2, 5/\tau_3, 7/\tau_4, 9/\tau_5} {
    \draw[thin] (\x,0.15) -- (\x,-0.15) node[tick label, below] {$\label$};
}

% ----------------------------------------------------
% Sample values
% ----------------------------------------------------
\coordinate (x1) at (1,1.0);
\coordinate (x2) at (3,2.2);
\coordinate (x3) at (5,3.5);
\coordinate (x4) at (7,4.4);
\coordinate (x5) at (9,3.2);

% ----------------------------------------------------
% FOH (linear interpolation)
% ----------------------------------------------------
\draw[foh] (x1) -- (x2) -- (x3) -- (x4) -- (x5);

% ----------------------------------------------------
% ZOH (piecewise constant / hold)
% ----------------------------------------------------
\draw[zoh] (1,1.0) -- (3,1.0);
\draw[zoh] (3,2.2) -- (5,2.2);
\draw[zoh] (5,3.5) -- (7,3.5);
\draw[zoh] (7,4.4) -- (9,4.4);

% vertical jumps for ZOH
\draw[zoh] (3,1.0) -- (3,2.2);
\draw[zoh] (5,2.2) -- (5,3.5);
\draw[zoh] (7,3.5) -- (7,4.4);
\draw[zoh] (9,4.4) -- (9,3.2);

% ----------------------------------------------------
% Nodes
% ----------------------------------------------------
\node[node] at (x1) {};
\node[node] at (x2) {};
\node[node] at (x3) {};
\node[node] at (x4) {};
\node[node] at (x5) {};

% ----------------------------------------------------
% Legend
% Labels are flush with the tip of the $\tau$ axis; each key is drawn
% relative to its label, so both rows share one right and one left edge.
% ----------------------------------------------------
\node[legend label] (zohlabel) at (10.5,5.4) {ZOH};
\draw[zoh] ([xshift=-16pt]zohlabel.west) -- ([xshift=-4pt]zohlabel.west);

\node[legend label] (fohlabel) at (10.5,4.8) {FOH};
\draw[foh] ([xshift=-16pt]fohlabel.west) -- ([xshift=-4pt]fohlabel.west);

\end{tikzpicture}%
    \caption{Control Parameterization for ZOH and FOH}
    \label{fig:control_parameterization}
\end{figure}
}

\subsubsection{Transcription}
\label{sec:lindisc}
\makeatletter
\edef\@currentlabel{\Roman{section}.\Alph{subsection}.\arabic{subsubsection}}
\label{sec:lindisc.dot}
\makeatother

To obtain a finite-dimensional representation of the \gls{ocp}, the control horizon is partitioned into $N-1$ intervals, problem is transcribed using a multiple shooting formulation and control is parameterized. The control parameterization step has been performed before; we now proceed with the remaining procedures of the transcription process.

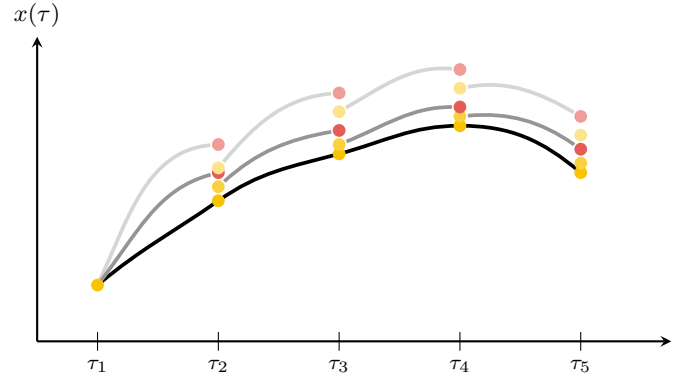
\begin{figure}[htbp]
    \centering
    % Units are chosen so the picture's natural width is one IEEEtran column
% (252pt); it is included at true size, without \resizebox.
\begin{tikzpicture}[>=stealth, x=0.799cm, y=0.62cm]

% Common styles for easy tuning. Lighter tints mark older iterations.
\tikzset{
    nodeYellow/.style={circle, fill=ttc-yellow, draw=none, inner sep=1.7pt},
    nodeYellowMid/.style={circle, fill=ttc-yellow!75!white, draw=none, inner sep=1.7pt},
    nodeYellowLight/.style={circle, fill=ttc-yellow!45!white, draw=none, inner sep=1.7pt},
    nodeRedMid/.style={circle, fill=ttc-red!75!white, draw=none, inner sep=1.7pt},
    nodeRedLight/.style={circle, fill=ttc-red!45!white, draw=none, inner sep=1.7pt},
    trajBlack/.style={line width=1.4pt, black},
    trajGray/.style={line width=1.4pt, ttc-gray, shorten >=3pt, shorten <=3pt},
    trajGrayLight/.style={line width=1.4pt, ttc-gray!40!white, shorten >=3pt, shorten <=3pt},
    axis/.style={thick, ->},
    axis label/.style={font=\small},
    tick label/.style={font=\footnotesize}
}

% Axes
\draw[axis] (0, 0) -- (10.5, 0);
\draw[axis] (0, 0) -- (0, 6.5) node[axis label, above] {$x(\tau)$};

% Ticks & Labels
\foreach \x/\i in {1/1, 3/2, 5/3, 7/4, 9/5} {
    \draw[thin] (\x, 0.2) -- (\x, -0.2) node[tick label, below] {$\tau_\i$};
}

% ----------------------------------------------------
% Trajectories
% ----------------------------------------------------

% 1. Continuous Black Trajectory (Fully Converged / Zero Defect)
\draw[trajBlack]
    (1, 1.2) to[out=40, in=215]
    (3, 3.0) to[out=35, in=195]
    (5, 4.0) to[out=15, in=180]
    (7, 4.6) to[out=0, in=140]
    (9, 3.6);

% 2. Oldest Iteration (Light Gray - large defects, highest above the final)
\draw[trajGrayLight, shorten <=0pt] (1, 1.2) to[out=65, in=185] (3, 4.2);
\draw[trajGrayLight]                (3, 3.7) to[out=50, in=185] (5, 5.3);
\draw[trajGrayLight]                (5, 4.9) to[out=30, in=175] (7, 5.8);
\draw[trajGrayLight]                (7, 5.4) to[out=10, in=150] (9, 4.8);

% 3. Middle Iteration (Dark Gray - smaller defects, squeezing downward)
\draw[trajGray, shorten <=0pt]      (1, 1.2) to[out=52, in=195] (3, 3.6);
\draw[trajGray]                     (3, 3.3) to[out=42, in=190] (5, 4.5);
\draw[trajGray]                     (5, 4.2) to[out=22, in=180] (7, 5.0);
\draw[trajGray]                     (7, 4.8) to[out=5,  in=145] (9, 4.1);

% ----------------------------------------------------
% Nodes (Discretization points / Defects)
% ----------------------------------------------------

% t_1 (Initial Condition - always fixed, all trajectories start here)
\node[nodeYellow] at (1, 1.2) {};

% t_2 Nodes
\node[nodeYellow]      at (3, 3.0) {}; % Converged guess / Black start
\node[nodeYellowMid]   at (3, 3.3) {}; % Mid iter guess
\node[nodeRedMid]      at (3, 3.6) {}; % Mid iter integrated end
\node[nodeYellowLight] at (3, 3.7) {}; % Old iter guess
\node[nodeRedLight]    at (3, 4.2) {}; % Old iter integrated end

% t_3 Nodes
\node[nodeYellow]      at (5, 4.0) {};
\node[nodeYellowMid]   at (5, 4.2) {};
\node[nodeRedMid]      at (5, 4.5) {};
\node[nodeYellowLight] at (5, 4.9) {};
\node[nodeRedLight]    at (5, 5.3) {};

% t_4 Nodes
\node[nodeYellow]      at (7, 4.6) {};
\node[nodeYellowMid]   at (7, 4.8) {};
\node[nodeRedMid]      at (7, 5.0) {};
\node[nodeYellowLight] at (7, 5.4) {};
\node[nodeRedLight]    at (7, 5.8) {};

% t_5 Nodes
\node[nodeYellow]      at (9, 3.6) {}; % Converged final state guess
\node[nodeYellowMid]   at (9, 3.8) {}; % Mid iter final state guess
\node[nodeRedMid]      at (9, 4.1) {}; % Mid iter integrated end
\node[nodeYellowLight] at (9, 4.4) {}; % Old iter final state guess
\node[nodeRedLight]    at (9, 4.8) {}; % Old iter integrated end

\end{tikzpicture}%
    \caption{Multiple shooting transcription. Early iterations (light gray) exhibit large defect gaps between integrated states (red) and the next state variables (yellow). As the algorithm converges, these defects are reduced, yielding a dynamically feasible trajectory (black).}
    \label{fig:multiple_shooting}
\end{figure}

In multiple shooting, the state at each knot is treated as an optimization variable. 
Dynamics are enforced by integrating the system over each interval, producing defect constraints that couple consecutive knots. 
The resulting nonlinear program is embedded in a \gls{scp} framework, where these defects are penalized and driven towards zero as the algorithm converges, as illustrated in \cref{fig:multiple_shooting}. 

The discretized dynamics and their linearizations can be obtained in two ways, which differ only in the order of differentiation and integration.

\paragraph{Discretize then Linearize}
The nonlinear dynamics are first integrated over each control interval to define a discrete-time flow map. 
This map is then differentiated to obtain the corresponding linearized dynamics.
In \openscvx, these derivatives are computed using automatic differentiation in \jax, propagated through the \textsf{Diffrax} integrator, avoiding finite-difference approximations. 
Two versions of this method are implemented in \openscvx, the first, \texttt{VectorizeDiscretizeLinearize} integrates all segments jointly and applies a single JVP pass for speed and the second, \texttt{DiscretizeLinearizeVectorize}, integrates and linearizes each segment separately for tighter per-segment error control on stiff or long-horizon problems.

\paragraph{Linearize then Discretize}
Alternatively, the continuous-time dynamics are first linearized to obtain time-varying Jacobians along the trajectory. 
These are augmented with the corresponding sensitivity equations and integrated over each interval to recover discrete-time state transition and control influence matrices.
This formulation corresponds to the variational equations and avoids repeated matrix inversions during discretization. 
A dense, \texttt{LinearizeDiscretize}, and sparse, \texttt{LinearizeDiscretizeSparse}, versions are both implemented in \openscvx.

\medskip

Both approaches yield equivalent discrete-time Jacobians after differentiation; the scheme is selected at problem assembly (\cref{lst:openscvx-api-options}).

We provide hereafter a mathematical description of the transcription step, assuming the control $\tilde{u}(\tau)$ has been parameterized with \gls{foh}, thus $\tilde{u}(\tau) = \tilde{u}(\tilde{u}_k, \tilde{u}_{k+1}), \text{for} \;\tau \in[\tau_k, \tau_{k+1})$.

Denote the flow of posterior states ${\tilde{x}}(\tau_k^+)$ from $\tau^+_{k}$ to $\tau^-_{k+1}$ with $\Phi(\tau^-_{k+1}, \tau^+_{k}; {\tilde{x}}(\tau_k^+), \tilde{u}_k, \tilde{u}_{k+1})$, which we define as follows. 
$$
\left.\begin{array}{l}
 \displaystyle \Phi(\tau^-_{k+1}, \tau^+_{k}; \tilde{x}(\tau_k^+), \tilde{u}_k, \tilde{u}_{k+1}) \coloneqq \tilde{x}(\tau_k^+) + \\ \displaystyle \qquad \qquad \qquad \qquad + \int_{\tau^+_{k}}^{\tau^-_{k+1}} \tilde{f}(\tilde{x}(\tau), \tilde{u}(\tau))d\tau
\end{array} \right| 
\;  k \in \mathcal{I}.
$$
Let $\Phi_k$ be the shorthand for $\Phi(\tau^-_{k+1}, \tau^+_{k}; {\tilde{x}}(\tau_k^+), \tilde{u}_k, \tilde{u}_{k+1})$. Transcription provides us with the reference flow $\bar{\Phi}_k \coloneqq \Phi(\tau^-_{k+1}, \tau^+_{k}; \bar{\tilde{x}}_k^+, \bar{\tilde{u}}_k, \bar{\tilde{u}}_{k+1})$. Chaining $\bar{\Phi}_k$ with the discrete-time dynamics $\tilde{f}_\text{d}$, we obtain the propagated reference $F_k$, defined as
$$
F_k \coloneqq \tilde{f}_\textrm{d}\left( \bar{\Phi}_k , \;\bar{\mu}_k \right),
\quad  k \in \mathcal{I}
$$
Jacobians with respect to posterior state, current control, next control, and impulsive control, respectively denoted by $A_k$, $B_k^-$, $B_k^+$, $C_k$, are computed by means of the chain rule, as follows:
\begin{align*}
    A_k &= \frac{\partial \tilde{f}_{\text{d}}}{\partial \tilde{x}}(\bar{\Phi}_k, \bar{\mu}_k) \frac{\partial \Phi_k}{\partial \tilde{x}}(\tau_{k+1}, \tau_k; \bar{\tilde{x}}_k^+, \bar{\tilde{u}}_k, \bar{\tilde{u}}_{k+1}), \\ 
    B_k^- &=  \frac{\partial \tilde{f}_{\text{d}}}{\partial \tilde{x}}(\bar{\Phi}_k, \bar{\mu}_k)\frac{\partial \Phi_k}{\partial \tilde{u}_k}(\tau_{k+1}, \tau_k; \bar{\tilde{x}}_k^+, \bar{\tilde{u}}_k, \bar{\tilde{u}}_{k+1}), \\
    B_k^+ &= \frac{\partial \tilde{f}_{\text{d}}}{\partial \tilde{x}}(\bar{\Phi}_k, \bar{\mu}_k)\frac{\partial \Phi_k}{\partial \tilde{u}_{k+1}}(\tau_{k+1}, \tau_k; \bar{\tilde{x}}_k^+, \bar{\tilde{u}}_k, \bar{\tilde{u}}_{k+1}), \\ 
    C_k &= \frac{\partial \tilde{f}_{\text{d}}}{\partial \mu}(\bar{\Phi}_k, \bar{\mu}_k),
\end{align*}

The linearized dynamics can be then written compactly as
\begin{multline*}
\tilde{x}_{k+1} =
A_k \Delta \tilde{x}^+_k + B_k^- \Delta \tilde{u}_k + B_k^+ \Delta \tilde{u}_{k+1} + \\
+ C_k \Delta \mu_{k+1} + F_k + \nu_k,
\end{multline*}
with slack $\nu_k$ absorbing discretization and linearization error. Here $\Delta\Box$ is used to denote the deviation of the solution variable $\Box$ from the reference solution, $\Delta\Box \coloneqq \Box - \bar{\Box}$. 

Let $z := \left[ \tilde{x}^{+\top} \;\; \tilde{u}^\top \;\; \mu^\top \right]^\top$ concatenate the state, continous controls and impulsive controls. Gather the convex inequality path and boundary constraints into $G_{\mathrm{cvx}}$, splitting it from the nonconvex inequality and affine equality $G$; similarly, split the affine equality path and boundary constraints $H_{\mathrm{aff}}$, from the nonaffine equality path and boundary constraints $H$. 
$$
G_{\mathrm{cvx}} \coloneqq \left[\begin{array}{c}
g_{\mathrm{cvx}} \\
P_{\mathrm{cvx}}
\end{array}\right] \quad G \coloneqq \left[\begin{array}{c}
g_{\mathrm{ncvx}} \\
P_{\mathrm{ncvx}}
\end{array}\right]
$$
$$
H_{\mathrm{aff}} \coloneqq \left[\begin{array}{c}
h_{\mathrm{aff}} \\
Q_{\mathrm{aff}}
\end{array}\right] \quad H \coloneqq \left[\begin{array}{c}
h_{\mathrm{naff}} \\
Q_{\mathrm{naff}}
\end{array}\right]
$$
Applying the transcription, linearizing the remaining nonconvex constraints and penalizing the linearized constraints provides the following surrogate problem.
As the user designates which states they wish to minimize or maximize in \cref{sec:objective}, the cost is recovered by indexing into
\(x(t_f)\).
Let
\begin{align*}
  \mathcal{I}_{\min}
  &=
  \bigl\{ i : x_i(t_f)\ \text{is Minimize} \bigr\},\\
  \mathcal{I}_{\max}
  &=
  \bigl\{ i : x_i(t_f)\ \text{is Maximize} \bigr\}.
  \label{eq:mayer-index}
\end{align*}
which are defined by the user when the states are created. Then
\begin{equation*}
  L(\tilde{x})
  =
  \sum_{i\in\mathcal{I}_{\min}}\lambda_i\,\tilde{x}_i(t_f)
  -
  \sum_{i\in\mathcal{I}_{\max}}\lambda_i\,\tilde{x}_i(t_f).
  \label{eq:mayer-subproblem}
\end{equation*}
\begin{mathnote}{Convex Subproblem}

\begin{equation}
    \label{eq:scp_subproblem}
    \begin{split}
        \min_{z, \nu, \nu_{\textrm{vb}}} \quad &
        L(\tilde{x})
        + \sum_i \Delta^\top z_i \mathbf{diag}(\wProx{i}) \Delta z_i\\
        & \qquad\qquad
        + \|\wVc \odot \nu\|_1
        + \wVbg^{\top} \max(0, \nu_G)\\
        & \qquad\qquad
        + \|\wVbh \odot \nu_H\|_1 \\
        \text{s.t.} \quad &
        \left.
        \begin{array}{l}
        \tilde{x}^{+}_{k+1}
        = F_k + A_k \Delta \tilde{x}^{+}_{k} + B^{-}_{k} \Delta \tilde{u}_k \\[0.5ex]
        \qquad\quad
        + B^{+}_{k} \Delta \tilde{u}_{k+1} + C_k \Delta \mu_{k+1} + \nu_k \\[0.5ex]
        y_{k+1} - y_k \le \varepsilon \, 1_{n_y}
        \end{array}
        \right|
        \forall k \in \mathcal{I}
        \\
        &
        \left.
        \begin{array}{l}
        G(\bar{z}_k,\theta) + \nabla_z G(\bar{z}_k,\theta)\Delta z_k = \nu_{G} \\[0.5ex]
        H(\bar{z}_k,\theta) + \nabla_z  H(\bar{z}_k,\theta)\Delta z_k = \nu_{H} \\[0.5ex]
        G_{\mathrm{cvx}}(z_k,\theta) \le 0 \\[0.5ex]
        H_{\mathrm{aff}}(z_k,\theta) = 0 \\[0.5ex]
        z_{\min} \le z_k \le z_{\max}
        \end{array}
        \right|
        \forall k \in \bar{\mathcal{I}}
    \end{split}
\end{equation}

where $\odot$ denotes the element-wise product, $\theta$ gathers fixed parameters.
\end{mathnote}

\subsubsection{Hyperparameter Update}
\label{sec:hyperparameter-update}
Both slacks and deviations are penalized in the objective function, thus requiring weight selection, an expert process referred to as \textit{tuning}. Choosing the right weights is important, but time-consuming; on the other hand, correctly updating weights can enable convergence even when weights are poorly chosen. \openscvx implements several weight-update policies and allows users to provide their own; a policy is selected at problem assembly (\cref{lst:openscvx-api-options}).

\paragraph{ConstantProximalWeight}
All weights in Prob.~\eqref{eq:scp_subproblem} are kept constant across iterations, and each iteration is accepted.

\paragraph{Adaptive Proximal Weight}
Based on how the proximal weight is updated in the Prox-linear method with Adaptive Penalty Updates in \cite[Section 4.5.3]{luo2025modeling}, at each \gls{scvx} iteration, candidate trajectories $z^{k+1}$ are evaluated against the true non-linear model using a composite non-linear metric $J_{\text{nonlin}}(z)$:
\begin{multline*}
    J_{\text{nonlin}}(z) = L + \sum_{k} \|\wVc \odot (x_{k+1} - x_{\text{prop}, k})\|_1 +\cr \sum_{k} \bigl( \wVbg^\top \max(0, G(z_k, \theta)) +\cr \|\wVbh \odot H(z_k, \theta)\|_1 \bigr),
\end{multline*}
where $x_{\text{prop}, k}$ is the forward state obtained by numerically integrating the non-linear continuous dynamics $F(\tilde{x}, \tilde{u})$ over $[\tau_k, \tau_{k+1}]$. This metric combines three intuitive terms:
\begin{enumerate}
    \item \textbf{Mayer Terminal Cost:} The un-linearized objective $J(z_N, \theta)$ evaluated at terminal node $N$.
    \item \textbf{Dynamics Mismatch:} The $l_1$-norm penalty on defect between decision states $x_{k+1}$ and forward-integrated states $x_{\text{prop}, k}$.
    \item \textbf{Constraint Violations:} The weighted $l_1$-penalties on non-linear inequality violations $\max(0, G_k)$ and equality residuals $|H_k|$.
\end{enumerate}
Natively convex constraints ($G_c \le 0$ and $H_c = 0$) are enforced directly as hard constraints within the Prob.~\eqref{eq:scp_subproblem}.
Because they suffer no linearization approximation error, candidate iterates satisfy them by construction without requiring virtual buffer slacks or penalty terms in $J_{\text{nonlin}}$.
To assess how well the convex subproblem predicted the true non-linear performance, \gls{scvx} computes the step acceptance ratio $\rho$:
\begin{equation}
    \rho = \frac{J_{\text{nonlin}}^k - J_{\text{nonlin}}^{k+1}}{J_{\text{nonlin}}^k - J_{\text{lin}}^{k+1}},
    \label{eq:rho_update}
\end{equation}
where $J_{\text{lin}}^{k+1}$ is the surrogate cost predicted by \cref{eq:scp_subproblem}.

\begin{algorithm}[h]
\caption{Adaptive Proximal Weight Autotuning}
\label{alg:adaptive_prox}
\begin{algorithmic}[1]

\STATE Solve convex subproblem to obtain candidate trajectory
$(\mathbf{x}^{k+1},\mathbf{u}^{k+1})$

\STATE Evaluate nonlinear objective
$J_{\mathrm{nonlin}}^{k+1}$

\IF{$k = 1$}
    \STATE Accept candidate trajectory
\ELSE
    \STATE Compute acceptance ratio $\rho$ (\cref{eq:rho_update})
    \IF{$\rho < \eta_0$}
        \STATE Reject candidate trajectory
        \STATE Increase $\lambda_{\mathrm{prox}}$
    \ELSIF{$\rho < \eta_1$}
        \STATE Accept candidate trajectory
        \STATE Increase $\lambda_{\mathrm{prox}}$
    \ELSIF{$\rho < \eta_2$}
        \STATE Accept candidate trajectory
        \STATE Leave $\lambda_{\mathrm{prox}}$ unchanged
    \ELSE
        \STATE Accept candidate trajectory
        \STATE Decrease $\lambda_{\mathrm{prox}}$
    \ENDIF
\ENDIF
\end{algorithmic}
\end{algorithm}

\paragraph{Augmented Lagrangian-inspired Autotuning}
To ensure exact penalization without manual weight tuning, the virtual control weights $\lambda_{\mathrm{vc}}$ and virtual buffer weights ($\lambda_{\mathrm{vb},G}$ for inequalities, $\lambda_{\mathrm{vb},H}$ for equalities) are updated dynamically using an augmented Lagrangian-inspired subgradient strategy (\cref{alg:augmented_lagrangian}). This is the linear penalty weight update scheme from \cite[Section 4.5.3]{luo2025modeling}.

Whenever a candidate iterate is accepted by \cref{alg:adaptive_prox}, the algorithm computes the true non-linear violations $\nu$ for dynamics integration mismatch ($\nu$), non-convex inequality violations ($\nu_G$), and non-convex equality residuals ($\nu_H$). All three penalty weight vectors share the identical update law, scaling the weight increase inversely with the current trust-region parameter $\lambda_{\mathrm{prox}}$ to prevent premature over-penalization early in the optimization process. Finally, each weight is capped at $\lambda_{\max}$ to maintain numerical stability in the convex subproblem.

\begin{algorithm}[h]
\caption{Augmented Lagrangian-inspired Autotuning}
\label{alg:augmented_lagrangian}
\begin{algorithmic}[1]

\STATE Execute \cref{alg:adaptive_prox}
to determine iterate acceptance and update
$\lambda_{\mathrm{prox}}$

\IF{candidate iterate accepted}

    \STATE Compute violations for dynamics defects, inequalities, and equalities:
    \begin{align*}
    \nu_{\mathrm{vc}} &= |\mathbf{x}_{k+1} - \mathbf{x}_{\mathrm{prop}}|,\\
    \nu_G &= \max(0, G(\mathbf{z}_k, \theta)), \\
    \nu_H &= |H(\mathbf{z}_k, \theta)|
    \end{align*}

    \FORALL{$(\lambda, \nu) \in \{(\lambda_{\mathrm{vc}}, \nu_{\mathrm{vc}}), (\lambda_{\mathrm{vb,G}}, \nu_G), (\lambda_{\mathrm{vb,H}}, \nu_H)\}$}

        \STATE Update virtual penalty weights
        \[
        \lambda \leftarrow
        \begin{cases}
        \lambda + \eta_\lambda \dfrac{\nu}{2\lambda_{\mathrm{prox}}}, & \nu > \epsilon, \\[8pt]
        \lambda + \eta_\lambda \dfrac{\nu^2}{2\epsilon\lambda_{\mathrm{prox}}}, & \nu \le \epsilon
        \end{cases}
        \]

        \STATE Clamp weight: $\lambda \leftarrow \min(\lambda, \lambda_{\max})$

    \ENDFOR

\ENDIF

\end{algorithmic}
\end{algorithm}

\subsubsection{Convergence Criterion}
\label{sec:convergence-criterion}

Regardless of the selected hyperparameter update strategy, \openscvx declares convergence when the candidate trajectory satisfies three conditions: (i) the trust-region step becomes sufficiently small, indicating that subsequent \gls{scp} iterations produce negligible changes to the solution, (ii) virtual control defects are sufficiently reduced, indicating dynamic feasibility, and (iii) virtual buffer violations are sufficiently reduced, indicating satisfaction of non-convex path constraints. Specifically, convergence is achieved when

\begin{equation*}
    \begin{aligned}
         \sum_i \Delta^\top z_i \mathbf{diag}(\wProx{i}) \Delta z_i &\leq \epsilon_{\mathrm{tr}},\\
        \|\wVc \odot \nu\|_1 &\leq \epsilon_{\mathrm{vc}},\\
        \|\wVbg \odot \nu_{\mathrm{G}}\|_1
        + \|\wVbh \odot \nu_{\mathrm{H}}\|_1
        &\leq \epsilon_{\mathrm{vb}},
    \end{aligned}
\end{equation*}

where $\epsilon_{\mathrm{tr}}$, $\epsilon_{\mathrm{vc}}$, and $\epsilon_{\mathrm{vb}}$ are user-defined convergence tolerances for the trust-region step, virtual control, and virtual buffer violations, respectively. The algorithm additionally terminates if the maximum number of \gls{scp} iterations $N_{\max}$ is reached. The overall \gls{scp} procedure is summarized in \cref{alg:scvx}.

\begin{algorithm}[h]
\caption{\openscvx}
\label{alg:scvx}
\begin{algorithmic}[1]

\STATE Initialize reference trajectory $z^0$, penalty weights, and trust-region parameters.

\FOR{$k=0,\ldots,N_{\max}-1$}

    \STATE Linearize and discretize about $z^k$.

    \STATE Solve the convex subproblem to obtain $z^{k+1}$,
    $\nu^{k+1}$, and $\nu_{\mathrm{vb}}^{k+1}$.

    \STATE Update trust-region parameters and penalty weights.

    \IF{
    $\sum_i (\Delta z_i^{k+1})^\top
    \mathbf{diag}(\wProx{i})\Delta z_i^{k+1}
    \leq \eta_{\mathrm{tr}}$,
    $\|\wVc \odot \nu^{k+1}\|_1 \leq \eta_{\mathrm{vc}}$,
    and
    $\|\wVbg \odot \nu_{\mathrm{G}}^{k+1}\|_1
    +\|\wVbh \odot \nu_{\mathrm{H}}^{k+1}\|_1
    \leq \eta_{\mathrm{vb}}$
    }
        \STATE Terminate.
    \ENDIF

\ENDFOR

\STATE Terminate: maximum iterations reached.

\end{algorithmic}
\end{algorithm}

\subsection{Modularity}
\label{sec:modularity}

\openscvx treats discretization, convex subproblem solvers, \gls{scp} algorithms, and autotuners as first-class, swappable modules rather than hard-coded parts of a monolithic solver. 
Each layer is defined by a lightweight abstract base class with a fixed lifecycle interface; a user implements that interface once and registers a custom component without touching the problem formulation or any other layer. 
The Problem object wires the chosen pieces together at setup time, so the same \gls{ocp} definition can be solved with different discretizers, solver backends, \gls{scp} variants, or weight-update rules—and a custom module developed for one problem transfers directly to another. 
This allows for \openscvx to be used for \gls{scvx} algorithm developement as well as a toolbox for solving \glspl{ocp}.

\section{Case Studies and Evaluation}
\label{sec:case_study}
To demonstrate the versatility of \openscvx, we consider a diverse collection of \glspl{ocp} spanning aerospace, robotics, autonomous vehicles, and hybrid systems. We first verify the accuracy of \openscvx by solving the Brachistochrone and Hohmann transfer problems, both of which have analytical solutions to compare against in \cref{sec:verification}. Then we compare \openscvx against other relevant trajectory optimization frameworks, \gpops, \maptor, \psopt, and \iclocs{} over the \glsentrylong{hs}, \glsentrylong{alv}, and \glsentrylong{rrv} from  \cite[Section 6]{Betts2010-vx} in \cref{sec:framework_comp}. We then demonstrate \openscvx's real-world feasibility by implementing a quadrotor racing, \gls{pdg} and logo tracing manipulator problem on real world systems in \cref{sec:exp_val}. A full suite of the provided reference problems, at the time of writing, can be found in \cref{apx:example_problems}.

\subsection{Verification Against Analytical Solutions}
\label{sec:verification}

To verify the correctness of \openscvx, we compare its solutions against
closed-form optima for two classical nonlinear \glspl{ocp}:
the Brachistochrone and the Hohmann transfer.
Rather than demonstrating convergence alone, these benchmarks verify that
the framework recovers trajectories whose objective values agree with the
analytical optimum to within numerical precision.
Beyond serving as validation examples for this paper, both problems are
included in the package's automated unit test and continuous integration
(CI) suite, ensuring that future code changes preserve numerical
correctness.

\subsubsection{Evaluation Metrics}

Solution accuracy is measured by the relative objective error with respect
to the analytical optimum.
For the brachistochrone, we report the relative final-time error
\[
\frac{|t_f - t_f^\star|}{t_f^\star},
\]
while for the Hohmann transfer we report the relative total impulse error
\[
\frac{\left|\sum_k \|\Delta v_k\| - \Delta v^\star\right|}{\Delta v^\star}.
\]

For completeness, we also report the wall-clock execution time of the three
major stages of an \openscvx workflow:
\texttt{problem.initialize()},
\texttt{problem.solve()}, and
\texttt{problem.post\_process()}.

\subsubsection{Brachistochrone}
\label{sec:verification-brachistochrone}

The brachistochrone problem seeks the curve of minimum travel time under
gravity between two fixed points in a vertical plane.
Its analytical solution is the cycloid,

\begin{align*}
x(\phi) &= x_0 + R(\phi - \sin\phi),\\
y(\phi) &= y_0 - R(1 - \cos\phi),
\end{align*}

with optimal travel time

\[
t_f^\star = \sqrt{\frac{R}{g}}\,\phi_f,
\]

where the radius $R$ and terminal parameter $\phi_f$ are determined by the
boundary conditions.

Within \openscvx, the problem is formulated as a free-final-time \gls{ocp} with state
$\mathbf{x} = [x,\, y,\, v]^\top$
and path-angle control $\theta$.
This benchmark verifies support for free-final-time optimization,
nonlinear continuous dynamics, and \gls{ctcs}.

\subsubsection{Hohmann Transfer}
\label{sec:verification-hohmann}

The second benchmark considers a planar LEO-to-GEO Hohmann transfer with
two impulsive maneuvers, following the formulation of
Weber~\cite{weber-hohmann}.
The spacecraft coasts under two-body dynamics between impulses, while the
transfer duration is fixed to the half-period of the transfer ellipse.
The objective minimizes the sum of the impulse magnitudes. The analytical optimal total impulse is
\begin{equation}
\Delta v^\star
=
|v_{\mathrm{LEO}} - v_{t,p}|
+
|v_{\mathrm{GEO}} - v_{t,a}|,
\end{equation}
where $v_{t,p}$ and $v_{t,a}$ denote the periapsis and apoapsis velocities
of the ellipse with semimajor axis equaling the halved sum of LEO and GEO radii.
Since $\Delta v$s are impulsive, this example verifies \openscvx's support for hybrid dynamics combining
continuous (time dilation in this case) with discrete impulsive controls.

\subsubsection{Results}

\Cref{tab:analytical-verification} summarizes the verification
results. For both benchmarks, \openscvx reproduces the analytical optimum with relative objective errors below $10^{-5}\%$, demonstrating that the framework accurately recovers known optimal solutions. Together, these examples exercise substantially different capabilities of the framework—including free-final-time optimization, continuous-time constraints, nonlinear dynamics, and impulsive controls—providing evidence that the underlying transcription and successive convexification implementation is numerically correct across multiple problem classes.

In addition to solution accuracy, the reported timing breakdown shows that solving each optimization problem requires only a small fraction of the total execution time, with initialization dominating the overall runtime. This reflects the one-time cost of constructing and compiling the problem, which can be amortized when solving the same problem repeatedly with different parameters.

Finally, these verification problems are executed as part of the package's automated unit testing and continuous integration (CI) pipeline. Comparing against analytical solutions enables the test suite to detect regressions in the optimization framework and ensures that future modifications continue to reproduce known optimal solutions within prescribed numerical tolerances.

\begin{table}[!h]
\centering
\caption{Verification of \openscvx against analytical optima for two
canonical nonlinear optimal control problems.}
\label{tab:analytical-verification}
\begin{tabular}{@{}lcccc@{}}
\toprule
Problem
  & Rel.\ error (\%)
  & Pos.\ RMSE
  & $t_{\mathrm{init}}$ (s)
  & $t_{\mathrm{solve}}$ (s)\\
\midrule
Brachistochrone
  & $6.65\times10^{-6}$
  & $1.01\times10^{-4}$ m
  & $1.614$
  & $0.053$\\
Hohmann transfer
  & $5.04\times10^{-7}$
  & $4.49\times10^{-2}$ km
  & $1.813$
  & $0.132$ \\
\bottomrule
\end{tabular}
\end{table}

\subsection{Framework Comparison}
\label{sec:framework_comp}

For a subset of benchmark problems, we compare \openscvx against the direct optimal control software packages \gpops{} \cite{Patterson2014GPOPII} and \iclocs{} \cite{Nie2018ICLOC2}, both featuring a MATLAB frontend, \psopt{} \cite{Becerra2010PSOPT}, a compiled C++ solver and \maptor{} \cite{maptor2025}, a solver using a Python frontend. All solvers except \openscvx employ third-party \gls{nlp} backends. \openscvx natively uses instead the second-order convex solver \textsf{QOCO} as backend \cite{Chari2026qoco}.  All methods are evaluated using a common set of metrics from trajectories obtained via forward simulation of the nonlinear system dynamics. 
\begin{enumerate*}[label=(\alph*), font=\itshape]
\item \textit{Constraint violation}, quantified with the following metrics:
(i) the maximum constraint violation over the trajectory, which captures worst-case infeasibility, and (ii) an aggregated violation measure that reflects both magnitude and duration of constraint violations over time.
\item \textit{Optimality}, evaluated by re-computing the objective function along with the simulated trajectory. This cost accounts for discrepancies introduced by discretization, linearization, and numerical integration in the optimization process.

\item \textit{Computational performance}, measured using total solve time from problem initialization to convergence. Average and standard deviation are estimated using 100 independent runs for each solver.
\end{enumerate*}

We then summarize ease of use of each package using different qualitative metrics: programming language, required user-specified information, back-end available solvers and main optimization solver mechanics.

\begin{figure}
\includegraphics[width=0.9\linewidth]{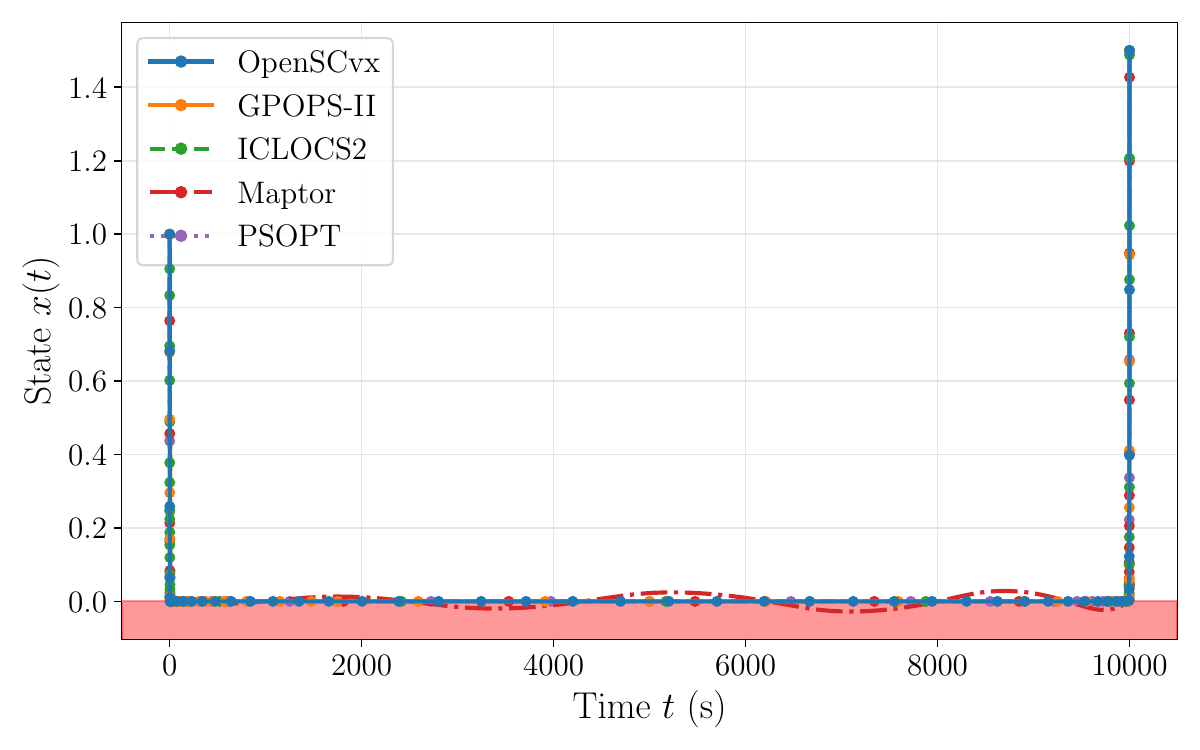}
\caption{Simulated optimal state -- \gls{hs}}
\label{fig:HS_state_trajectories}
\end{figure}

All examples are tested on a laptop equipped with Intel Core i9-14900HX, with 32 GB of installed RAM. For these examples we adopt \openscvx \texttt{0.6.0}, \iclocs{} \texttt{2.5.0}, \maptor{} \texttt{0.2.1} and \psopt{} release of \texttt{2026-06-26}.
\paragraph{\glsentrytitlecase{hs}{long} (\glsentryshort{hs})\glsunset{hs}}
Retrieved from \cite{Rao2000-dj}, \gls{hs} stresses a solver since the control magnitude of the optimal solution is larger near the domain boundaries than over the majority of the domain. Given scalar state $x$ and control $u$, dynamics read
$
\dot{x}(t) = -x^3(t) + u(t)
$
whereas the objective to minimize is
$
J = \frac{1}{2}\int_{0}^{t_f} \left( x^2(t) + u^2(t) \right) \text{d}t.
$
Boundary conditions and final time are fixed. The constraint $x\geq 0$ shall be valid over the whole domain. Further details can be found at \cite{Rao2000-dj}. A representation of the simulated solution from all solvers is reported in \cref{fig:HS_state_trajectories}.

As shown in \cref{tab:hypersensitive_computational_performance}, \openscvx ranks behind \gpops and shows superior performance with respect to \maptor{}, \iclocs{} and \psopt. All \gls{nlp} solvers are run with \ipopt{} backend \cite{Wachter2006ipopt}.
\begin{table}[h]
\centering
\begin{tabular}{lcc}
\toprule
Solver & Avg CPU (s) $\downarrow$ & Std CPU (s) $\downarrow$ \\
\midrule
\openscvx & 1.208 & 0.266 \\
\gpops{} & \textbf{0.751} & \textbf{0.014} \\
\iclocs{} & 1.898 & 0.025 \\
\maptor{} & 10.540 & 0.136 \\
\psopt{} & 17.138 & 0.807 \\
\bottomrule
\end{tabular}
\caption{Computational performance -- \gls{hs}}
\label{tab:hypersensitive_computational_performance}
\end{table}

 Furthermore, the reported objective is higher than all solvers. The lower computational and objective performance of \openscvx are a consequence of its fixed-node discretization: while the other solvers add nodes between iterations, \openscvx uses a fixed number of nodes, trading off simplicity and convergence guarantees \cite{Elango2025auto} for performance. On the other hand, deviation of the objective reported by the solver from the objective resulting from the simulation is the lowest, mirroring the high precision enabled by \openscvx. This is reported in the last column of \cref{tab:hypersensitive_objective_performance}.
 
\begin{figure}[h!]
\includegraphics[width=0.9\linewidth]{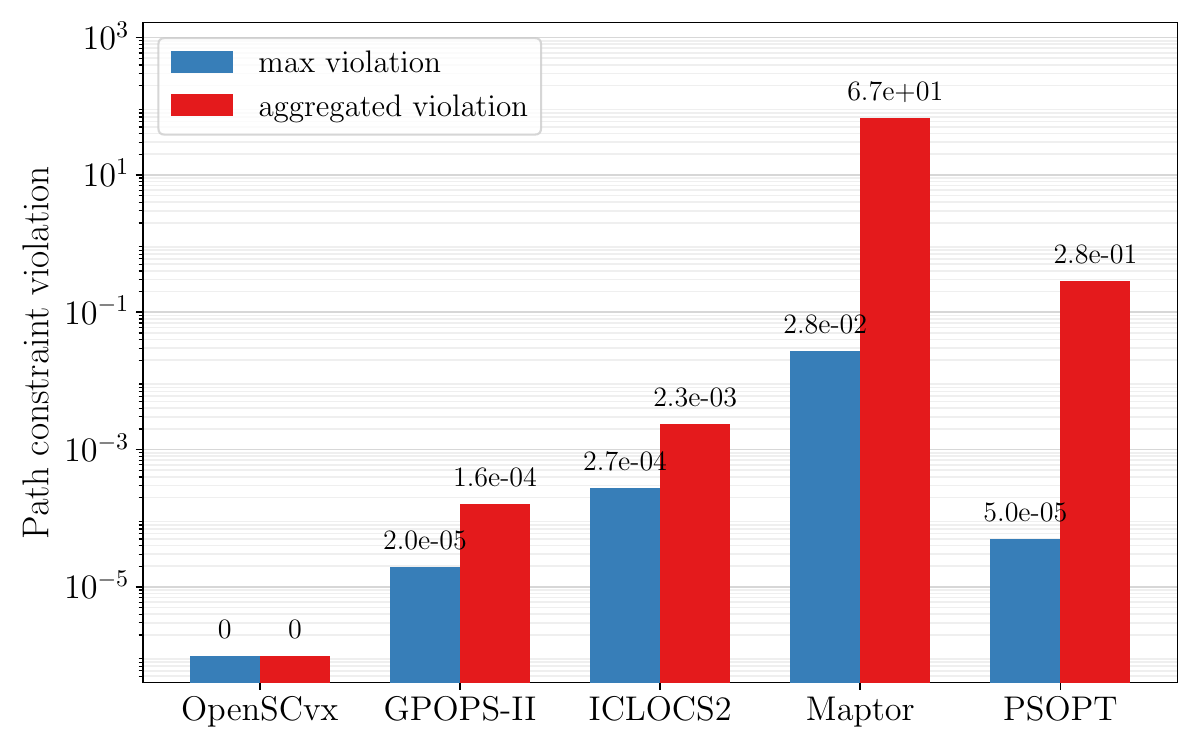}
\caption{Path constraint violations -- \gls{hs}}
\label{fig:HS_violation}
\end{figure}

\begin{table}[h]
\centering
\begin{tabular}{lccc}
\toprule
Solver & Reported $J$ $\downarrow$& Simulated $J$ $\downarrow$ & \% error $\downarrow$ \\
\midrule
\openscvx & 3.890 & 3.890 & $\approx$ \textbf{0.00}\% \\
\gpops{} & \textbf{3.362} & 3.351 & 0.33\% \\
\iclocs{} & \textbf{3.362} & \textbf{3.277} & 2.53\% \\
\maptor{} & \textbf{3.362} & 4.556 & 35.52\% \\
\psopt{} & \textbf{3.362} & 3.361 & 0.03\% \\
\bottomrule
\end{tabular}
\caption{Solver vs. simulated objective -- \gls{hs}}
\label{tab:hypersensitive_objective_performance}
\end{table}
Finally, the constraint reformulation used by \openscvx ensures that path constraint violation is essentially absent, independently of the considered number of nodes. This aspect, critical to avoid loss of the controlled system, is shown in \cref{fig:HS_violation}. This aspect is evident from \cref{fig:HS_state_trajectories}, where the only solution avoiding the forbidden zone in red is the solution of \openscvx. 

\begin{figure}[h!]
    \centering
    \includegraphics[width=\linewidth]{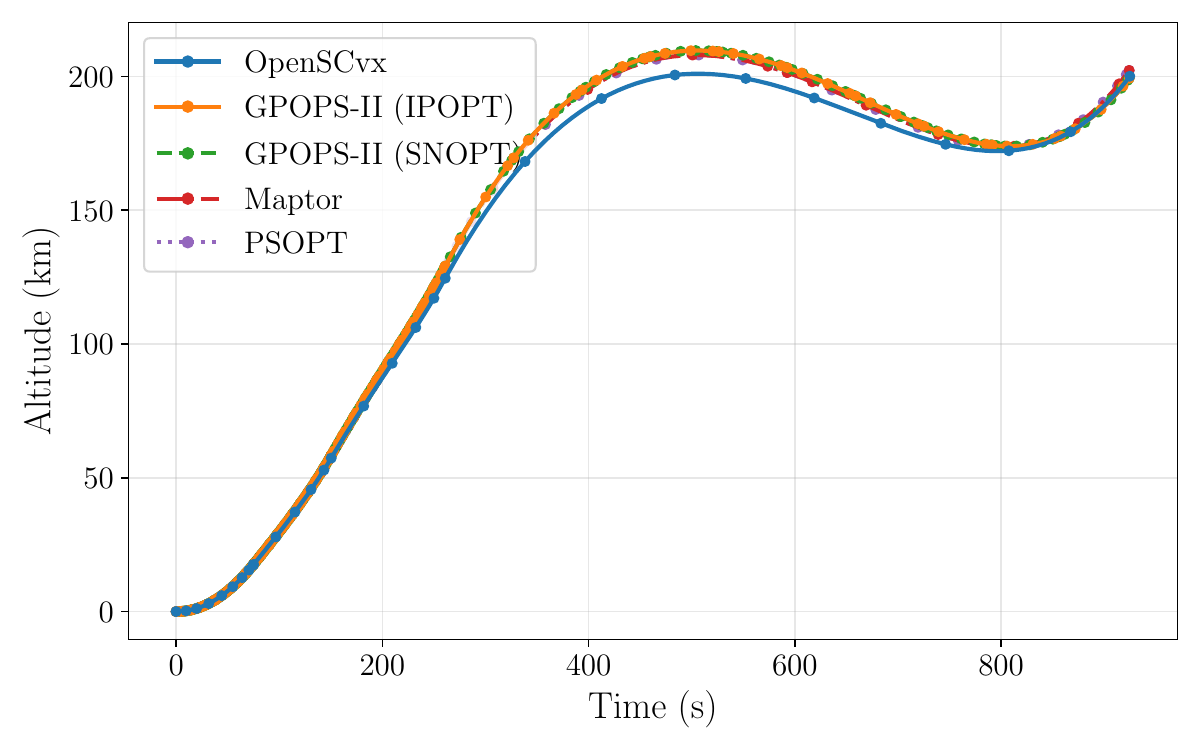}
    \caption{Simulated optimal altitude profile -- \gls{alv}}
    \label{fig:ALV_state_trajectories}
\end{figure}

\paragraph{\glsentrytitlecase{alv}{long} (\glsentryshort{alv})\glsunset{alv}}
This example, retrieved from \cite{Betts2010-vx}, optimizes a multi-stage vehicle ascent scenario with four different phases originating from three mass-depletion events. Given dynamics and thrust profile (interested reader is referred to \cite{Betts2010-vx}), the vehicle must optimize its steering angle to minimize its total expended mass, while injecting on an orbit with given orbital parameters $a,e,i,\Omega,\omega$. The final true anomaly is left free. Altitude profiles at convergence are depicted in \cref{fig:ALV_state_trajectories}, \iclocs{} being bypassed since it does not natively ship with this specific example.

Computational performance is summarized in \cref{tab:launch_computational_performance}. \openscvx sits behind the compiled \psopt{} solver, while overcoming the remaining solvers. All \gls{nlp} solvers are run with \ipopt{} backend, except \gpops. Notably, the open-source \ipopt{} interface of \gpops{} requires a much longer computational time with respect to the commercial \snopt{} interface, an aspect that we explain in the section about the ease of use.

\begin{table}[h!]
\centering
\begin{tabular}{lcc}
\toprule
Solver & Avg CPU (s) $\downarrow$ & Std CPU (s) $\downarrow$ \\
\midrule
\openscvx & 2.125 & 0.379 \\
\gpops{} (\ipopt) & 115.483 & 1.323 \\
\gpops{} (\snopt) & 8.442 & 0.161 \\
\maptor{} &  3.844 & \textbf{0.152} \\
\psopt{} & \textbf{1.681} & 0.302 \\
\bottomrule
\end{tabular}
\caption{Computational performance -- \gls{alv}}
\label{tab:launch_computational_performance}
\end{table}

The \gls{alv} problem does not feature any continuous-time path constraints, therefore we bypass the constraint satisfaction analysis. However, we report the precision of the simulated position and velocity profiles in \cref{tab:launch_integration_accuracy}: for this purpose, we reintegrate the state using the optimal control solution and compare the final state with the state provided by the solver. \openscvx sits on the same order of magnitude of the most precise solver in terms of position $r$, \gpops{} with \ipopt, and the best in terms of velocity $v$, \psopt, demonstrating that a simple first-order architecture can compete reliably against state-of-the-art \gls{nlp} solvers.

\begin{table}[h!]
\centering
\begin{tabular}{lcc}
\toprule
Solver & Final $r$ err.\ (km) $\downarrow$ & Final $v$ err.\ (m/s) $\downarrow$\\
\midrule
\openscvx & 1.67e-03 & 6.26e-03 \\
\gpops{} (\ipopt) & \textbf{7.03e-04} & 4.27e-01 \\
\gpops{} (\snopt) & 2.20e-02 & 5.76e-02 \\
\maptor{} & 3.77e-03 & 1.16e+00 \\
\psopt{} & 1.01e-03 & \textbf{1.95e-03}  \\
\bottomrule
\end{tabular}
\caption{Position/velocity integration accuracy -- \gls{alv}}
\label{tab:launch_integration_accuracy}
\end{table}

\gls{alv} problem boundary conditions are posed in terms of orbital parameters; we further report in \cref{tab:orbital_parameters_errors} the error with respect to the prescribed orbital parameters.
\begin{table}[t]
  \centering
  \small
  \setlength{\tabcolsep}{4pt}
  \begin{tabular}{l
      S[table-format=-1.3]
      S[table-format=-2.3]
      S[table-format=-2.4]
      S[table-format=-2.3]
      S[table-format=-2.4]}
  \toprule
   & {$|\Delta a|\,\downarrow$} & {$|\Delta e|\,\downarrow$} & {$|\Delta i|\,\downarrow$} & {$|\Delta\Omega|\,\downarrow$} & {$|\Delta\omega|\,\downarrow$} \\
  Solver & {(km)} & {($10^{-6}$)} & {($10^{-6}\,^\circ$)} & {($10^{-6}\,^\circ$)} & {($10^{-3}\,^\circ$)} \\
  \midrule
  OpenSCvx          &  0.212 &  2.42 & \bfseries 0.0022 & \bfseries 0.029 & 0.0402 \\
  \gpops{} (I.)  &  0.810 & 10.3  & 0.532            & 1.04           & 5.60   \\
  \gpops{} (S.)  & 0.213 & 2.71 & 22.1             & 94.7            & 0.601  \\
  \maptor{}            &  2.030 & 25.9  & 0.883            & 1.68           & 15.4   \\
  \psopt{}             & \bfseries 0.013 & \bfseries 0.113 & 0.289 & 2.57 & \bfseries 0.0219 \\
  \bottomrule
  \end{tabular}
  \caption{Terminal orbital-element absolute errors against the target orbit
  ($a=24361140$~m, $e=0.7308$, $i=28.5^\circ$, $\Omega=269.8^\circ$,
  $\omega=130.5^\circ$) -- \gls{alv}}
  \label{tab:orbital_parameters_errors}
  \end{table}

The objective, as entirely control-dependent, does not vary between solved solution and simulated solution. Performances are reported in \cref{tab:launch_objective_performance}, with \openscvx paying with respect to other solvers because of low amount of used nodes. This aspect can be explained from \cref{fig:ALV_state_trajectories}, which revels that the solution from \openscvx is slightly different with respect to the other solvers. 
\begin{table}[h]
\centering
\begin{tabular}{lc}
\toprule
Solver & $J$ (kg) $\uparrow$  \\
\midrule
\openscvx & 7514.45 \\
\gpops{} (\ipopt) & 7529.71 \\
\gpops{} (\snopt) & \textbf{7529.67} \\
\maptor{} & 7529.71 \\
\psopt{} & 7529.71 \\
\bottomrule
\end{tabular}
\caption{Solver/simulated objective -- \gls{alv}}
\label{tab:launch_objective_performance}
\end{table}

\begin{figure}[h!]
    \centering
    \includegraphics[width=1.0\linewidth]{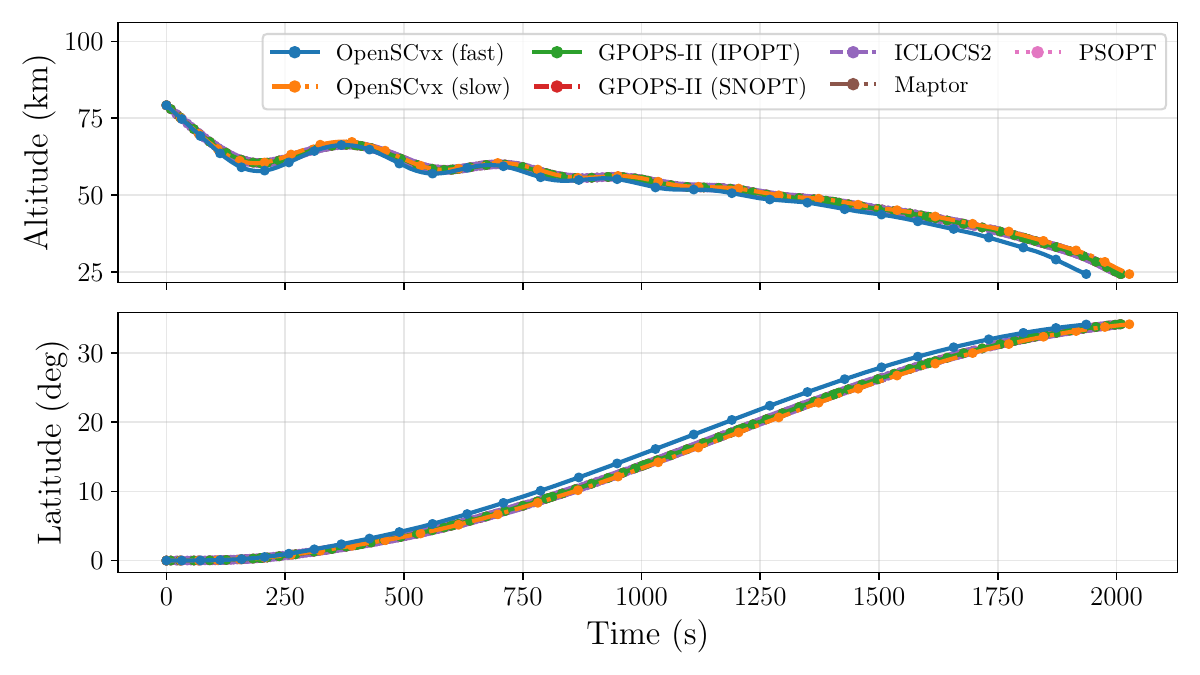}
    \caption{Simulated optimal altitude and latitude profiles - \gls{rrv}}
    \label{fig:RRV_state_trajectories}
\end{figure}

\paragraph{\glsentrytitlecase{rrv}{long} (\glsentryshort{rrv})\glsunset{rrv}} This example, retrieved from \cite{Betts2010-vx}, aims at maximizing the final latitude (i.e. the crossrange) of a Shuttle-like vehicle reentering Earth's atmosphere in the hypersonic regime. Dynamics and model description can be found in \cite{Betts2010-vx}. The high complexity of this scenario is caused by the sensitivity of dynamics to variations in control profiles, final time and initial conditions. For these reasons, we consider two tunings of \openscvx (\texttt{fast}, prioritizing computational time with aggressive tuning and \texttt{slow}, prioritizing optimality with looser tuning) and report both solutions. A depiction of the converged trajectories is reported in \cref{fig:RRV_state_trajectories}. 

The slow \openscvx tuning has a final time closer to the solutions of the other solvers; for this reason the trajectory has indeed higher optimality, as reported in \cref{tab:reentry_objective_performance}. On the other hand, the user interested in faster execution can privilege a fast tuning, providing lower computational time and trading them off for optimality, as reported in \cref{tab:reentry_computational_performance}. All \gls{nlp} solvers in \cref{tab:reentry_computational_performance} are run with \ipopt{} backend.

\begin{table}[h!]
\centering
\begin{tabular}{lccc}
\toprule
Solver & Reported $J$ ($^\circ$) $\uparrow$ & Simulated $J$ ($^\circ$) $\uparrow$ & \% error $\downarrow$ \\
\midrule
\openscvx (f.) & 34.1163 & 34.1160 & \textbf{0.00}\% \\
\openscvx (s.) & 34.1544 & 34.1544 & \textbf{0.00}\% \\
\gpops{} & \textbf{34.1641} & \textbf{34.1641} & \textbf{0.00}\% \\
\iclocs{} & 34.1422 & 34.1554 & 0.04\% \\
\maptor{} & \textbf{34.1641} & \textbf{34.1641} & \textbf{0.00}\% \\
\psopt{} & 34.1412 & 34.1521 & 0.03\% \\
\bottomrule
\end{tabular}
\caption{Solver vs. simulated objective -- \gls{rrv}}
\label{tab:reentry_objective_performance}
\end{table}

\begin{table}[h]
\centering
\begin{tabular}{lcc}
\toprule
Solver & Avg CPU (s) $\downarrow$ & Std CPU (s) $\downarrow$\\
\midrule
\openscvx (f.) & 1.869 & \textbf{0.027} \\
\openscvx (s.) & 2.314 & 0.114 \\
\gpops{} & \textbf{1.025} & 0.129 \\
\iclocs{} & 4.052 & 0.313 \\
\maptor{} & 4.197 & 0.278 \\
\psopt{} & 1.643 & 0.418 \\
\bottomrule
\end{tabular}
\caption{Computational performance -- \gls{rrv} }
\label{tab:reentry_computational_performance}
\end{table}

\subsubsection{Ease of use}
Each analyzed solver presents distinctive features, and \openscvx partially contrasts with them. All considered packages use second-order solvers, hence their \gls{nlp} backends require either exact or approximated Hessians of the constraints; \openscvx uses first-order information, requiring only Jacobians. \psopt, \textsf{ICLOCS} and \gpops{} adopt finite differences to compute Jacobians and Hessians; \maptor{} adopts a symbolic backend to evaluate Jacobians and Hessians based on \casadi, whereas \openscvx adopts \jax{} automatic differences to compute Jacobians up to machine precision.
\openscvx adopts a \gls{scp} scheme, and the adopted convex solver constitutes its backend. \openscvx therefore can adopt both \cvxpy-installed backends, using the \cvxpy{} constraints interface, and custom backends, as \textsf{QOCO} \cite{Chari2026qoco}, \textsf{moreau} \cite{moreau2026} or \textsf{qpax} \cite{Arrizabalaga2026qpax}, casting the solver matrices without help of a \gls{dsl}. The other considered solvers, instead, adopt a more general \gls{nlp} approach. \psopt{} and \gpops{} accept \ipopt/\snopt, \maptor{} accepts \ipopt, whereas \iclocs{} accepts \ipopt, MATLAB's fmincon and \textsf{WORHP} \cite{Buskens2012-ag}. Most notably, each \gls{nlp} backend requires in turn its own rules: for example, using the open-source \ipopt{} in place of the commercial \snopt{} for \gpops{} in the \gls{alv} example required a large overhead due to Hessian computation. The proposed \openscvx is written in Python and uses a \jax{} backend, in contrast with \maptor{} that is Python-based, but relies on \casadi. \iclocs{} and \gpops{} are written in MATLAB, whereas \psopt{} uses C++. \psopt{} is written in a natively compiled language, thus reliably offering high-end computational performance across all analyzed examples; \openscvx achieves similar performance by means of solver-dedicated \jax{} interfaces and batchable solves, but features a gentler learning curve thanks to use of an interpreted language. Similarly, \maptor{} adopts a \casadi{} interface, which makes its setup on a Python environment extremely flexible, but does not allow compilation of a full problem instance. \gpops{} and \iclocs{} use MEX functions to call the \gls{nlp} solvers, thus adding a layer of architectural complexity to the setup of the solver itself. 
At last, all \gls{nlp} solvers adopt mesh-refinement approaches to attempt ensuring constraint satisfaction in continuous time and improve solution precision. This requires multiple calls to the \gls{nlp} solvers, with problem structure that changes between each \gls{nlp} call. On the other hand, \openscvx adopts \gls{ctscvx}, ensuring constraint satisfaction in continuous time by design within the limits of \cref{eq:ctcs_relaxation}. 

\subsection{Experimental Demonstration}
\label{sec:exp_val}

To demonstrate the versatility of \openscvx beyond simulation benchmarks, we
validate the framework on representative hardware experiments spanning robotic
manipulation, autonomous flight, and powered descent. The three case studies
below exercise complementary modeling features---symbolic rigid-body kinematics,
parameterized nodal constraints, and compound \glspl{stc}---while
sharing the same modeling and solve pipeline.

\subsubsection{UR5e SVG Path Tracing}
\label{sec:exp_arm}
\begin{figure}[!t]
    \centering
    \subfloat[Real World Experiment]{\includegraphics[width=0.64\linewidth]{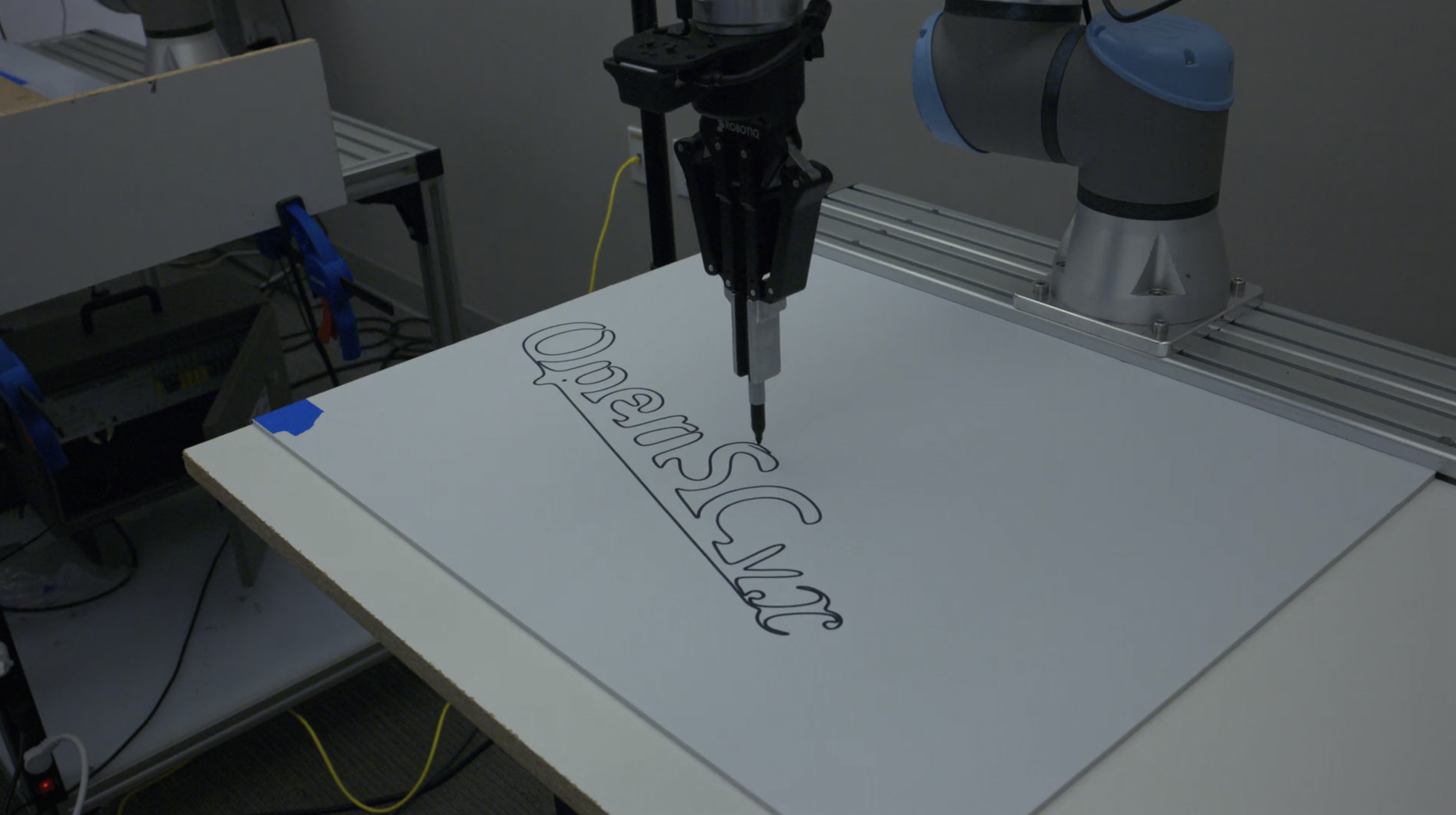}\label{fig:real_logo}}
    \hfil
    \subfloat[Simulation]{\includegraphics[width=0.35\linewidth]{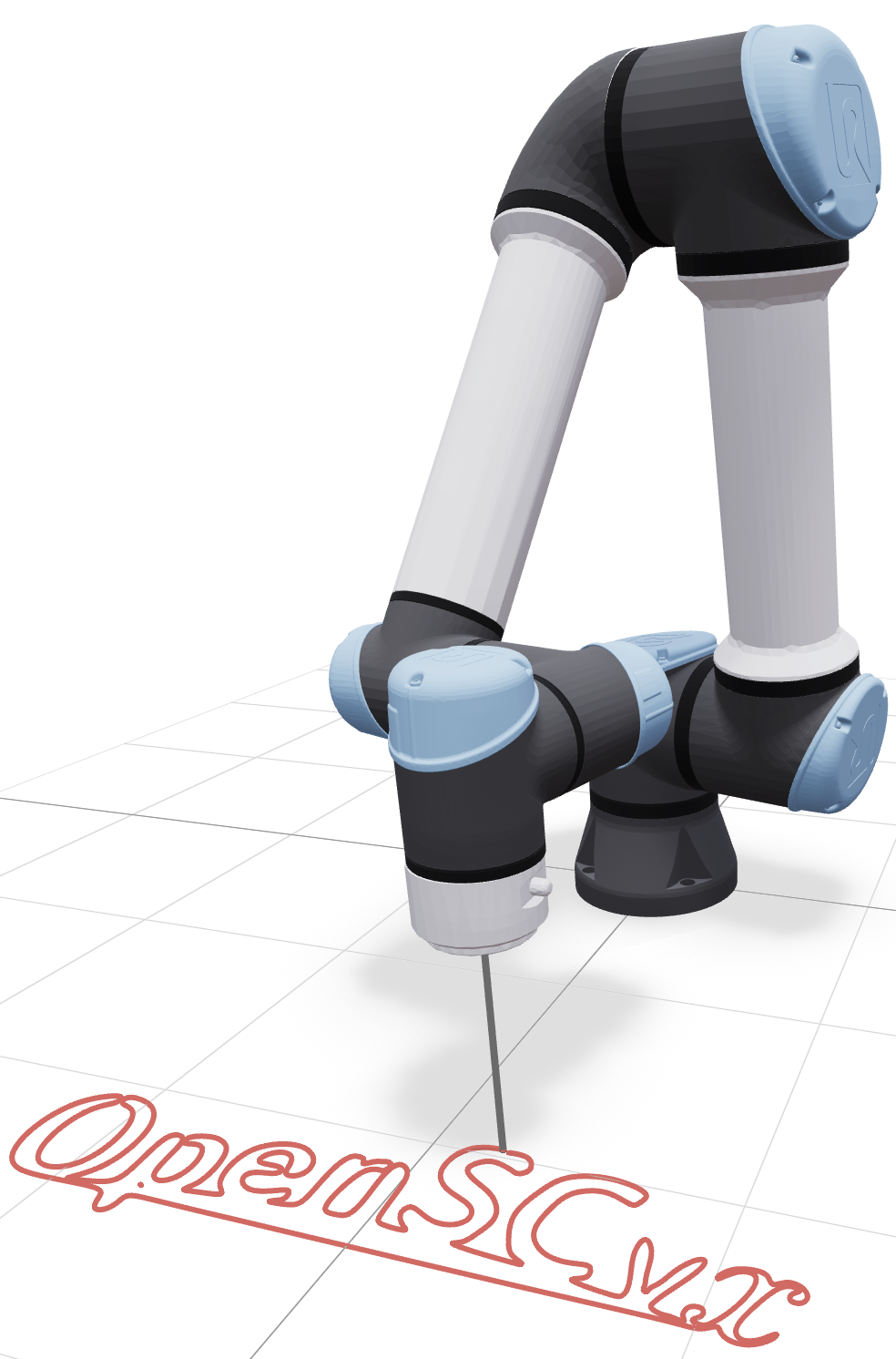}\label{fig:sim_logo}}
    \caption{UR5e pen-tracing experiment. Left: sequence composite of the hardware execution. Right: optimized simulation.}
    \label{fig:logo_Tracing}
\end{figure}

We first consider a seven-degree-of-freedom manipulator tasked with tracing a
prescribed SVG path with a pen tip mounted at the end effector
(\cref{fig:logo_Tracing}).
The plant is modeled with symbolic rigid-body dynamics and Product-of-Exponentials
forward kinematics, so the end-effector pose is an algebraic function of the joint
configuration inside the constraint graph.
Contact with the drawing surface is enforced continuously through \gls{ctcs}, keeping the
pen tip on the writing plane for the duration of the stroke rather than only at
discretization nodes.
The resulting trajectory is executed on a UR5e, with the hardware composite and
the corresponding nonlinear simulation shown side by side in
\cref{fig:logo_Tracing}.

\subsubsection{Time-Optimal Quadrotor Gate Racing}
\label{sec:exp_drone}
\begin{figure}[!t]
    \centering
    \includegraphics[width=\linewidth]{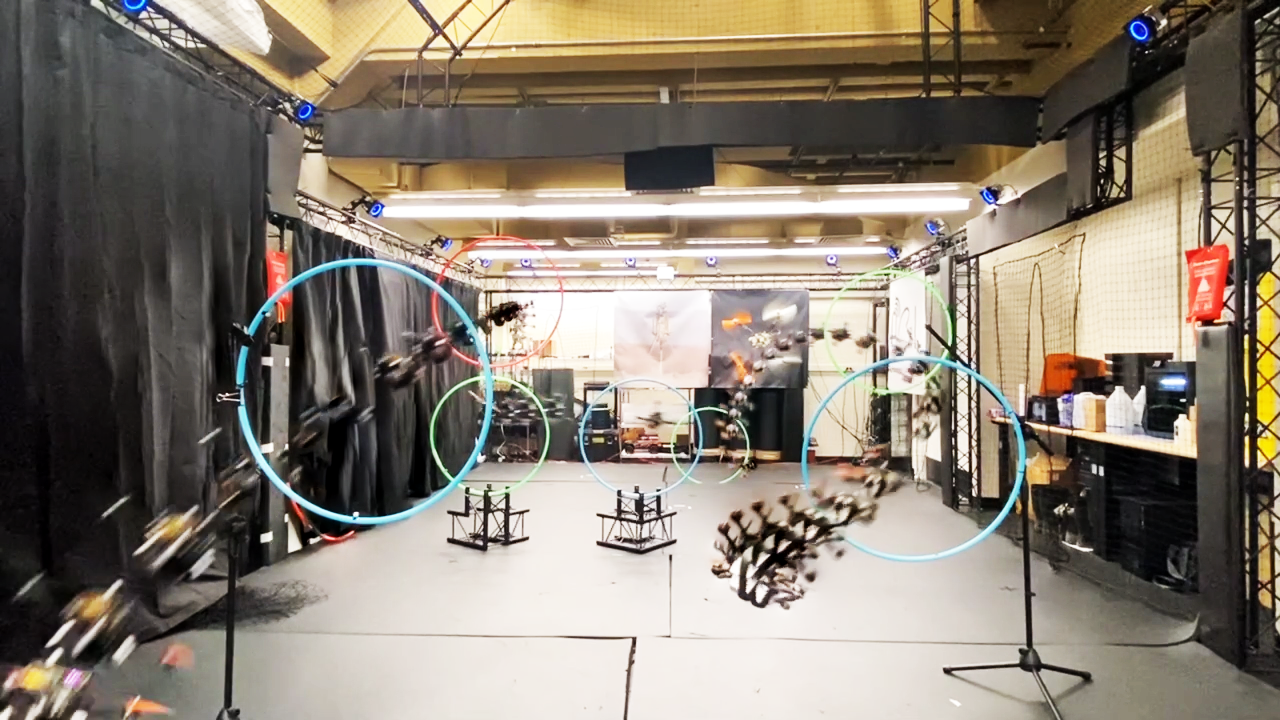}
    \caption{Time-optimal 6-DoF quadrotor racing trajectory.}
    \label{fig:drone_racing}
\end{figure}

The second experiment is a time-optimal six-degree-of-freedom quadrotor racing
problem (\cref{fig:drone_racing}).
The vehicle must traverse an ordered sequence of gates as quickly as possible
while respecting continuously enforced state and control limits.
Gate passage is encoded with nodal containment constraints, and the gate poses
themselves are declared as symbolic parameters.
Replanning for a new course layout therefore reduces to updating parameter values
and re-solving---without rebuilding or recompiling the problem---which is the
same mechanism used by the interactive realtime examples.
\Cref{fig:drone_racing} shows a representative minimum-time trajectory through
the gate sequence.

\subsubsection{6-DoF Powered Descent Guidance with Compound State-Triggered Constraints}
\label{sec:exp_pdg_stc}
To demonstrate \openscvx's support for temporal logic and compound \glspl{stc}, we consider a six-degree-of-freedom \gls{pdg} problem
for autonomous planetary landing.
While the vehicle is governed by the standard nonlinear rigid-body equations of
motion, the complexity of the problem arises from the large collection of
conditional flight rules that become active throughout different phases of the
descent.
These rules are represented symbolically within \openscvx using logical
implication, conjunction, and disjunction operators before being automatically
reformulated into continuous optimization constraints.

As the vehicle descends below an altitude of $200~\mathrm{m}$, the landing target
is required to remain within the sensor boresight throughout the remainder of the
descent,
\begin{equation*}
    (h < 200)
    \land
    (h > 2)
    \Longrightarrow
    \alpha_{\mathrm{LOS}}
    \le
    \alpha_{\max},
\end{equation*}
ensuring continuous visibility of the landing site during the terminal approach.

Below $100~\mathrm{m}$, the guidance system transitions into the terminal landing
phase, activating a collection of tighter flight constraints.
The engine gimbal deflection is restricted,
\begin{equation*}
    h < 100
    \Longrightarrow
    |\delta_e| \le 1^\circ,
\end{equation*}
while tighter vehicle tilt, angular-rate, and velocity limits are enforced,
\begin{align*}
    h < 100 &\Longrightarrow \theta \le \theta_{\max},\\
    h < 100 &\Longrightarrow \|\boldsymbol{\omega}\| \le \omega_{\max},\\
    h < 100 &\Longrightarrow \|\mathbf{v}\| \le v_{\max}.
\end{align*}
A glideslope corridor is also activated while the vehicle remains above the
landing surface,
\begin{equation*}
    (h < 100)
    \land
    (h > 2)
    \Longrightarrow
    \gamma
    \le
    \gamma_{\max}.
\end{equation*}

\begin{figure}[!t]
    \centering
    \subfloat[Translational Velocity \glsentryshort{stc}]{\includegraphics[width=0.48\linewidth]{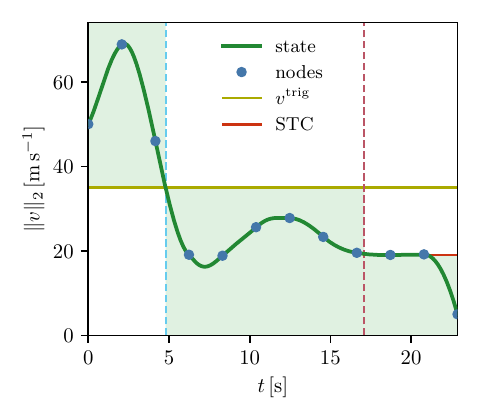}\label{fig:vel}}
    \hfil
    \subfloat[Angular Velocity \glsentryshort{stc}]{\includegraphics[width=0.48\linewidth]{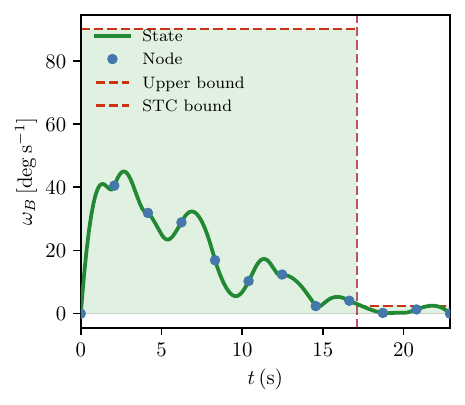}\label{fig:ang_vel}}
    \caption{Demonstration of implication-based compound state-triggered constraints in a 6-DoF powered descent guidance problem. The shaded areas denote the feasible regions. Note the intersample constraint satisfaction in \cref{fig:vel}.}
    \label{fig:stc}
\end{figure}

\begin{figure}[!t]
    \centering
    \includegraphics[width=1\linewidth]{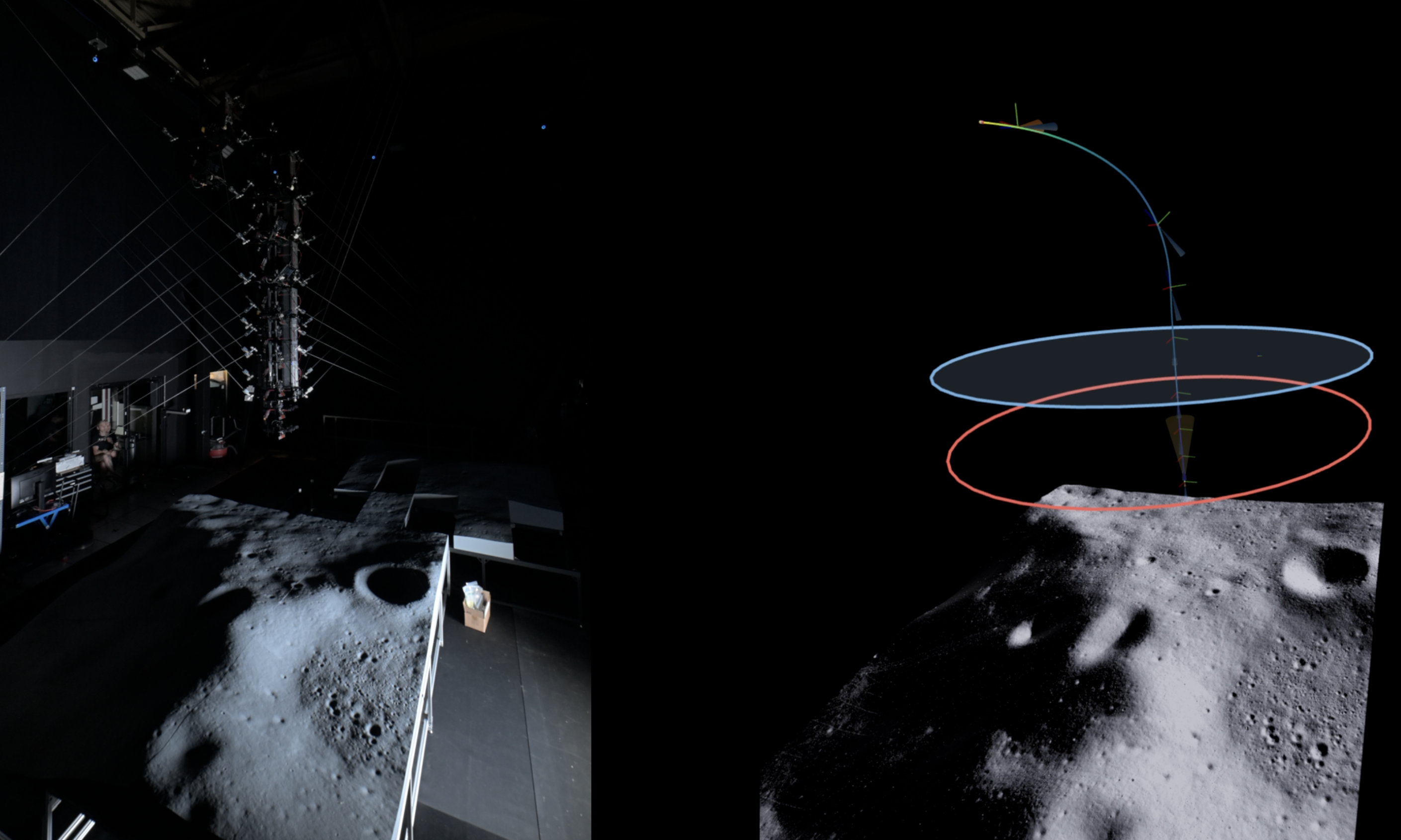}
    \caption{Real-world experiment of the 6-DoF compound-\glsentryshort{stc} problem. Left: NASA SENSS 6-DoF tendon robot executing the trajectory. Right: aligned nonlinear single-shot simulation. Both panels are sequence composites.}
    \label{fig:6dof_stc_exp}
\end{figure}

Finally, the propulsion system switches operating modes according to the vehicle
state.
When both the vehicle speed and tilt angle are sufficiently small, thrust is
restricted to the single-engine operating envelope,
\begin{equation*}
    (\|\mathbf{v}\| < 35)
    \land
    (\theta < 60^\circ)
    \Longrightarrow
    T_{\min}^{(1)}
    \le
    T
    \le
    T_{\max}^{(1)},
\end{equation*}
whereas exceeding either threshold activates the three-engine thrust limits,
\begin{equation*}
    (\|\mathbf{v}\| > 35)
    \lor
    (\theta > 60^\circ)
    \Longrightarrow
    T_{\min}^{(3)}
    \le
    T
    \le
    T_{\max}^{(3)}.
\end{equation*}
Collectively, these constraints demonstrate \openscvx's ability to express complex logical and temporal flight rules directly from their engineering specification, including nested implications, conjunctions, and disjunctions.
\Cref{fig:stc} shows two representative \glspl{stc} along the optimized descent, including continuous-time (intersample) satisfaction of the speed limit.
The optimized trajectory is then executed on the six-degree-of-freedom tendon actuated robot (STAR) \cite{Adams2023star} in the SENSS Lab at NASA JSC as a qualitative hardware demonstration (\cref{fig:6dof_stc_exp}), illustrating deployment of trajectories generated from high-level optimization models onto a real robotic system.

\section{Conclusion}
\openscvx provides an open-source framework for formulating, solving, and developing nonlinear trajectory optimization problems using continuous-time successive convexification. 
By separating problem specification from transcription, convexification, and numerical solution, the framework combines a high-level symbolic modeling interface with the flexibility needed to develop and evaluate new optimization algorithms. 
Its support for continuous-time constraints, temporal and logical specifications, automatic differentiation, parameterized problem updates, batched optimization, and interchangeable computational components enables a broad range of robotics and aerospace applications. 
The verification and case studies presented in this work demonstrate that this abstraction can recover analytical optima, achieve competitive performance with established trajectory optimization frameworks, and generate trajectories that can be executed on physical systems.
Beyond providing another trajectory optimization package, our broader goal is for \openscvx to serve as a common platform for research and deployment of successive-convexification methods. 
For researchers, the modular architecture is intended to make new discretization schemes, convexification strategies, solver interfaces, and algorithmic updates easier to implement, compare, and reproduce without rebuilding the surrounding optimization infrastructure. 
For practitioners, the same abstraction can shorten the path from a mathematical optimal-control formulation to a working implementation, facilitating the use of trajectory optimization in increasingly complex autonomous systems. 
In this sense, we envision \openscvx as infrastructure that can help close the gap between advances in trajectory optimization algorithms and their adoption in practical robotic and aerospace systems.

The current framework is a foundation rather than a finished ecosystem. 
Future development will focus on expanding the supported modeling and algorithmic capabilities, improving computational scalability and real-time performance, and broadening integrations with robotics, simulation, and embedded computing platforms. 
Equally important, continued growth of the open-source package and its benchmark and validation suite can provide a shared environment for the community to contribute new methods and evaluate them on common problems. 
We hope that \openscvx ultimately becomes not only a tool for solving trajectory optimization problems, but also a platform through which new trajectory optimization ideas can be developed, validated, and transitioned toward real-world autonomous systems.

\bibliographystyle{IEEEtran}
\bibliography{references}

@article{diamond2016cvxpy,
  author  = {Steven Diamond and Stephen Boyd},
  title   = {{CVXPY}: {A} {P}ython-embedded modeling language for convex optimization},
  journal = {Journal of Machine Learning Research},
  year    = {2016},
  volume  = {17},
  number  = {83},
  pages   = {1--5},
}

@article{agrawal2018rewriting,
  author  = {Agrawal, Akshay and Verschueren, Robin and Diamond, Steven and Boyd, Stephen},
  title   = {A rewriting system for convex optimization problems},
  journal = {Journal of Control and Decision},
  year    = {2018},
  volume  = {5},
  number  = {1},
  pages   = {42--60},
}

@article{uzun2026successive,
  title={Successive Convexification for Trajectory Optimization with Continuous-time Satisfaction of Signal Temporal Logic Specifications},
  author={Uzun, Samet and A{\c{c}}{\i}kme{\c{s}}e, Beh{\c{c}}et},
  journal={arXiv preprint arXiv:2606.06896},
  year={2026},
  eprint        = {2606.06896},
  archivePrefix = {arXiv},
  primaryClass  = {math.OC},
  url = {https://doi.org/10.48550/arXiv.2606.06896},
  doi = {10.48550/arXiv.2606.06896}
}

@article{uzun2024optimizationtemporallogicalspecifications,
  author = {Uzun, Samet and Elango, Purnanand and Garoche, Pierre-Loic and A{\c{c}}{\i}kme{\c{s}}e, Beh{\c{c}}et},
  title = {Optimization with Temporal and Logical Specifications via Generalized Mean-based Smooth Robustness Measures},
  journal = {arXiv preprint arXiv:2405.10996},
  year = {2024},
  doi = {10.48550/arxiv.2405.10996},
  url = {https://doi.org/10.48550/arXiv.2405.10996},
  eprint = {2405.10996},
  archiveprefix = {arXiv}
}

@inproceedings{uzun2025sequential,
  author = {Uzun, Samet and A{\c{c}}{\i}kme{\c{s}}e, Beh{\c{c}}et and Carson, John M},
  title = {Sequential convex programming for 6-DoF powered descent guidance with continuous-time compound state-triggered constraints},
  booktitle = {AIAA SCITECH 2025 Forum},
  pages = {1895},
  year = {2025},
  month = {jan},
  publisher = {American Institute of Aeronautics and Astronautics},
  doi = {10.2514/6.2025-1895},
  url = {https://doi.org/10.2514/6.2025-1895}
}

@article{uzun2025information_acquisition_ctscvx,
  author  = {Uzun, Samet and A{\c{c}}{\i}kme{\c{s}}e, Beh{\c{c}}et and {Di Cairano}, Stefano},
  title   = {Motion Planning for Information Acquisition via Continuous-Time Successive Convexification},
  journal = {IEEE Control Systems Letters},
  volume  = {9},
  pages   = {751--756},
  year    = {2025},
  doi     = {10.1109/LCSYS.2025.3576066}
}

@ARTICLE{Uzun2025proxconvex,
  title         = "A proximal method for composite optimization with smooth and
                   convex components",
  author        = "Uzun, Samet and Luo, Dayou and Açıkmeşe, Behçet and Aravkin,
                   Aleksandr Y",
  journal={arXiv preprint arXiv:2512.20602},
  month         =  dec,
  year          =  2025,
  archivePrefix = "arXiv",
  primaryClass  = "math.OC"
}

@book{luo2025modeling,
  title={Modeling and Algorithms for Nonconvex Trajectory Generation Problems From Constraint Reformulations and First-Order Proximal Methods to Structure-Exploiting Convex Solvers},
  author={Luo, Dayou},
  year={2025},
  publisher={University of Washington}
}

@ARTICLE{Amos2018-ex,
  title        = "{Differentiable MPC for end-to-end planning and control}",
  author       = "Amos, Brandon and Rodriguez, Ivan Dario Jimenez and Sacks,
                  Jacob and Boots, Byron and Zico Kolter, J",
  journal      = "arXiv [cs.LG]",
  month        =  "31~" # oct,
  year         =  2018,
  primaryClass = "cs.LG",
  doi          = "10.48550/arXiv.1810.13400"
}

@INPROCEEDINGS{Howell2019-we,
  title     = "{ALTRO: A fast solver for constrained trajectory optimization}",
  author    = "Howell, Taylor A and Jackson, Brian E and Manchester, Zachary",
  booktitle = "{2019 IEEE/RSJ International Conference on Intelligent Robots and
               Systems (IROS)}",
  publisher = "IEEE",
  month     =  nov,
  year      =  2019,
  doi       = "10.1109/iros40897.2019.8967788"
}

@INPROCEEDINGS{Mastalli2020-on,
  title     = "{Crocoddyl: An Efficient and Versatile Framework for Multi-Contact Optimal Control}",
  author    = "Mastalli, Carlos and Budhiraja, Rohan and Merkt, Wolfgang and
               Saurel, Guilhem and Hammoud, Bilal and Naveau, Maximilien and
               Carpentier, Justin and Righetti, Ludovic and Vijayakumar, Sethu
               and Mansard, Nicolas",
  booktitle = "{2020 IEEE International Conference on Robotics and Automation
               (ICRA)}",
  publisher = "IEEE",
  month     =  may,
  year      =  2020,
  doi       = "10.1109/icra40945.2020.9196673"
}

@ARTICLE{Jeon2025-iw,
  title     = "{CusADi: A GPU parallelization framework for symbolic expressions
               and optimal control}",
  author    = "Jeon, Se Hwan and Hong, Seungwoo and Lee, Ho Jae and Khazoom,
               Charles and Kim, Sangbae",
  journal   = "IEEE Robotics and Automation Letters",
  publisher = "Institute of Electrical and Electronics Engineers (IEEE)",
  volume    =  10,
  number    =  2,
  pages     = "899--906",
  month     =  feb,
  year      =  2025,
  doi       = "10.1109/lra.2024.3512254"
}

@ARTICLE{Jallet2025-lg,
  title     = "{ProxDDP: Proximal Constrained Trajectory Optimization}",
  author    = "Jallet, Wilson and Bambade, Antoine and Arlaud, Etienne and
               El-Kazdadi, Sarah and Mansard, Nicolas and Carpentier, Justin",
  journal   = "IEEE Transactions on Robotics",
  publisher = "Institute of Electrical and Electronics Engineers (IEEE)",
  volume    =  41,
  pages     = "2605--2624",
  year      =  2025,
  doi       = "10.1109/tro.2025.3554437"
}

@article{andersson2018casadi,
  title={CasADi—A software framework for nonlinear optimization and optimal control},
  author={Andersson, Joel and Gillis, Joris and Horn, Greg and Rawlings, Jim and Diehl, Moritz},
  journal={Mathematical Programming Computation},
  volume={11},
  number={1},
  pages={1--36},
  year={2018},
  publisher={Springer Verlag}
}

@software{jax2018github,
  author = {James Bradbury and Roy Frostig and Peter Hawkins and Matthew James Johnson and Yash Katariya and Chris Leary and Dougal Maclaurin and George Necula and Adam Paszke and Jake Vander{P}las and Skye Wanderman-{M}ilne and Qiao Zhang},
  title = {{JAX}: composable transformations of {P}ython+{N}um{P}y programs},
  url = {http://github.com/jax-ml/jax},
  version = {0.3.13},
  year = {2018},
}

@article{meuer2017sympy,
 title = {SymPy: symbolic computing in Python},
 author = {Meurer, Aaron and Smith, Christopher P. and Paprocki, Mateusz and \v{C}ert\'{i}k, Ond\v{r}ej and Kirpichev, Sergey B. and Rocklin, Matthew and Kumar, Amit and Ivanov, Sergiu and Moore, Jason K. and Singh, Sartaj and Rathnayake, Thilina and Vig, Sean and Granger, Brian E. and Muller, Richard P. and Bonazzi, Francesco and Gupta, Harsh and Vats, Shivam and Johansson, Fredrik and Pedregosa, Fabian and Curry, Matthew J. and Terrel, Andy R. and Rou\v{c}ka, \v{S}t\v{e}p\'{a}n and Saboo, Ashutosh and Fernando, Isuru and Kulal, Sumith and Cimrman, Robert and Scopatz, Anthony},
 year = 2017,
 month = Jan,
 volume = 3,
 pages = {e103},
 journal = {PeerJ Computer Science},
 issn = {2376-5992},
 url = {https://doi.org/10.7717/peerj-cs.103},
 doi = {10.7717/peerj-cs.103}
}

@misc{drake,
  author = "Russ Tedrake and the Drake Development Team",
  title = "Drake: Model-based design and verification for robotics",
  year = 2019,
  url = "https://drake.mit.edu"
}

@ARTICLE{Amatucci2026-rr,
  title     = "{Primal-dual iLQR for GPU-accelerated learning and control in
               legged robots}",
  author    = "Amatucci, Lorenzo and Sousa-Pinto, Jo\~{a}o and Turrisi, Giulio
               and Orban, Dominique and Barasuol, Victor and Semini, Claudio",
  journal   = "IEEE Robotics and Automation Letters",
  publisher = "Institute of Electrical and Electronics Engineers (IEEE)",
  volume    =  11,
  number    =  1,
  pages     = "1010--1017",
  month     =  jan,
  year      =  2026,
  doi       = "10.1109/lra.2025.3632610"
}

@software{maptor2025,
  title={MAPTOR: Multiphase Adaptive Trajectory Optimizer},
  author={Timothy, David},
  year={2025},
  url={https://github.com/maptor/maptor},
  version={0.2.1}
}

@INPROCEEDINGS{Becerra2010PSOPT,
  title     = "Solving complex optimal control problems at no cost with {PSOPT}",
  author    = "Becerra, Victor M",
  booktitle = "2010 IEEE International Symposium on Computer-Aided Control
               System Design",
  publisher = "IEEE",
  pages     = "1391--1396",
  month     =  sep,
  year      =  2010
}

@INPROCEEDINGS{Nie2018ICLOC2,
  title     = "{ICLOCS2}: Try this Optimal Control Problem Solver Before you Try
               the Rest",
  author    = "Nie, Yuanbo and Faqir, Omar and Kerrigan, Eric C",
  booktitle = "2018 UKACC 12th International Conference on Control (CONTROL)",
  publisher = "IEEE",
  pages     = "336--336",
  month     =  sep,
  year      =  2018
}

@ARTICLE{Agamawi2020CGPOP,
  title     = "{CGPOPS}",
  author    = "Agamawi, Yunus M and Rao, Anil V",
  journal   = "ACM Trans. Math. Softw.",
  publisher = "Association for Computing Machinery (ACM)",
  volume    =  46,
  number    =  3,
  pages     = "1--38",
  month     =  sep,
  year      =  2020,
  language  = "en"
}

@ARTICLE{Patterson2014GPOPII,
  title     = "Gpops-ii",
  author    = "Patterson, Michael A and Rao, Anil V",
  journal   = "ACM Trans. Math. Softw.",
  publisher = "Association for Computing Machinery (ACM)",
  volume    =  41,
  number    =  1,
  pages     = "1--37",
  month     =  oct,
  year      =  2014,
  language  = "en"
}

@INCOLLECTION{Byrd2006knitro,
  title     = "Knitro: An integrated package for nonlinear optimization",
  author    = "Byrd, Richard H and Nocedal, Jorge and Waltz, Richard A",
  booktitle = "Nonconvex Optimization and Its Applications",
  publisher = "Springer US",
  address   = "Boston, MA",
  pages     = "35--59",
  series    = "Nonconvex Optimization and Its Applications",
  year      =  2006
}

@ARTICLE{Wachter2006ipopt,
  title     = "On the implementation of an interior-point filter line-search
               algorithm for large-scale nonlinear programming",
  author    = "Wächter, Andreas and Biegler, Lorenz T",
  journal   = "Math. Program.",
  publisher = "Springer Science and Business Media LLC",
  volume    =  106,
  number    =  1,
  pages     = "25--57",
  month     =  mar,
  year      =  2006,
  language  = "en"
}

@ARTICLE{Gill2002snopt,
  title     = "{SNOPT}: An {SQP} algorithm for large-scale constrained
               optimization",
  author    = "Gill, Philip E and Murray, Walter and Saunders, Michael A",
  journal   = "SIAM J. Optim.",
  publisher = "Society for Industrial \& Applied Mathematics (SIAM)",
  volume    =  12,
  number    =  4,
  pages     = "979--1006",
  month     =  jan,
  year      =  2002
}

@ARTICLE{Verschueren2022acados,
  title     = "Acados—a modular open-source framework for fast embedded optimal
               control",
  author    = "Verschueren, Robin and Frison, Gianluca and Kouzoupis, Dimitris
               and Frey, Jonathan and van Duijkeren, Niels and Zanelli, Andrea
               and Novoselnik, Branimir and Albin, Thivaharan and Quirynen, Rien
               and Diehl, Moritz",
  journal   = "Math. Program. Comput.",
  publisher = "Springer Science and Business Media LLC",
  volume    =  14,
  number    =  1,
  pages     = "147--183",
  month     =  mar,
  year      =  2022,
  language  = "en"
}

@ARTICLE{Cederberg2026dnlp,
  title         = "Disciplined nonlinear programming",
  author        = "Cederberg, Daniel and Zhang, William and Nobel, Parth and
                   Boyd, Stephen",
  journal       = "arXiv [math.OC]",
  month         =  jun,
  year          =  2026,
  archivePrefix = "arXiv",
  primaryClass  = "math.OC"
}

@BOOK{David-H-Jacobson1970-ddp,
  title     = "Differential Dynamic {Programming–A} Unified Approach to the
               Optimization of Dynamic Systems",
  author    = "David H. Jacobson, David Q Mayne",
  publisher = "American Elsevier Publishing Company",
  year      =  1970
}

@INPROCEEDINGS{Todorov2005ilqr,
  title     = "A generalized iterative {LQG} method for locally-optimal feedback
               control of constrained nonlinear stochastic systems",
  author    = "Todorov, E and Li, Weiwei",
  booktitle = "Proceedings of the 2005, American Control Conference, 2005",
  publisher = "IEEE",
  pages     = "300--306 vol. 1",
  year      =  2005
}

@ARTICLE{Kavuncu2021-dpilqr,
  title         = "Potential {iLQR}: A potential-minimizing controller for
                   planning multi-agent interactive trajectories",
  author        = "Kavuncu, Talha and Yaraneri, Ayberk and Mehr, Negar",
  journal       = "arXiv [cs.RO]",
  month         =  jul,
  year          =  2021,
  archivePrefix = "arXiv",
  primaryClass  = "cs.RO"
}

@software{sun_lqrax_2025,
    author = {["Sun"], Max Muchen},
    license = {GPL-3.0},
    month = march,
    title = {{LQRax: JAX-enabled continuous-time LQR solver}},
    url = {https://github.com/MaxMSun/lqrax},
    version = {0.0.6},
    year = {2025}
}

@article{SCPToolboxCSM2022,
  doi = {10.1109/mcs.2022.3187542},
  url = {https://doi.org/10.1109/mcs.2022.3187542},
  year = {2022},
  month = oct,
  publisher = {Institute of Electrical and Electronics Engineers ({IEEE})},
  volume = {42},
  number = {5},
  pages = {40--113},
  author = {Danylo Malyuta and Taylor P. Reynolds and Michael Szmuk and Thomas Lew
            and Riccardo Bonalli and Marco Pavone and Behçet Açıkmeşe},
  title = {Convex Optimization for Trajectory Generation: A Tutorial on Generating
           Dynamically Feasible Trajectories Reliably and Efficiently},
  journal = {{IEEE} Control Systems},
  note = {Free preprint available at https://arxiv.org/abs/2106.09125}
}

@ARTICLE{Elango2025auto,
  title     = "Continuous-time successive convexification for constrained
               trajectory optimization",
  author    = "Elango, Purnanand and Luo, Dayou and Kamath, Abhinav G and Uzun,
               Samet and Kim, Taewan and Açıkmeşe, Behçet",
  journal   = "Automatica (Oxf.)",
  publisher = "Elsevier BV",
  volume    =  180,
  number    =  112464,
  pages     =  112464,
  month     =  oct,
  year      =  2025,
  language  = "en"
}

@ARTICLE{Elango2024ctscvx,
  title         = "Successive convexification for trajectory optimization with
                   continuous-time constraint satisfaction",
  author        = "Elango, Purnanand and Luo, Dayou and Kamath, Abhinav G and
                   Uzun, Samet and Kim, Taewan and Açıkmeşe, Behçet",
  journal       = "arXiv [math.OC]",
  month         =  apr,
  year          =  2024,
  archivePrefix = "arXiv",
  primaryClass  = "math.OC"
}

@ARTICLE{Oguri2023alscvx,
  author        = {Oguri, Kenshiro},
  title         = {Successive Convexification with Feasibility Guarantee via Augmented Lagrangian for Non-Convex Optimal Control Problems},
  booktitle     = {2023 62nd IEEE Conference on Decision and Control (CDC)},
  year          = {2023},
  pages         = {3296--3302},
  doi           = {10.1109/CDC49753.2023.10383462},
  eprint        = {2304.14564},
  archivePrefix = {arXiv},
  primaryClass  = {math.OC}
}

@INPROCEEDINGS{Mao2016-yr,
  title     = "Successive convexification of non-convex optimal control problems
               and its convergence properties",
  author    = "Mao, Yuanqi and Szmuk, Michael and Açıkmeşe, Behçet",
  booktitle = "2016 IEEE 55th Conference on Decision and Control (CDC)",
  publisher = "IEEE",
  pages     = "3636--3641",
  month     =  dec,
  year      =  2016
}

@ARTICLE{Mao2017-zj,
  title    = "Successive Convexification of Non-Convex Optimal Control Problems
              with State Constraints",
  author   = "Mao, Yuanqi and Dueri, Daniel and Szmuk, Michael and Açıkmeşe,
              Behçet",
  journal  = "IFAC-PapersOnLine",
  volume   =  50,
  number   =  1,
  pages    = "4063--4069",
  month    =  jul,
  year     =  2017
}

@inproceedings{spada2025impulsive_relative_ctcs,
  author        = {Spada, Fabio and Elango, Purnanand and A{\c{c}}{\i}kme{\c{s}}e, Beh{\c{c}}et},
  title         = {Impulsive Relative Motion Control with Continuous-Time Constraint Satisfaction for Cislunar Space Missions},
  booktitle     = {2025 American Control Conference (ACC)},
  year          = {2025},
  month         = jul,
  doi           = {10.23919/ACC63710.2025.11108021},
  eprint        = {2502.00215},
  archivePrefix = {arXiv},
  primaryClass  = {eess.SY}
}

@article{kumagai2024adaptive_mesh_scp_space,
  author  = {Kumagai, Naoya and Oguri, Kenshiro},
  title   = {Adaptive-Mesh Sequential Convex Programming for Space Trajectory Optimization},
  journal = {Journal of Guidance, Control, and Dynamics},
  volume  = {47},
  number  = {10},
  pages   = {2213--2220},
  year    = {2024},
  doi     = {10.2514/1.G008107}
}

@ARTICLE{Harris2020numpy,
  title     = "Array programming with {NumPy}",
  author    = "Harris, Charles R and Millman, K Jarrod and van der Walt, Stéfan
               J and Gommers, Ralf and Virtanen, Pauli and Cournapeau, David and
               Wieser, Eric and Taylor, Julian and Berg, Sebastian and Smith,
               Nathaniel J and Kern, Robert and Picus, Matti and Hoyer, Stephan
               and van Kerkwijk, Marten H and Brett, Matthew and Haldane, Allan
               and Del Río, Jaime Fernández and Wiebe, Mark and Peterson, Pearu
               and Gérard-Marchant, Pierre and Sheppard, Kevin and Reddy, Tyler
               and Weckesser, Warren and Abbasi, Hameer and Gohlke, Christoph
               and Oliphant, Travis E",
  journal   = "Nature",
  publisher = "Springer Science and Business Media LLC",
  volume    =  585,
  number    =  7825,
  pages     = "357--362",
  month     =  sep,
  year      =  2020,
  language  = "en"
}

@ARTICLE{Arrizabalaga2026qpax,
  title         = "A differentiable interior-point method in single precision",
  author        = "Arrizabalaga, Jon and Tracy, Kevin and Manchester, Zachary",
  journal       = "arXiv [math.OC]",
  month         =  may,
  year          =  2026,
  archivePrefix = "arXiv",
  primaryClass  = "math.OC"
}

@ARTICLE{schaller2022cvxpygen,
  author={Schaller, Maximilian and Banjac, Goran and Diamond, Steven and Agrawal, Akshay and Stellato, Bartolomeo and Boyd, Stephen},
  journal={IEEE Control Systems Letters}, 
  title={Embedded Code Generation With CVXPY}, 
  year={2022},
  volume={6},
  number={},
  pages={2653-2658},
  doi={10.1109/LCSYS.2022.3173209}}

@INPROCEEDINGS{Bonalli2019gusto,
  title     = "{GuSTO}: Guaranteed sequential trajectory optimization via
               sequential convex programming",
  author    = "Bonalli, Riccardo and Cauligi, Abhishek and Bylard, Andrew and
               Pavone, Marco",
  booktitle = "2019 International Conference on Robotics and Automation (ICRA)",
  publisher = "IEEE",
  pages     = "6741--6747",
  month     =  may,
  year      =  2019
}

@ARTICLE{Kamath2023-pq,
  title    = "Real-Time Sequential Conic Optimization for Multi-Phase Rocket
              Landing Guidance",
  author   = "Kamath, Abhinav G and Elango, Purnanand and Yu, Yue and Mceowen,
              Skye and Chari, Govind M and Carson, III, John M and Açıkmeşe,
              Behçet",
  journal  = "IFAC-PapersOnLine",
  volume   =  56,
  number   =  2,
  pages    = "3118--3125",
  month    =  jan,
  year     =  2023
}

@ARTICLE{Rao2000-dj,
  title     = "{Eigenvector approximate dichotomic basis method for solving
               hyper-sensitive optimal control problems}",
  author    = "Rao, Anil V and Mease, Kenneth D",
  journal   = "Optimal Control Applications \& Methods",
  publisher = "Wiley",
  volume    =  21,
  number    =  1,
  pages     = "1--19",
  month     =  jan,
  year      =  2000,
  doi       = "10.1002/(sici)1099-1514(200001/02)21:1<1::aid-oca646>3.0.co;2-v",
  language  = "en"
}

@BOOK{Betts2010-vx,
  title     = "{Practical methods for optimal control and estimation using
               nonlinear programming}",
  author    = "Betts, John T",
  publisher = "Society for Industrial and Applied Mathematics",
  month     =  "1~" # jan,
  year      =  2010,
  doi       = "10.1137/1.9780898718577"
}

@article{lavalle1998rrt,
author="Lavalle, S.",
title="Rapidly-exploring random trees : a new tool for path planning",
journal="Research Report 9811",
publisher="Department of Computer Science, Iowa State University",
year="1998",
URL="https://cir.nii.ac.jp/crid/1573950399665672960"
}

@ARTICLE{Karaman2011rrt*,
  title     = "Sampling-based algorithms for optimal motion planning",
  author    = "Karaman, Sertac and Frazzoli, Emilio",
  journal   = "Int. J. Rob. Res.",
  publisher = "SAGE Publications Ltd STM",
  volume    =  30,
  number    =  7,
  pages     = "846--894",
  month     =  jun,
  year      =  2011
}

@ARTICLE{Kavraki1996prm,
  title     = "Probabilistic roadmaps for path planning in high-dimensional
               configuration spaces",
  author    = "Kavraki, L E and Svestka, P and Latombe, J-C and Overmars, M H",
  journal   = "IEEE Trans. Rob. Autom.",
  publisher = "Institute of Electrical and Electronics Engineers (IEEE)",
  volume    =  12,
  number    =  4,
  pages     = "566--580",
  year      =  1996
}

@ARTICLE{Ross2010dagger,
  title         = "A reduction of imitation learning and structured prediction
                   to no-regret online learning",
  author        = "Ross, Stephane and Gordon, Geoffrey J and Andrew Bagnell, J",
  journal       = "arXiv [cs.LG]",
  month         =  nov,
  year          =  2010,
  archivePrefix = "arXiv",
  primaryClass  = "cs.LG"
}

@ARTICLE{Schulman2017ppo,
  title         = "Proximal Policy Optimization Algorithms",
  author        = "Schulman, John and Wolski, Filip and Dhariwal, Prafulla and
                   Radford, Alec and Klimov, Oleg",
  journal       = "arXiv [cs.LG]",
  month         =  jul,
  year          =  2017,
  archivePrefix = "arXiv",
  primaryClass  = "cs.LG"
}

@inproceedings{Konda1999actorcritic,
 author = {Konda, Vijay and Tsitsiklis, John},
 booktitle = {Advances in Neural Information Processing Systems},
 editor = {S. Solla and T. Leen and K. M\"{u}ller},
 pages = {},
 publisher = {MIT Press},
 title = {Actor-Critic Algorithms},
 url = {https://proceedings.neurips.cc/paper_files/paper/1999/file/6449f44a102fde848669bdd9eb6b76fa-Paper.pdf},
 volume = {12},
 year = {1999}
}

@ARTICLE{Janner2022diffusion,
  title         = "Planning with diffusion for flexible behavior synthesis",
  author        = "Janner, Michael and Du, Yilun and Tenenbaum, Joshua B and
                   Levine, Sergey",
  journal       = "arXiv [cs.LG]",
  month         =  may,
  year          =  2022,
  archivePrefix = "arXiv",
  primaryClass  = "cs.LG"
}

@ARTICLE{Ross2004pseudospectral,
  title     = "Pseudospectral knotting methods for solving nonsmooth optimal
               control problems",
  author    = "Ross, I Michael and Fahroo, Fariba",
  journal   = "J. Guid. Control Dyn.",
  publisher = "American Institute of Aeronautics and Astronautics (AIAA)",
  volume    =  27,
  number    =  3,
  pages     = "397--405",
  month     =  may,
  year      =  2004,
  language  = "en"
}

@BOOK{Boyd2004cvx,
  title     = "Convex Optimization",
  author    = "Boyd, Stephen and Vandenberghe, Lieven",
  publisher = "Cambridge University Press",
  address   = "Cambridge, England",
  month     =  mar,
  year      =  2004,
  language  = "en"
}

@ARTICLE{Chari2026qoco,
  title     = "{QOCO: a quadratic objective conic optimizer with custom solver
               generation}",
  author    = "Chari, Govind M and A\c{c}\i{}kme\c{s}e, Beh\c{c}et",
  journal   = "Mathematical Programming Computation",
  publisher = "Springer Science and Business Media LLC",
  year      =  2026,
  doi       = "10.1007/s12532-026-00311-8",
  language  = "en"
}

@BOOK{weber-hohmann,
  title     = "Attainability Heavenly Bodies ({NASA} Technical Translation
               {F}-44)",
  author    = "Hohmann, W",
  publisher = "National Aeronautics Space Administration",
  address   = "Washington, D.C.",
  year      =  1960
}

@inproceedings{yi2021iros,
    author={Brent Yi and Michelle Lee and Alina Kloss and Roberto Mart\'in-Mart\'in and Jeannette Bohg},
    title = {Differentiable Factor Graph Optimization for Learning Smoothers},
    year = 2021,
    BOOKTITLE = {2021 IEEE/RSJ International Conference on Intelligent Robots and Systems (IROS)}
}

@ARTICLE{Dunning2015-ut,
  title         = "{JuMP}: A modeling language for mathematical optimization",
  author        = "Dunning, Iain and Huchette, Joey and Lubin, Miles",
  journal       = "arXiv [math.OC]",
  month         =  aug,
  year          =  2015,
  archivePrefix = "arXiv",
  primaryClass  = "math.OC"
}

@software{moreau2026,
  author  = {Barratt, Shane and Nobel, Parth and Diamond, Steven},
  title   = {Moreau: {GPU}-Native Differentiable Optimization},
  year    = {2026},
  url     = {https://moreau.so},
}

@INCOLLECTION{Buskens2012-ag,
  title     = "{The ESA NLP Solver {WORHP}}",
  author    = "B{\"{u}}skens, Christof and Wassel, Dennis",
  booktitle = "{Springer}",
  publisher = "Springer",
  series    = "Springer Optimization and Its Applications",
  year      =  2012,
}

@BOOK{Pontryagin2018-fn,
  title     = "Mathematical Theory of Optimal Processes",
  author    = "Pontryagin, L S",
  publisher = "Routledge",
  year      =  1987
}

@BOOK{Bryson2018-sl,
  title     = "Applied Optimal Control: Optimization, Estimation, and Control",
  author    = "Bryson, Arthur E and Ho, Yu-Chi",
  publisher = "Routledge",
  year      =  1975
}

@ARTICLE{Hargraves1987-bk,
  title     = "Direct trajectory optimization using nonlinear programming and
               collocation",
  author    = "Hargraves, C R and Paris, S W",
  journal   = "J. Guid. Control Dyn.",
  publisher = "American Institute of Aeronautics and Astronautics (AIAA)",
  volume    =  10,
  number    =  4,
  pages     = "338--342",
  month     =  jul,
  year      =  1987,
  language  = "en"
}

@ARTICLE{Hicks1971single,
  title     = "Approximation methods for optimal control synthesis",
  author    = "Hicks, G A and Ray, W H",
  journal   = "Can. J. Chem. Eng.",
  publisher = "Wiley",
  volume    =  49,
  number    =  4,
  pages     = "522--528",
  month     =  aug,
  year      =  1971,
  language  = "en"
}

@ARTICLE{Bock1984multiple,
  title     = "A multiple shooting algorithm for direct solution of optimal
               control problems",
  author    = "Bock, H G and Plitt, K J",
  journal   = "IFAC Proc. Vol.",
  publisher = "Elsevier BV",
  volume    =  17,
  number    =  2,
  pages     = "1603--1608",
  month     =  jul,
  year      =  1984
}

@ARTICLE{Elnagar1995-fs,
  title     = "The pseudospectral Legendre method for discretizing optimal
               control problems",
  author    = "Elnagar, G and Kazemi, M A and Razzaghi, M",
  journal   = "IEEE Trans. Automat. Contr.",
  publisher = "Institute of Electrical and Electronics Engineers (IEEE)",
  volume    =  40,
  number    =  10,
  pages     = "1793--1796",
  year      =  1995
}

@ARTICLE{Kelly2017-zv,
  title     = "An introduction to trajectory optimization: How to do your own
               direct collocation",
  author    = "Kelly, Matthew",
  journal   = "SIAM Rev. Soc. Ind. Appl. Math.",
  publisher = "Society for Industrial \& Applied Mathematics (SIAM)",
  volume    =  59,
  number    =  4,
  pages     = "849--904",
  month     =  jan,
  year      =  2017,
  language  = "en"
}

@INCOLLECTION{Diehl2007-qf,
  title     = "Fast direct multiple shooting algorithms for optimal robot
               control",
  author    = "Diehl, M and Bock, H G and Diedam, H and Wieber, P-B",
  booktitle = "Lecture Notes in Control and Information Sciences",
  publisher = "Springer Berlin Heidelberg",
  address   = "Berlin, Heidelberg",
  pages     = "65--93",
  year      =  2007
}

@ARTICLE{Garg2010-yu,
  title     = "A unified framework for the numerical solution of optimal control
               problems using pseudospectral methods",
  author    = "Garg, Divya and Patterson, Michael and Hager, William W and Rao,
               Anil V and Benson, David A and Huntington, Geoffrey T",
  journal   = "Automatica (Oxf.)",
  publisher = "Elsevier BV",
  volume    =  46,
  number    =  11,
  pages     = "1843--1851",
  month     =  nov,
  year      =  2010,
  language  = "en"
}

@ARTICLE{Schulman2014-cm,
  title     = "Motion planning with sequential convex optimization and convex
               collision checking",
  author    = "Schulman, John and Duan, Yan and Ho, Jonathan and Lee, Alex and
               Awwal, Ibrahim and Bradlow, Henry and Pan, Jia and Patil, Sachin
               and Goldberg, Ken and Abbeel, Pieter",
  journal   = "Int. J. Rob. Res.",
  publisher = "SAGE Publications",
  volume    =  33,
  number    =  9,
  pages     = "1251--1270",
  month     =  aug,
  year      =  2014,
  language  = "en"
}

@ARTICLE{Mukadam2018-jl,
  title     = "Continuous-time Gaussian process motion planning via
               probabilistic inference",
  author    = "Mukadam, Mustafa and Dong, Jing and Yan, Xinyan and Dellaert,
               Frank and Boots, Byron",
  journal   = "Int. J. Rob. Res.",
  publisher = "SAGE Publications",
  volume    =  37,
  number    =  11,
  pages     = "1319--1340",
  month     =  sep,
  year      =  2018,
  language  = "en"
}

@INPROCEEDINGS{Kalakrishnan2011-cc,
  title     = "{STOMP}: Stochastic trajectory optimization for motion planning",
  author    = "Kalakrishnan, Mrinal and Chitta, Sachin and Theodorou, Evangelos
               and Pastor, Peter and Schaal, Stefan",
  booktitle = "2011 IEEE International Conference on Robotics and Automation",
  publisher = "IEEE",
  month     =  may,
  year      =  2011
}

@INPROCEEDINGS{Ratliff2009-cx,
  title     = "{CHOMP}: Gradient optimization techniques for efficient motion
               planning",
  author    = "Ratliff, Nathan and Zucker, Matt and Bagnell, J Andrew and
               Srinivasa, Siddhartha",
  booktitle = "2009 IEEE International Conference on Robotics and Automation",
  pages     = "489--494",
  month     =  may,
  year      =  2009
}

@ARTICLE{Janson2015-br,
  title     = "Fast Marching Tree: A fast marching sampling-based method for
               optimal motion planning in many dimensions",
  author    = "Janson, Lucas and Schmerling, Edward and Clark, Ashley and
               Pavone, Marco",
  journal   = "Int. J. Rob. Res.",
  publisher = "SAGE Publications",
  volume    =  34,
  number    =  7,
  pages     = "883--921",
  month     =  jun,
  year      =  2015,
  language  = "en"
}

@INPROCEEDINGS{kazukidiffusion,
  title     = "{CoBL}-diffusion: Diffusion-based conditional robot planning in
               dynamic environments using control barrier and lyapunov functions",
  author    = "Mizuta, Kazuki and Leung, Karen",
  booktitle = "2024 IEEE/RSJ International Conference on Intelligent Robots and
               Systems (IROS)",
  publisher = "IEEE",
  pages     = "13801--13808",
  month     =  oct,
  year      =  2024
}

@ARTICLE{Kapoor2025-mp,
  title     = "{STLCG++}: A masking approach for differentiable signal temporal
               logic specification",
  author    = "Kapoor, Parv and Mizuta, Kazuki and Kang, Eunsuk and Leung, Karen",
  journal   = "IEEE Robot. Autom. Lett.",
  publisher = "Institute of Electrical and Electronics Engineers (IEEE)",
  volume    =  10,
  number    =  9,
  pages     = "9240--9247",
  month     =  sep,
  year      =  2025
}

@INPROCEEDINGS{Adams2023star,
  title     = "Entry, descent, and landing {GN\&C} system evaluation via
               cable-driven emulation robotics",
  author    = "Adams, Davis W and Tse, Teming and Downs, Sean and Sostaric,
               Ronald R and Sooknanan, Joshua and Bankieris, Derek and O'Meara,
               Sarah and Peck, Caleb and Majji, Manoranjan and Sanchez, Hector L
               and Mohammadi, Ebrahim and Bunju Antoinette, Ringnyu and Walton,
               James and Owens, Chris",
  booktitle = "AIAA SCITECH 2023 Forum",
  publisher = "American Institute of Aeronautics and Astronautics",
  address   = "Reston, Virginia",
  month     =  jan,
  year      =  2023
}

@ARTICLE{Drusvyatskiy2019-sg,
  title     = "Efficiency of minimizing compositions of convex functions and
               smooth maps",
  author    = "Drusvyatskiy, D and Paquette, C",
  journal   = "Math. Program.",
  publisher = "Springer Science and Business Media LLC",
  volume    =  178,
  number    = "1-2",
  pages     = "503--558",
  month     =  nov,
  year      =  2019,
  language  = "en"
}

@INPROCEEDINGS{Mceowen2026autoscvx,
  title     = "Auto-tuned successive convexification for entry guidance with
               continuous-time constraint satisfaction",
  author    = "Mceowen, Skye and Morales, Carlos M and Johnson, Breanna J and
               Calderone, Daniel J and Cook, Edgerton M and Carson, John M and
               Acikmese, Behcet",
  booktitle = "AIAA SCITECH 2026 Forum",
  publisher = "American Institute of Aeronautics and Astronautics",
  address   = "Reston, Virginia",
  month     =  jan,
  year      =  2026
}

@ARTICLE{Hart1968a*,
  title     = "A formal basis for the heuristic determination of minimum cost
               paths",
  author    = "Hart, Peter and Nilsson, Nils and Raphael, Bertram",
  journal   = "IEEE Trans. Syst. Sci. Cybern.",
  publisher = "Institute of Electrical and Electronics Engineers (IEEE)",
  volume    =  4,
  number    =  2,
  pages     = "100--107",
  year      =  1968,
  language  = "en"
}

@ARTICLE{Stentz1994d*,
  title    = "Optimal and efficient path planning for partially-known
              environments",
  author   = "Stentz, Anthony",
  journal  = "IEEE Trans. Rob. Autom.",
  year     =  1994,
  language = "en"
}

@MISC{Dmitri2008hybrida*,
  title   = "Practical search techniques in path planning for autonomous driving",
  author  = "{Dmitri Dolgov, Sebastian Thrun, Michael Montemerlo, James Diebel}",
  journal = "AAAI",
  year    =  2008
}

@ARTICLE{Dijkstra1959-ib,
  title     = "A note on two problems in connexion with graphs",
  author    = "Dijkstra, E W",
  journal   = "Numer. Math. (Heidelb.)",
  publisher = "Springer Nature",
  volume    =  1,
  number    =  1,
  pages     = "269--271",
  month     =  dec,
  year      =  1959,
  language  = "en"
}

@PHDTHESIS{Kamath2025thesis,
  title    = "Real-Time Trajectory Optimization for High-Performance Guidance \&
              Control",
  author   = "Kamath, Abhinav Girish",
  editor   = "Açıkmeşe, Behçet",
  year     =  2025,
  school   = "University of Washington",
  language = "en"
}

\newpage

\begin{appendices}
\crefalias{section}{appendix}

\section{\openscvx Reference Problems}
\label{apx:example_problems}

% =============================================================================
% Catalog of OpenSCvx examples, grouped by folder, with the primary library
% features each example is intended to showcase.
%
% Requires: booktabs, tabularx, array
% Usage (appendix): \input{figures/examples_catalog}
%
% Two-column safe: non-floating tabularx blocks (no longtable, no \onecolumn).
% Names are wrapped with \expath (parbox + break after every character).
% =============================================================================

% Force-wrap monospace paths to the name-column width.
\makeatletter
\def\expath#1{%
  \parbox[t]{\linewidth}{%
    \ttfamily\footnotesize\raggedright
    \expandafter\expath@break\detokenize{#1}\relax
  }}
\def\expath@break#1{%
  \ifx\relax#1\else
    #1\hspace{0pt}%
    \expandafter\expath@break
  \fi}
\makeatother

\medskip
\noindent\textbf{Abstract}\\[0.35em]
\begingroup
\footnotesize
\renewcommand{\arraystretch}{1.15}
\begin{tabularx}{\linewidth}{@{} >{\raggedright\arraybackslash}p{0.46\linewidth} >{\raggedright\arraybackslash}X @{}}
\toprule
\textbf{\textrm{Example}} & \textbf{\textrm{OpenSCvx features showcased}} \\
\midrule
\expath{abstract/brachistochrone.py} & Minimal CTCS OCP; free-final-time brachistochrone \\
\expath{abstract/brachistochrone_batched.py} & solve\_batched; Qpax backend; batched ICs \\
\expath{abstract/chen_allgoewer.py} & Unstable OCP benchmark; CTCS \\
\expath{abstract/flappy_bird.py} & Impulsive controls; obstacle avoidance; Viser \\
\expath{abstract/hypersensitive.py} & Long-horizon hypersensitive OCP; CTCS \\
\expath{abstract/impulsive.py} & Mixed continuous / impulsive inputs \\
\expath{abstract/stl_integer_variable.py} & STL IntegerVariable (discrete-valued states) \\
\expath{abstract/stl_or.py} & STL Or (GMSR) operator \\
\addlinespace
\multicolumn{2}{@{}l@{}}{\textbf{\textrm{Aircraft}}} \\
\midrule
\expath{aircraft/dynamic_soaring.py} & Free-final-time aircraft OCP; CTCS \\
\expath{aircraft/supersonic_time_to_climb.py} & Minimum-time climb; free final time \\
\bottomrule
\end{tabularx}
\endgroup
\vspace{0.6\baselineskip}

\medskip
\noindent\textbf{Manipulator Arms}\\[0.35em]
\begingroup
\footnotesize
\renewcommand{\arraystretch}{1.15}
\begin{tabularx}{\linewidth}{@{} >{\raggedright\arraybackslash}p{0.46\linewidth} >{\raggedright\arraybackslash}X @{}}
\toprule
\expath{arm/3_dof_arm.py} & Lie-group PoE FK; Parameterized EE target \\
\expath{arm/7_dof_arm.py} & 7-DoF redundant arm; PoE FK; Viser \\
\expath{arm/7_dof_arm_collision.py} & Self-collision (vmap ellipsoids); PoE FK \\
\expath{arm/7_dof_arm_vp.py} & Wrist-camera viewcone; PoE FK \\
\expath{arm/franka_fr3v2_pick_place.py} & Franka pick-and-place; obstacle avoidance; Viser \\
\expath{arm/franka_fr3v2_viewplanning.py} & CTCS viewcone constraints; PoE FK \\
\expath{arm/franka_fr3v2_viewplanning_nodal.py} & Nodal (non-CTCS) viewcone constraints \\
\bottomrule
\end{tabularx}
\endgroup
\vspace{0.6\baselineskip}

\medskip
\noindent\textbf{Dubins Car}\\[0.35em]
\begingroup
\footnotesize
\renewcommand{\arraystretch}{1.15}
\begin{tabularx}{\linewidth}{@{} >{\raggedright\arraybackslash}p{0.46\linewidth} >{\raggedright\arraybackslash}X @{}}
\toprule
\expath{car/dubins_car.py} & CTCS obstacle avoidance; Parameters \\
\expath{car/dubins_car_disjoint.py} & Disjoint waypoint visiting \\
\expath{car/dubins_car_obstacle_conditional.py} & ox.Cond state-dependent speed limit \\
\expath{car/dubins_car_obstacle_stl.py} & STL conditional / temporal constraints \\
\expath{car/dubins_car_stl_or.py} & STL Or via stljax bridge \\
\expath{car/dubins_car_waypoint_stl.py} & STL Eventually waypoint in ball \\
\bottomrule
\end{tabularx}
\endgroup
\vspace{0.6\baselineskip}

\medskip
\noindent\textbf{Double Integrator}\\[0.35em]
\begingroup
\footnotesize
\renewcommand{\arraystretch}{1.15}
\begin{tabularx}{\linewidth}{@{} >{\raggedright\arraybackslash}p{0.46\linewidth} >{\raggedright\arraybackslash}X @{}}
\toprule
\expath{double_integrator/double_integrator_static.py} & Minimal CTCS box / terminal constraints \\
\expath{double_integrator/maze_rrt_scp.py} & Hybrid wavefront guess + SCP; Viser \\
\expath{double_integrator/moving_safe_zones.py} & STL moving safe zones; Parameters \\
\expath{double_integrator/multiphase_velocity.py} & Multiphase CTCS (.over node ranges) \\
\expath{double_integrator/obstacle_avoidance_vmap.py} & Vmap parallel obstacle constraints \\
\expath{double_integrator/obstacle_avoidance_vmap_2d.py} & Vmap ellipsoidal obstacles (2D) \\
\bottomrule
\end{tabularx}
\endgroup
\vspace{0.6\baselineskip}

\medskip
\noindent\textbf{Quadrotor}\\[0.35em]
\begingroup
\footnotesize
\renewcommand{\arraystretch}{1.15}
\begin{tabularx}{\linewidth}{@{} >{\raggedright\arraybackslash}p{0.46\linewidth} >{\raggedright\arraybackslash}X @{}}
\toprule
\expath{drone/cinema_vp.py} & CTCS continuous FOV / viewpoint constraints \\
\expath{drone/cinema_vp_blur_expr.py} & Symbolic motion-blur cost; viewpoint \\
\expath{drone/cinema_vp_ellipsoid.py} & Oriented ellipsoid in polytope FOV \\
\expath{drone/cinema_vp_ellipsoid_2norm.py} & Ellipsoid containment in 2-norm viewcone \\
\expath{drone/cinema_vp_ellipsoid_2norm_radii.py} & Eroding ellipsoid radii in viewcone \\
\expath{drone/cinema_vp_ellipsoid_2norm_radii_min.py} & Minimax eroding ellipsoid radii \\
\expath{drone/cinema_vp_nodal.py} & Nodal (non-CTCS) viewpoint constraints \\
\expath{drone/cinema_vp_occlusion_corridor.py} & LoS occlusion through corridor geometry \\
\expath{drone/cinema_vp_occlusion_shapes.py} & Mixed-shape LoS occlusion \\
\expath{drone/cinema_vp_polytope.py} & Subject polytope in FOV \\
\expath{drone/dr_vp.py} & Drone racing + CTCS multi-target LoS; vmap \\
\expath{drone/dr_vp_nodal.py} & Drone racing + nodal LoS constraints \\
\expath{drone/dr_vp_polytope.py} & Polytope target arrangement; CTCS LoS \\
\expath{drone/drone_racing.py} & Sequential gate racing; Parameters \\
\expath{drone/drone_racing_batched_gates.py} & solve\_batched over gate layouts \\
\expath{drone/logo.py} & Path-tracing (ACL logo); viewpoint \\
\expath{drone/lunar_terrain_agl.py} & DEM / AGL terrain following \\
\expath{drone/maze_scp.py} & Wavefront guess + SCP maze; Viser \\
\expath{drone/moving_safe_columns.py} & STL moving safe columns; Parameters \\
\expath{drone/obstacle_avoidance.py} & CTCS ellipsoidal obstacles; Parameters \\
\expath{drone/obstacle_avoidance_nodal.py} & Nodal ellipsoidal obstacle constraints \\
\expath{drone/obstacle_avoidance_vmap_2d.py} & Vmap parallel obstacle constraints \\
\expath{drone/openscvx_logo.py} & Path-tracing (OpenSCvx logo); Parameters \\
\expath{drone/quadrotor_hector.py} & Simple quadrotor dynamics port \\
\expath{drone/quadrotor_zigzag.py} & Waypoint tracking benchmark port \\
\bottomrule
\end{tabularx}
\endgroup
\vspace{0.6\baselineskip}

\medskip
\noindent\textbf{Frax Robot Dynamics}\\[0.35em]
\begingroup
\footnotesize
\renewcommand{\arraystretch}{1.15}
\begin{tabularx}{\linewidth}{@{} >{\raggedright\arraybackslash}p{0.46\linewidth} >{\raggedright\arraybackslash}X @{}}
\toprule
\expath{frax/panda_frax.py} & FRAX rigid-body dynamics integration \\
\expath{frax/panda_frax_pick_place.py} & FRAX dynamics; collision avoidance \\
\expath{frax/panda_frax_viewplanning.py} & FRAX dynamics; wrist-camera viewcone \\
\expath{frax/panda_frax_waypoint.py} & FRAX dynamics; Parameterized waypoint \\
\bottomrule
\end{tabularx}
\endgroup
\vspace{0.6\baselineskip}

\medskip
\noindent\textbf{MuJoCo MJX}\\[0.35em]
\begingroup
\footnotesize
\renewcommand{\arraystretch}{1.15}
\begin{tabularx}{\linewidth}{@{} >{\raggedright\arraybackslash}p{0.46\linewidth} >{\raggedright\arraybackslash}X @{}}
\toprule
\expath{mjx/cartpole_mjx.py} & MJX dynamics adapter; swing-up \\
\expath{mjx/double_cartpole_mjx.py} & MJX dynamics; double cartpole \\
\expath{mjx/skydio_x2_mjx.py} & MJX quadrotor; gate racing \\
\expath{mjx/triple_cartpole_3d_mjx.py} & MJX dynamics; 3D triple cartpole \\
\expath{mjx/triple_cartpole_game.py} & Interactive MJX balancing game \\
\expath{mjx/triple_cartpole_mjx.py} & MJX dynamics; triple cartpole \\
\bottomrule
\end{tabularx}
\endgroup
\vspace{0.6\baselineskip}

\medskip
\noindent\textbf{MPC / MPCC}\\[0.35em]
\begingroup
\footnotesize
\renewcommand{\arraystretch}{1.15}
\begin{tabularx}{\linewidth}{@{} >{\raggedright\arraybackslash}p{0.46\linewidth} >{\raggedright\arraybackslash}X @{}}
\toprule
\expath{mpc/double_integrator_discrete.py} & MPCC path following; discrete MPC loop \\
\expath{mpc/double_integrator_drone_racing.py} & MPCC gate racing; receding horizon \\
\expath{mpc/dubins_car_circle_analytical.py} & MPCC with analytical reference \\
\expath{mpc/dubins_car_circle_discrete.py} & MPCC discrete circle tracking \\
\expath{mpc/realtime_double_integrator_drone_racing.py} & Realtime Parameter updates; MPCC; Viser \\
\expath{mpc/realtime_quadrotor_viewpoint_mpcc.py} & Realtime keypoint Parameter; viewpoint MPC \\
\bottomrule
\end{tabularx}
\endgroup
\vspace{0.6\baselineskip}

\medskip
\noindent\textbf{Multi-agent}\\[0.35em]
\begingroup
\footnotesize
\renewcommand{\arraystretch}{1.15}
\begin{tabularx}{\linewidth}{@{} >{\raggedright\arraybackslash}p{0.46\linewidth} >{\raggedright\arraybackslash}X @{}}
\toprule
\expath{multi_agent/ilqgames_three_agent.py} & iLQGames differential game; multi-agent \\
\expath{multi_agent/ilqgames_three_agent_lqr.py} & iLQGames with soft collision costs \\
\expath{multi_agent/multi_agent_circle_swap.py} & Multi-agent CTCS collision avoidance; vmap \\
\bottomrule
\end{tabularx}
\endgroup
\vspace{0.6\baselineskip}

\medskip
\noindent\textbf{Pendulum}\\[0.35em]
\begingroup
\footnotesize
\renewcommand{\arraystretch}{1.15}
\begin{tabularx}{\linewidth}{@{} >{\raggedright\arraybackslash}p{0.46\linewidth} >{\raggedright\arraybackslash}X @{}}
\toprule
\expath{pendulum/pendulum_swingup_ocp.py} & Classic swing-up OCP; CTCS \\
\bottomrule
\end{tabularx}
\endgroup
\vspace{0.6\baselineskip}

\medskip
\noindent\textbf{Race Cars}\\[0.35em]
\begingroup
\footnotesize
\renewcommand{\arraystretch}{1.15}
\begin{tabularx}{\linewidth}{@{} >{\raggedright\arraybackslash}p{0.46\linewidth} >{\raggedright\arraybackslash}X @{}}
\toprule
\expath{race_cars/race_car_hybrid.py} & Minimum-lap-time; MPCC; Qpax; batching \\
\expath{race_cars/race_car_ice.py} & Minimum-lap-time ICE ablation; MPCC \\
\expath{race_cars/race_car_mpc.py} & Receding-horizon race-car MPC \\
\expath{race_cars/race_car_multi_agent.py} & Batched multi-agent MPC; Parameters \\
\expath{race_cars/race_car_multi_agent_mpcc.py} & Batched multi-agent MPCC tracking \\
\expath{race_cars/race_car_multi_agent_mpcc_ice.py} & Batched multi-agent MPCC (ICE) \\
\expath{race_cars/race_car_openscvx.py} & Minimum-lap-time single-shot SCP \\
\expath{race_cars/race_car_viser.py} & Viser animation of race-car trajectory \\
\bottomrule
\end{tabularx}
\endgroup
\vspace{0.6\baselineskip}

\medskip
\noindent\textbf{Realtime interactive}\\[0.35em]
\begingroup
\footnotesize
\renewcommand{\arraystretch}{1.15}
\begin{tabularx}{\linewidth}{@{} >{\raggedright\arraybackslash}p{0.46\linewidth} >{\raggedright\arraybackslash}X @{}}
\toprule
\expath{realtime/3DoF_pdg_realtime.py} & Realtime Parameter updates; interactive Viser \\
\expath{realtime/6DoF_pdg_realtime.py} & Realtime Parameter updates; 6-DoF PDG \\
\expath{realtime/cinema_vp_realtime.py} & Realtime viewpoint replan; Parameters \\
\expath{realtime/drone_racing_realtime.py} & Realtime gate racing; Parameters \\
\expath{realtime/dubins_car_realtime.py} & Realtime Dubins replan; Parameters \\
\expath{realtime/obstacle_avoidance_realtime.py} & Realtime obstacle Parameter updates \\
\bottomrule
\end{tabularx}
\endgroup
\vspace{0.6\baselineskip}

\medskip
\noindent\textbf{Rocket / PDG}\\[0.35em]
\begingroup
\footnotesize
\renewcommand{\arraystretch}{1.15}
\begin{tabularx}{\linewidth}{@{} >{\raggedright\arraybackslash}p{0.46\linewidth} >{\raggedright\arraybackslash}X @{}}
\toprule
\expath{rocket/3DoF_pdg.py} & 3-DoF PDG; Parameters; free final time \\
\expath{rocket/6DoF_pdg.py} & 6-DoF PDG; CTCS; free final time \\
\expath{rocket/6DoF_pdg_batched_ic.py} & solve\_batched over initial conditions \\
\expath{rocket/6DoF_pdg_stc.py} & Compound state-triggered constraints (cSTC) \\
\expath{rocket/6DoF_pdg_stc_ifthen.py} & cSTC via ox.stl.IfThen \\
\expath{rocket/ascent_launch_vehicle.py} & Multiphase ascent; impulsive staging \\
\expath{rocket/reusable_launch_vehicle.py} & RLV entry / landing; Parameters \\
\expath{rocket/rocket_ilqr.py} & iLQR-style rocket landing OCP \\
\bottomrule
\end{tabularx}
\endgroup
\vspace{0.6\baselineskip}

\medskip
\noindent\textbf{NASA SENSS}\\[0.35em]
\begingroup
\footnotesize
\renewcommand{\arraystretch}{1.15}
\begin{tabularx}{\linewidth}{@{} >{\raggedright\arraybackslash}p{0.46\linewidth} >{\raggedright\arraybackslash}X @{}}
\toprule
\expath{rocket/senss/6DoF_pdg_dem.py} & DEM terrain; realtime Parameters; Viser \\
\expath{rocket/senss/6DoF_pdg_dem_static.py} & DEM terrain landing (static solve) \\
\expath{rocket/senss/6DoF_pdg_stc_senss.py} & cSTC on SENSS DEM terrain \\
\expath{rocket/senss/6DoF_pdg_stc_senss_gimble.py} & cSTC + gimbal triggers on DEM \\
\expath{rocket/senss/6DoF_pdg_stc_senss_gimble_hop.py} & cSTC hop trajectory; DEM \\
\expath{rocket/senss/6DoF_pdg_stc_senss_hop.py} & cSTC hop trajectory; DEM \\
\expath{rocket/senss/6DoF_pdg_stc_senss_planar.py} & Planar cSTC on DEM \\
\bottomrule
\end{tabularx}
\endgroup
\vspace{0.6\baselineskip}

\medskip
\noindent\textbf{Spacecraft}\\[0.35em]
\begingroup
\footnotesize
\renewcommand{\arraystretch}{1.15}
\begin{tabularx}{\linewidth}{@{} >{\raggedright\arraybackslash}p{0.46\linewidth} >{\raggedright\arraybackslash}X @{}}
\toprule
\expath{spacecraft/dual_deputy_inspection_cw.py} & CW dual-deputy; LoS inspection; impulsive \\
\expath{spacecraft/dual_quaternion_rendezvous.py} & Dual-quaternion SE(3); FOV / glide-slope \\
\expath{spacecraft/halo_orbit.py} & CR3BP Halo orbit IC definition \\
\expath{spacecraft/hohmann_transfer.py} & Impulsive delta-v; Hohmann transfer \\
\expath{spacecraft/let_transfer.py} & Low-energy CR3BP transfer; impulsive \\
\expath{spacecraft/proxops_cw.py} & CW proximity operations; CTCS \\
\expath{spacecraft/relative_loitering.py} & CR3BP relative loitering; CTCS regularization \\
\bottomrule
\end{tabularx}
\endgroup
\vspace{0.6\baselineskip}

\section{Atomic Operators}
\label{apx:atoms}

\Crefrange{tab:math_ops}{tab:advanced_ops} catalog the atomic operators \openscvx provides, grouped by domain.
Because every operator composes freely with the rest of the vocabulary, this single set of primitives suffices to express arbitrary dynamics, costs, and constraints necessary to formulate \cref{prob:mayer}.
The \jax{} and \cvxpy{} columns record which translation rules each operator carries: an operator with a \cvxpy{} rule may appear in convex constraints handed directly to the solver, while a \jax-only operator can still appear anywhere in the nonconvex pipeline, where lowering to \jax{} precedes linearization (\cref{sec:lowering}).

\paragraph{Basic Mathematical Operators}

\Cref{tab:math_ops} collects the core primitives: elementwise arithmetic and comparisons, linear-algebra operations, array stacking, and trigonometric functions.
This modest vocabulary is already sufficient for most complete applications. Typical vehicle dynamics, costs and constraints are built from these operators alone. The families in the tables that follow extend it into more specialized domains and problems.

\begin{table}[!t]
    \caption{Basic Mathematical Operators}
    \label{tab:math_ops}
    \centering
    \footnotesize
    \renewcommand{\arraystretch}{1.25}
    \begin{tabularx}{\linewidth}{@{} >{\ttfamily}l X cc @{}}
        \toprule
        \multicolumn{1}{@{}l}{\textbf{Arithmetic}} & & \jax{} & \cvxpy{} \\
        \midrule
        $+, -, *, /$ & Addition, subtraction, multiplication, division & \checkmark & \checkmark \\
        $\leq$, == & Inequality, equality & \checkmark & \checkmark \\
        @ & Matrix multiplication & \checkmark & \checkmark \\
        ** & Power & \checkmark & \checkmark \\
        Sqrt() & Element-wise square root & \checkmark & \checkmark \\
        Exp() & Element-wise exponential & \checkmark & \checkmark \\
        Log() & Element-wise logarithm & \checkmark & \checkmark \\
        Abs() & Element-wise absolute value & \checkmark & \checkmark \\
        Min(), Max() & Element-wise minimum and maximum over operands & \checkmark & \checkmark \\
        \addlinespace
        \multicolumn{1}{@{}l}{\textbf{Linear Algebra}} & & & \\
        \midrule
        .T & Transpose & \checkmark& \checkmark \\
        Diag() & Square diagonal matrix from 1D vector & \checkmark & \checkmark \\
        Inv() & Inverse of square matrix & \checkmark \\
        Norm() & Norm of an expression & \checkmark & \checkmark \\
        Sum() & Sum of all elements of an expression & \checkmark & \checkmark \\
        \addlinespace
        \multicolumn{1}{@{}l}{\textbf{Array}} & & & \\
        \midrule
        Vstack() & Vertical stacking for expressions & \checkmark & \checkmark \\
        Hstack() & Horizontal stacking for expressions & \checkmark & \checkmark \\
        Block() & Block matrix/tensor from nested expression arrays & \checkmark & \checkmark \\
        \addlinespace
        \multicolumn{1}{@{}l}{\textbf{Trigonometric}} & & & \\
        \midrule
        Sin(), Asin() & Element-wise sine and arcsine & \checkmark \\
        Cos(), Acos() & Element-wise cosine and arccosine & \checkmark \\
        Tan(), Atan() & Element-wise tangent and arctangent & \checkmark \\
        Atan2() & Two-argument arctangent & \checkmark \\
        \bottomrule
    \end{tabularx}
\end{table}

\paragraph{Logic \& Signal-Temporal Logic Operators}
The logical operators in \cref{tab:advanced_ops} serve two roles.
General Boolean primitives such as \texttt{All}, \texttt{Any}, and \texttt{Cond} express branching and aggregation, while \gls{stl} operators provide a robust mathematical framework for encoding complex mission-level specifications, such as sequencing, state-dependent behaviors, deadlines, and conditional obligations directly into continuous \glspl{ocp}.
\gls{stl} expressions are enforced smoothly and exactly through \gls{gmsr} \cite{uzun2024optimizationtemporallogicalspecifications, uzun2026successive} and are represented in the \jax{} backend.

\paragraph{Spatial Operators}
Rigid-body models, from quadrotors to manipulators, require rotations and homogeneous transforms.
The spatial operators in \cref{tab:advanced_ops} provide quaternion utilities alongside the exponential, logarithmic, and adjoint maps of the matrix Lie groups $\mathrm{SO}(3)$ and $\mathrm{SE}(3)$.
Under the hood, these Lie group operations are lowered to the \textsf{jaxlie} package \cite{yi2021iros}, demonstrating how new nodes can leverage the existing \jax{} ecosystem.

The vocabulary shown here is representative rather than exhaustive, and, as \cref{sec:lowering} described, it is built to grow.
With the operators established, the remainder of this section turns from what \openscvx \emph{can} express to \emph{how} a user expresses it, assembling the expression \gls{dag} of \cref{sec:symbolic-expression-layer} piece by piece, beginning with the leaf nodes.
 
\begin{table}[!t]
    \caption{Specialized Operators (\jax-only)}
    \label{tab:advanced_ops}
    \centering
    \footnotesize
    \renewcommand{\arraystretch}{1.25}
    \begin{tabularx}{\linewidth}{@{} >{\ttfamily}l X @{}}
        \toprule
        \multicolumn{1}{@{}l}{\textbf{Logic \& Signal Temporal Logic}} & \\
        \midrule
        All(), Any() & Logical AND / OR over all predicates \\
        And(), Or(), Not() & Conjunction, disjunction, negation \\
        Always() & Temporal operator, formula holds over a node or time interval \\
        IfThen() & Implication \\
        Cond() & Conditional, selects a branch by predicate \\
        IntegerVariable() & Discrete/integer variable \\
        \addlinespace
        \multicolumn{1}{@{}l}{\textbf{Spatial \& Lie Groups}} & \\
        \midrule
        QDCM() & Unit quaternion to direction cosine matrix \\
        SSM() & $3 \times 3$ skew-symmetric (cross-product) matrix of a vector in $\mathbb{R}^3$ \\
        SSMP() & $4 \times 4$ skew-symmetric matrix $\Omega(\omega)$ for quaternion kinematics \\
        SO3Exp() & Exponential map $\mathfrak{so}(3) \to \mathrm{SO}(3)$, 3D rotation vector to $3 \times 3$ rotation matrix \\
        SO3Log() & Logarithmic map $\mathrm{SO}(3) \to \mathfrak{so}(3)$, $3 \times 3$ rotation matrix to 3D rotation vector \\
        SE3Exp() & Exponential map $\mathfrak{se}(3) \to \mathrm{SE}(3)$, 6D twist to $4 \times 4$ homogeneous transform \\
        SE3Log() & Logarithmic map $\mathrm{SE}(3) \to \mathfrak{se}(3)$, $4 \times 4$ homogeneous transform to 6D twist \\
        Adjoint() & Adjoint action $\mathrm{ad}_{\xi}$ (Lie bracket) on twists \\
        AdjointDual() & Coadjoint action $\mathrm{ad}^*_{\xi}$ on momenta, e.g.\ Coriolis and centrifugal terms \\
        \bottomrule
    \end{tabularx}
\end{table}

\section{External Dynamics Integrations}
\label{apx:integrations}

The flexibility of \jax{} enables seamless integration with a broad ecosystem of \jax-based packages.
\mujoco{} is a highly prevalent physics engine in robotics that features \textsf{MuJoCo XLA}, exposing \jax-based dynamics functions for its model library; \openscvx can parse \mujoco{} XML models directly via \texttt{MjxDynamics}.
However, the current implementation of contact dynamics within \textsf{MuJoCo XLA} does not support forward differentiation, necessitating the disabling of contact during optimization.
Complementarily, \fraxpkg{} provides URDF-based rigid-body kinematics and dynamics in \jax{} and integrates through \texttt{FraxDynamics}, which auto-populates joint and torque bounds from the robot description. \fraxpkg{} provides Franka Panda and Unitree G1 dynamics models.

\begin{listing}
\caption{Interfacing with \mujoco{} \textsf{MJX} dynamics in \openscvx.}
\begin{lstlisting}[style=oxapi]
import mujoco
import mujoco.mjx as mjx
import openscvx as ox

# 1. Load standard MuJoCo model from XML
mj_model = mujoco.MjModel.from_xml_path("skydio_x2.xml")

# 2. Disable contact solver
mj_model.opt.disableflags |= mujoco.mjtDisableBit.mjDSBL_CONTACT

# 3. Convert to MJX model
mjx_model = mjx.put_model(mj_model)

# 4. Pass to OpenSCvx MJX dynamics adapter
dyn = ox.MjxDynamics(mjx_model)

# 5. Extract state and control objects for optimization
qpos, qvel = dyn.states
(ctrl,) = dyn.controls
\end{lstlisting}
\end{listing}

\begin{listing}
\caption{Interfacing with \fraxpkg{} dynamics in \openscvx.}
\begin{lstlisting}[style=oxapi]
import frax
import openscvx as ox

# 1. Load robot model from URDF (bundled with frax)
robot = frax.load_panda()

# 2. Pass to OpenSCvx frax dynamics adapter
dyn = ox.FraxDynamics(robot)

# 3. Extract state and control objects for optimization
q, qd = dyn.states
(tau,) = dyn.controls
\end{lstlisting}
\end{listing}

\ifprintgloss
  \printglossary[type=\acronymtype,title={List of Acronyms},nonumberlist]
\fi
\end{appendices}

\end{document}